\documentclass[10pt, conference, compsocconf]{IEEEtran}

\usepackage{times}
\usepackage{epsfig}
\usepackage{graphicx}
\usepackage{amsmath}
\usepackage{amssymb}
\usepackage{calc}  

\let\labelindent\relax
\usepackage{enumitem}  

\usepackage{subfig}
\usepackage{lib}
\usepackage{booktabs}
\usepackage{multirow}

\usepackage{pgf}

\begin{document}
%
\title{Hierarchical Surface Prediction for 3D Object Reconstruction}


\author{\IEEEauthorblockN{Christian H{\"a}ne, Shubham Tulsiani, Jitendra Malik}
\IEEEauthorblockA{University of California, Berkeley \\
{\tt\small\{chaene, shubhtuls, malik\}@eecs.berkeley.edu}}
}

\maketitle

\begin{abstract}
Recently, Convolutional Neural Networks have shown promising results for 3D geometry prediction. They can make predictions from very little input data such as a single color image.
A major limitation of such approaches is that they only predict a coarse resolution voxel grid, which does not capture the surface of the objects well.
We propose a general framework, called hierarchical surface prediction (HSP), which facilitates prediction of high resolution voxel grids. 
The main insight is that it is sufficient to predict high resolution voxels around the predicted surfaces.
The exterior and interior of the objects can be represented with coarse resolution voxels.
Our approach is not dependent on a specific input type. We show results for geometry prediction from color images, depth images and shape completion from partial voxel grids. Our analysis shows that our high resolution predictions are more accurate than low resolution predictions.
\end{abstract}

\IEEEpeerreviewmaketitle

\section{Introduction}

We live in a world composed of 3D objects bounded by 2D surfaces. Fundamentally, this means that we can either represent geometry implicitly as 3D volume or explicitly as 2D mesh surface which lives within the 3D space. When working with 3D geometry in practice for many tasks a specific representation is more suited than the other. 

\begin{figure}
 \subfloat[Input Image]{\includegraphics[width=0.31\linewidth, trim={0.4cm 1.1cm 0.4cm 1.1cm}, clip]{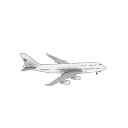}}
 \subfloat[$16^3$]{\includegraphics[width=0.31\linewidth, trim={3.5cm 5.5cm 3.5cm 5.5cm}, clip]{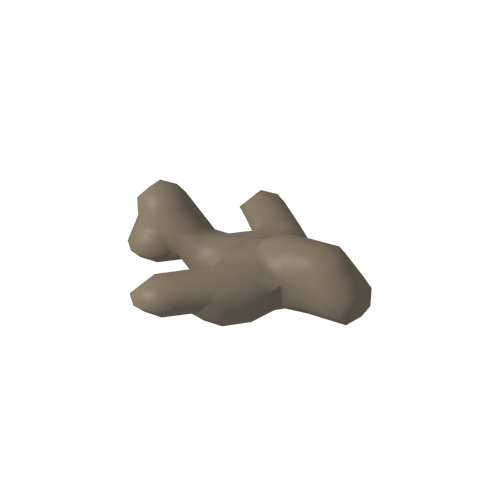}}
 \subfloat[$32^3$]{\includegraphics[width=0.31\linewidth, trim={3.5cm 5.5cm 3.5cm 5.5cm}, clip]{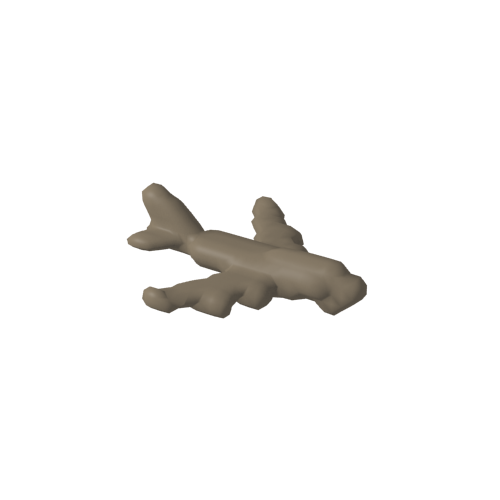}} \\
 \subfloat[$64^3$]{\includegraphics[width=0.31\linewidth, trim={3.5cm 5.5cm 3.5cm 5.5cm}, clip]{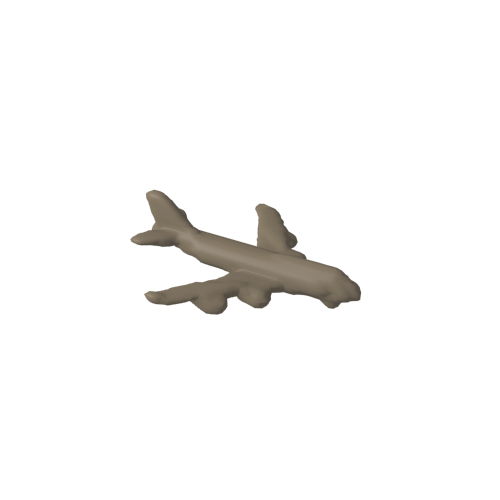}}
 \subfloat[$128^3$]{\includegraphics[width=0.31\linewidth, trim={3.5cm 5.5cm 3.5cm 5.5cm}, clip]{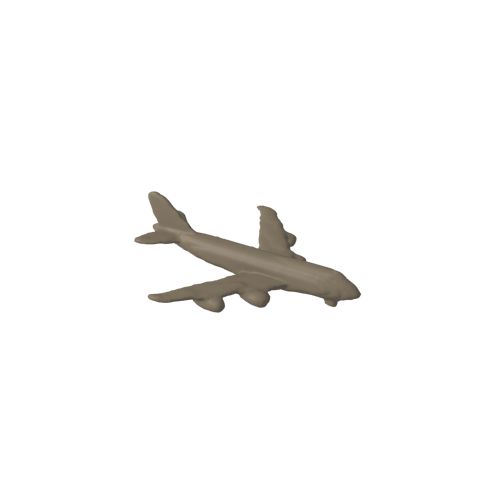}}
 \subfloat[$256^3$]{\includegraphics[width=0.31\linewidth, trim={3.5cm 5.5cm 3.5cm 5.5cm}, clip]{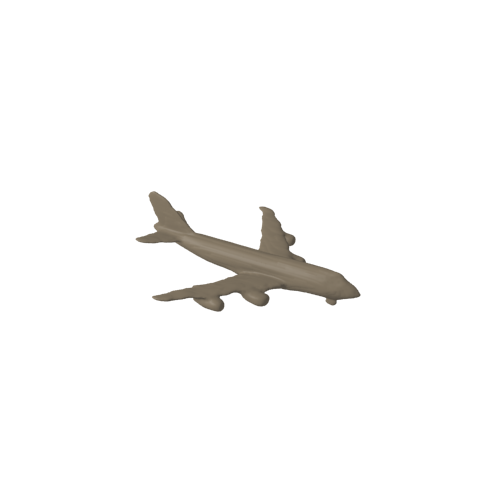}}
 \caption{Illustration of our method on the task of predicting geometry from a single color image. From the image our method hierarchically generates a surface with increasing resolution. In order to have sufficient information for the prediction of the higher resolution we predict at each level multiple feature channels which serve as input for the next level and at the same time allow us to generate the output of the current level. We start with a resolution of $16^3$ and divide the voxel side length by two on each level reaching a final resolution of $256^3$.}
 \label{fig:teaser}
\end{figure}

Recently, 3D prediction approaches which directly learn a function to map an input image to the output geometry have emerged \cite{choy20163d,girdhar2016learning,yan2016perspective}. The function is represented as Convolutional Neural Network (CNN) and the geometry is represented as voxel grid, which is a natural choice for CNNs. The advantage of such an approach is that it can represent arbitrary topology and allows for large shape variations. However, such approaches have a major limitation. Due to the cubic growth of the volume with increasing resolution only a coarse voxel grid is predicted. A common resolution is $32\times32\times32$. Using higher resolutions very quickly becomes computationally infeasible. Moreover, due to the 2D nature of the surface, the ratio between surface or close to the surface and non-surface voxels becomes smaller and smaller with increasing resolution. Voxels far from the surface are generally predicted correctly quite early in the training and hence do not induce any gradients which can be used to teach the system where the surface lies. Therefore, with increasing resolution also the work done on computing gradients which are uninformative increases. The underlying reason why this problem exists is that current systems do not exploit the fact that surfaces are only two dimensional. Can we build a system that does?

In this paper we introduce a general framework, hierarchical surface prediction (HSP), for high resolution 3D object reconstruction which is organized around the observation that only a few of the voxels are in the vicinity of the object's surface \ie the boundary between free and occupied space, while most of the voxels will be ``boring'', either completely inside or outside the object. The basic principle therefore is to only predict voxels around the surface, which we can now afford to do at higher resolution.  The key insight to enable our method to achieve this is to change the standard prediction of free and occupied space into a three label prediction with the labels \emph{free space}, \emph{boundary} and \emph{occupied space}.  Furthermore, we do not directly predict high resolution voxels. Instead, we hierarchically predict small blocks of voxels from coarse to fine resolution in an octree, which we call voxel block octree. Thereby we have at each resolution the signal of the \emph{boundary} label which tells us that the descendants of the current voxel will contain both,  \emph{free space} and \emph{occupied space}. Therefore, only nodes containing voxels with the \emph{boundary} label assigned need higher resolution prediction. By hierarchically predicting only those voxels we build up our octree. This effectively reduces the computational cost and hence allows us to predict significantly higher resolution voxel grids than previous approaches. In our experiments we predict voxels at a resolution of $256\times256\times256$ (\cf Fig.\ \ref{fig:teaser}).

One of the fundamental questions which arises once higher resolution predictions are feasible, is whether or not a higher resolution prediction leads to more accurate results. In order to analyze this we compare our high resolution predictions to upsampled low resolution predictions in our quantitative evaluation and determine that our method outperforms these baselines. Our results are not only quantitatively more accurate, they are also qualitatively more detailed and have a higher surface quality.

The remainder of the paper is organized as follows. In Sec.\ \ref{sec:relWork} we discuss the related work. We then introduce our framework in Sec.\ \ref{sec:formulation}. Quantitative and qualitative evaluations are done in Sec.\ \ref{sec:exp} and eventually we draw the conclusions in Sec.\ \ref{sec:conclusion}.

\section{Related Work}
\label{sec:relWork}

Traditionally dense 3D reconstruction from images has been done using a large collection of images. Geometry is extracted by dense matching or direct minimization of reprojection errors. The common methods can be broadly grouped as implicit volumetric reconstruction~\cite{curless1996volumetric, kolev2007continuous, kutulakos2000theory, lempitsky2007global, zach2007globally} or explicit mesh based~\cite{delaunoy2008minimizing, gargallo2007minimizing} approaches. All these approaches facilitate a reconstruction of arbitrary geometry. However, in general a lot of input data is required to constrain the geometry enough. In difficult cases it is not always possible to recover the geometry of an object with multi-view stereo matching. This can be due to challenging materials such as transparent and reflective surfaces or lack of images from all around an object. For such cases object shape priors have been proposed~\cite{dame2013dense, hane2014class,  yingze2013dense}.

All the methods mentioned above in general use multiple images as input data. Approaches which are primarily designed to reconstruct 3D geometry from a single color image were also studied. Vetter \etal~\cite{blanz1999morphable} propose to build a morphable shape model for faces, which allows for the reconstruction of faces from just a single image. In order to build such a model a lot of manual interaction and high quality scanning of faces is required. Other works propose to learn a shape model from silhouettes~\cite{cashman2013shape} or silhouettes and keypoints~\cite{kar2015category}.

Recently, CNNs have also been used to predict geometry from single images. Choy \etal~\cite{choy20163d} propose to use an encoder decoder architecture in a recurrent network to facilitate prediction of coarse resolution voxel grids from a single or multiple color images. An alternative approach~\cite{girdhar2016learning} proposes to first train an autoencoder on the coarse resolution voxel grids and then train a regressor which predicts the shape code from the input image. It has also been shown that CNNs can be leveraged to predict alternative primitive based representations~\cite{abstractionTulsiani17}. While these approaches rely on ground truth shapes, CNNs can also be trained to predict 3D shapes using weaker supervision~\cite{rezende2016unsupervised,yan2016perspective,tulsiani2017ray} for example image silhouettes. These types of supervisions utilize ray formulations which originally have been proposed for multi-view reconstruction \cite{liu2011complete,savinov2016semantic}.

Predicting geometry from color images has also been addressed in a 2.5D setting where the goal is, given a color image to predict a depth map~\cite{eigen2014depth,ladicky2014pulling,saxena2005learningdepth}. These approaches can generally produce high resolution output but do only capture the geometry from a single viewpoint. Tatarchenko \etal~\cite{tatarchenko2016multi} propose to predict a collection of RGBD (color and depth) images from a single color image and fuse them in a post-processing step.

The volumetric approach has the major limitation that using high resolution voxel grids is computationally demanding and memory intensive. Multi-view approaches alleviate this, by the use of data adaptive discretization of the volume with Delaunay tedrahedrization~\cite{labatut2007efficient}, voxel block hashing~\cite{niessner2013real} or octrees~\cite{chen2013scalable, steinbrucker2014volumetric}. Our work is inspired by these earlier works on octrees. A crucial difference is that in these works the structure of the octree can be determined from the input geometry but in our case we are looking at a more general problem where the structure of the tree needs to be predicted. Similarly, a very recent work~\cite{riegler2016octnet} which proposes to use octrees in a CNN,  also assumes that the structure of the octree is given as input. We predict the structure of the tree together with its content.

Our hierarchical surface prediction framework also relates to coarse-to-fine approaches in optical flow computation. Similar to volumetric 3D reconstruction also dense optical flow is computationally demanding due the the large label space. By computing optical flow hierarchically on a spatial pyramid, e.g.~\cite{brox2004high}, real-time computation of optical flow is feasible~\cite{zach2007duality}. The benefits of using a coarse-to-fine approach in learning based optical flow prediction has also recently been pointed out~\cite{ranjan2016optical,ilg2016flownet}.

\section{Formulation}
\label{sec:formulation}

In this Section we describe our hierarchical surface prediction framework, which allows for high resolution voxel grid predictions. We first briefly discuss how the state-of-the-art, coarse resolution baseline works and then introduce the voxel block octree structure which we are predicting.

\begin{figure*}
 \includegraphics[width=0.95\linewidth]{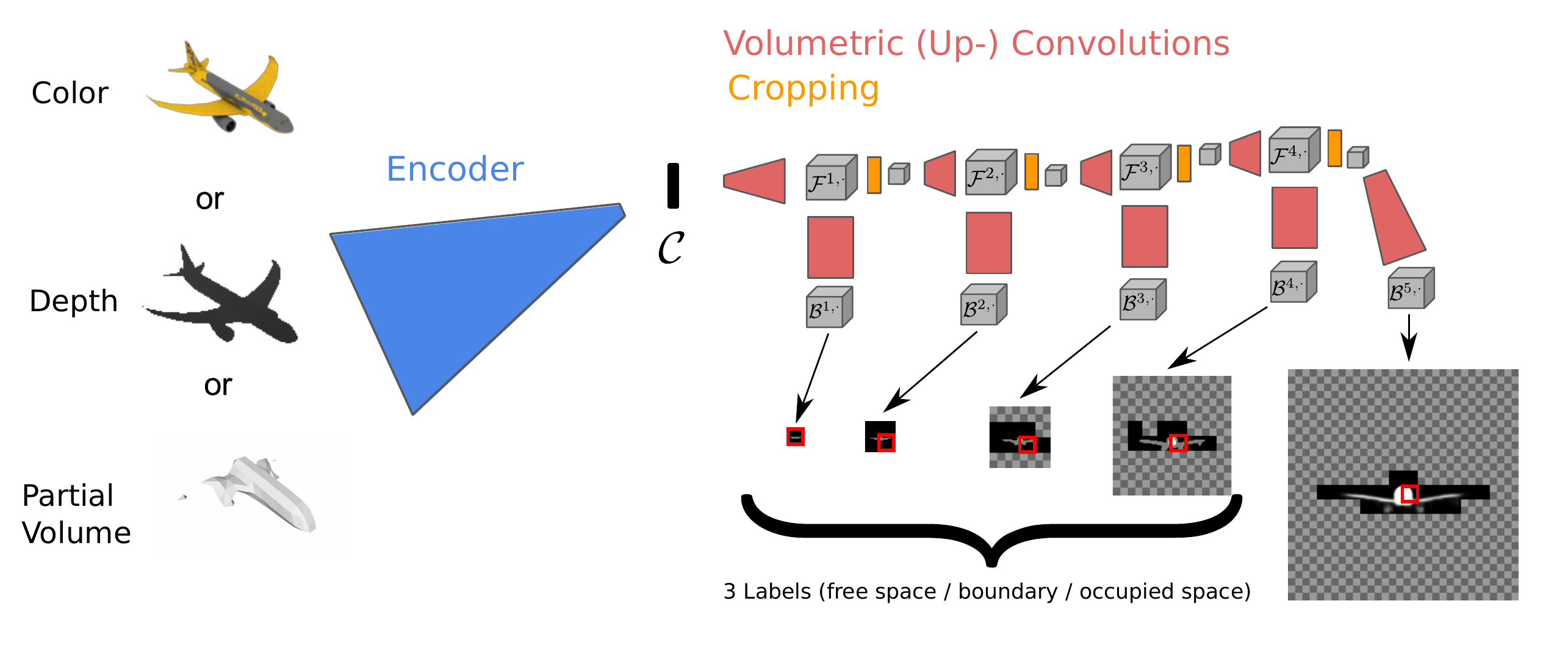}
 \caption{Overview of our system}
 \label{fig:architecture}
\end{figure*}

\subsection{Voxel Prediction}
\label{sec:voxPred}

The basic voxel prediction framework is adapted from \cite{choy20163d,girdhar2016learning}. We consider an encoder/decoder architecture which takes an input $\mathcal{I}$, which in our experiments is either a color image, depth image or a partial low resolution voxel grid. A convolutional encoder  encodes the input to a feature vector or shape code $\mathcal{C}$ which, in all our experiments, is a 128 dimensional vector. An up-convolutional decoder then decodes $\mathcal{C}$ into a predicted voxel grid $\mathcal{V}$. A labeling of the voxel space into free and occupied space can be determined using a suitable threshold $\tau$ (\cf Sec.\ \ref{sec:quantEval}) or alternatively a mesh can be extracted using marching cubes \cite{lorensen1987marching} directly on the predicted voxel occupancies at an iso value $\sigma$.

The main problem which prevents us from directly utilizing this formulation with high resolutions for $\mathcal{V}$ is that each voxel, even if it is far from the surface, needs to be represented and a prediction is made for it. Given that the surface area grows quadratically and the volume cubically  with respect to edge division, the ratio of surface to non-surface voxels becomes smaller with increasing resolution.

\subsection{Voxel Block Octree}

In our hierarchical surface prediction method, we propose to predict a data structure with an up-convolutional decoder architecture, which we call `voxel block octree'. It is inspired from octree formulations \cite{chen2013scalable,steinbrucker2014volumetric} used in traditional multi-view reconstruction approaches. The key insight which allows us to use such a data structure in a prediction framework is to extend the standard two label formulation to a three label formulation with labels \emph{inside}, \emph{boundary} and \emph{outside}. As we will see later our data structure allows us to generate a complete voxel grid at high resolution while only making predictions around the surface. This leads to an approach which facilitates efficient training of an encoder/decoder architecture end-to-end.

An octree is a tree data structure which is used to partition the 3D space. The root node describes the cube of interest in the 3D space. Each internal node has up to 8 child nodes which describe the subdivision of the current node's cube into the eight octants. Note that we slightly deviate from the standard definition of the octree where either none or all the 8 child nodes are present.

\begin{figure}
    \centering
    \hfill
    \includegraphics[width=0.31\linewidth]{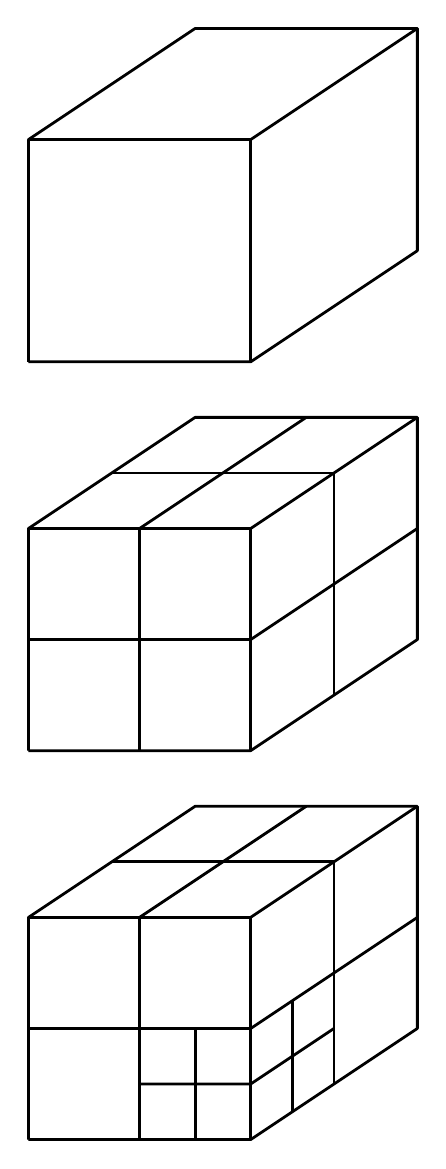} \hfill
    \includegraphics[width=0.31\linewidth]{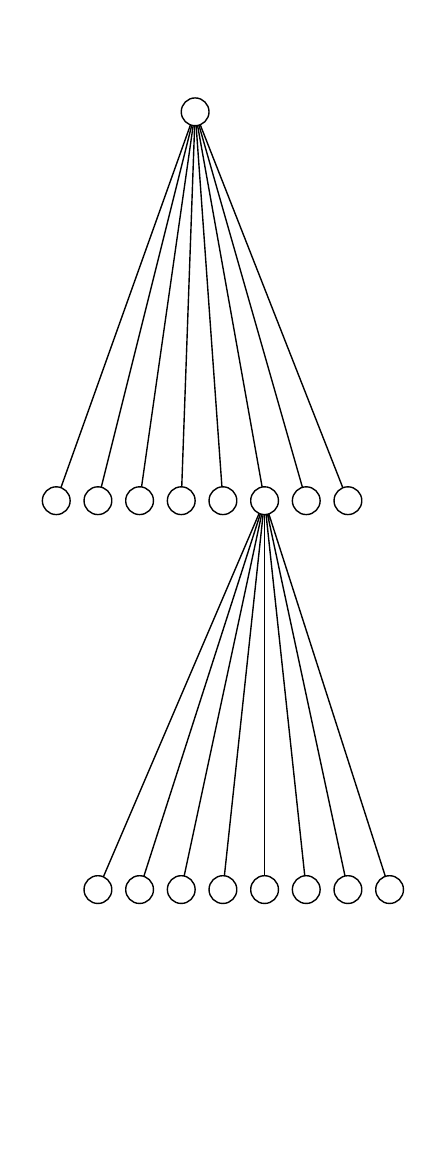} \hfill
    \includegraphics[width=0.19\linewidth]{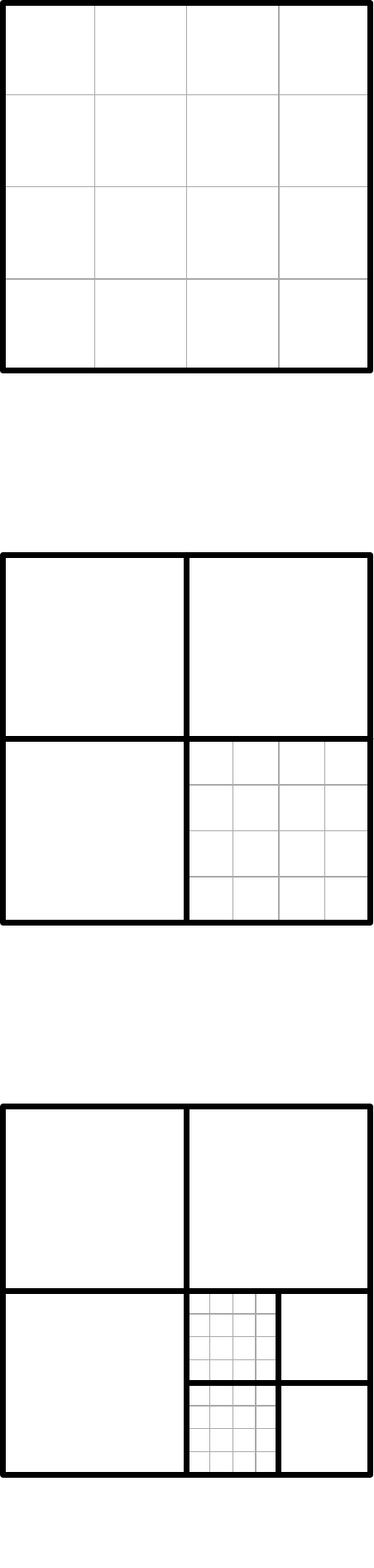} \hspace{0.2cm}
    \caption{Visualization of the voxel block octree. The (left) part depicts the space subdivision which is induced by the octree in the (middle). The (right) side is a 2D visualization of parts of the voxel blocks with a voxel block size of $b=4$.}
    \label{fig:treeVis}
\end{figure}

We consider a tree with levels $\ell \in \{1,\ldots,L\}$. Each node of the tree contains a `voxel block' -- a subdivision of the node's 3D space into a voxel grid of size $b^3$. Each voxel in the voxel block at tree levels $\ell < L$ contains classifier responses for the three labels \emph{occupied space}, \emph{boundary} and \emph{free space}.
The nodes in layers $\ell \in \{1, \ldots, L-1 \}$, therefore, contain voxel blocks $\mathcal{B}^{\ell,s} \in [0,1]^{b^3 \times 3}$ where the index $s$ describes the position of the voxel block with respect to the global voxel grid. In the lowest level $L$ we do not need the boundary label, therefore it is composed of nodes which contain a voxel block $\mathcal{B}^{\ell,s} \in [0,1]^{b^3}$, where each element of the block describes the classifier response for the binary classification into  free or occupied space.
Note that each child node describes a space which has a cube side length of only half of the current node. By keeping the voxel block resolution fixed at $b^3$ we also divide the voxel side length by a factor of two. This means that the prediction in the child nodes are of higher resolution. 
The voxel block resolution $b$ is chosen such that it is big enough for efficient prediction using an up-convolutional decoder network and at the same time small enough such that local predictions around the surface can be made. In our experiments we use voxel blocks with $b=16$ and a tree with depth $L = 5$. A visualization of the voxel block octree is given in Fig.\ \ref{fig:treeVis}.

The predictions stored at any given level of the tree might be sparse. In order to reconstruct a complete model in high resolution, we upsample all the voxels which have not been predicted on the highest resolution from the closest predicted resolution. As we will see later, the voxels which get upsampled are in the inside and outside of the object not directly next to the boundary. Therefore the resolution of the actual surface remains high. In order to be able to extract a smooth high quality surface using marching cubes it is crucial that all the voxels which are close to the surface are predicted at high resolution and the predicted values are smooth. Therefore we aim to always evaluate the highest resolution around the boundary.
At this point we would like to note that we can also extract a model at intermediate resolution by considering the boundary label as occupied space. This choice is consistent with the choice of considering any voxel as occupied space which intersects the ground truth mesh during voxelization (\cf Sec.\ \ref{sec:voxelization}).

\subsection{Network Architecture}

In the previous section we introduced our voxel block octree data structure. We will now describe the encoder / decoder architecture which we are using to predict voxel block octrees using CNNs. An overview of our architecture is depicted in Fig.\ \ref{fig:architecture}.

The encoder part is identical to the one of the basic voxel prediction pipeline from Sec.\ \ref{sec:voxPred}, \ie the input $\mathcal{I}$ gets first transformed into a shape code $\mathcal{C}$ by a convolutional encoder network. An up-convolutional also called de-convolutional decoder architecture \cite{dosovitskiy2017learning,zeiler2010deconvolutional} predicts the voxel block octree. In order to be able to predict the blocks $\mathcal{B}^{\ell,s}$ and at the same time also being able to capture the information which allows for predictions of the higher resolution levels $\ell+1, \ldots$, we introduce feature blocks $\mathcal{F}^{\ell,s} \in \mathbb{R}^{(b+2p)^3 \times c}$. The spatial extent of the feature blocks is bigger or equal to the ones of the voxels blocks to allow for a padding $p \geq 0$. As we will see later the padding allows for an overlap when predicting neighbouring octants which leads to smoother results. The fourth dimension is the number of feature channels which can be chosen depending on the application. In our experiments we use $c=32$ and $p=2$.

\begin{figure}
\subfloat[Cropping]{\includegraphics[width=0.48\linewidth]{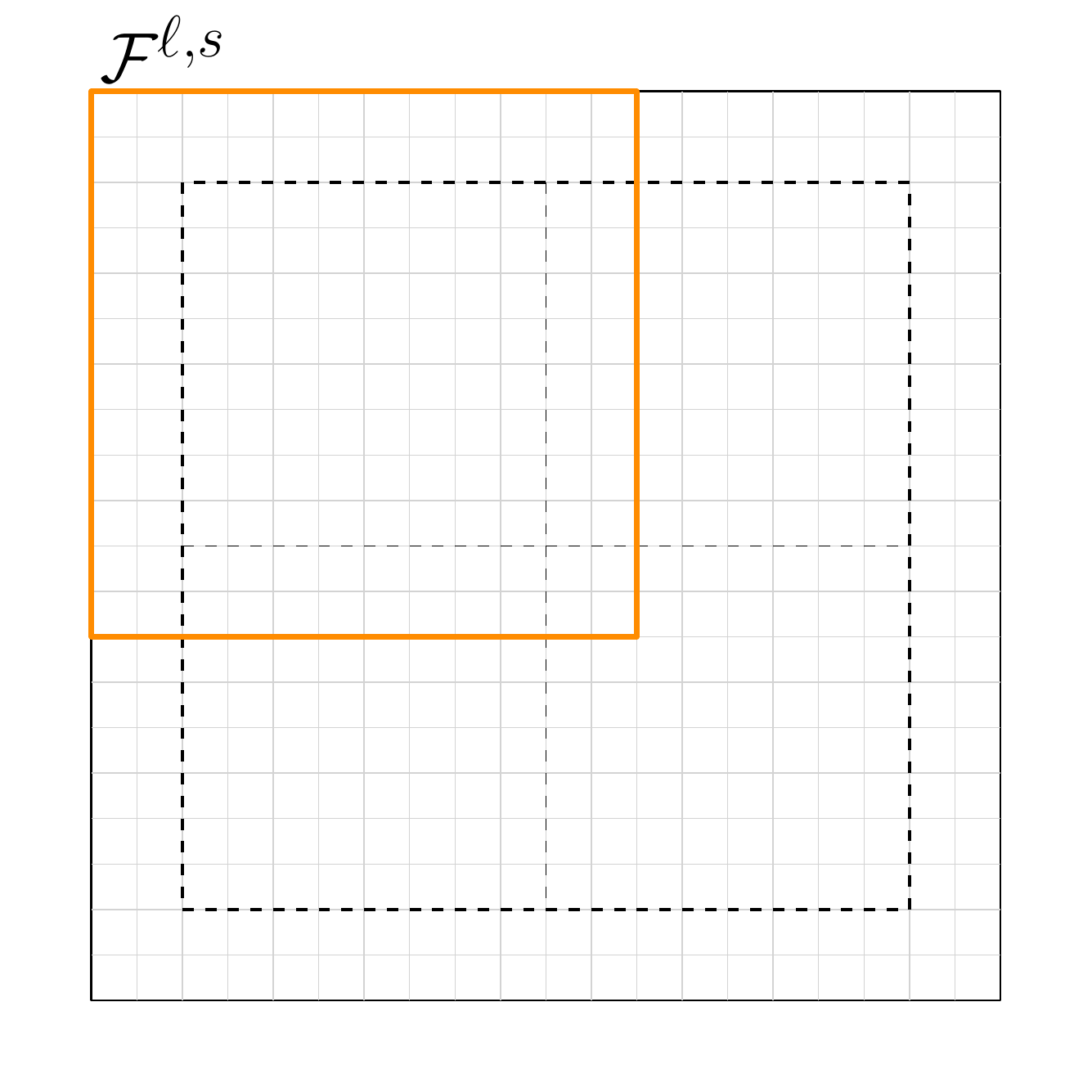}}
\subfloat[Upsampling]{\includegraphics[width=0.48\linewidth]{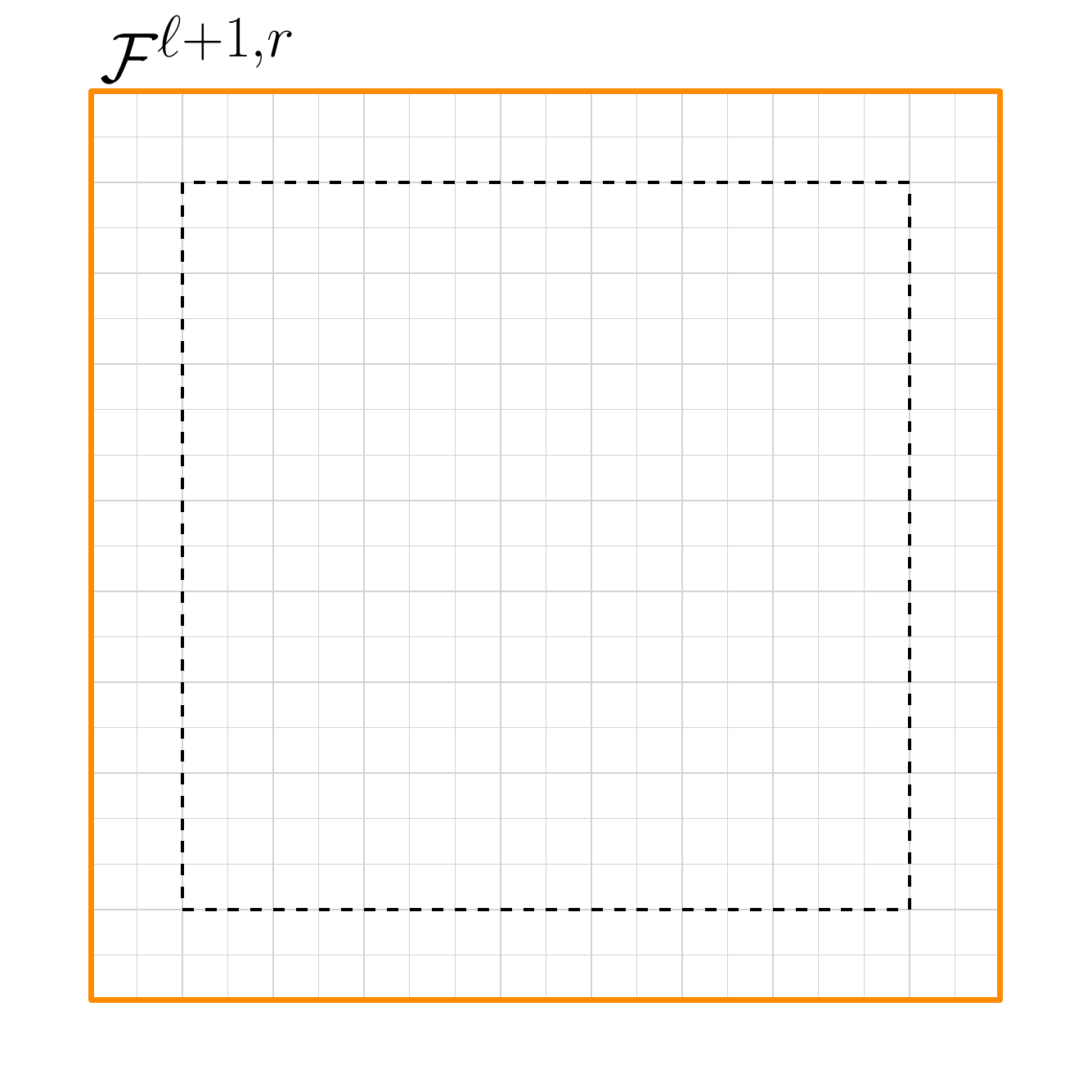}}
\caption{2D visualization of the cropping and upsampling module. The cropping module (left) crops out the part of the feature block centered around the child node's octant. The upsampling module (right) then upsamples the feature map using up-convolutional layers to a new feature block with higher spatial resolution. The dashed lines indicate the size of the output blocks and the octants.}
\label{fig:cropUpsampling}
\end{figure}

The most important part for the understanding of our decoder architecture is how we predict level $\ell+1$ given level $\ell$. In the following we will discuss this procedure in detail. It consist of three basic steps.

\begin{enumerate}
 \item Feature Cropping
 \item Upsampling
 \item Output Generation
\end{enumerate}

\begin{figure}
  \subfloat[Input]{\includegraphics[width=0.31\linewidth]{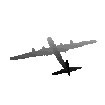}}
  \subfloat[No Overlap]{\includegraphics[width=0.31\linewidth]{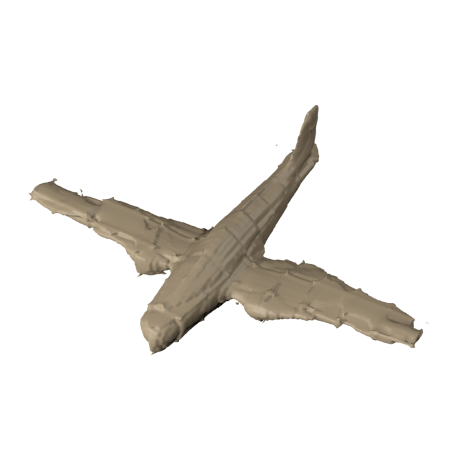}}
  \subfloat[2 Voxels Overlap]{\includegraphics[width=0.31\linewidth]{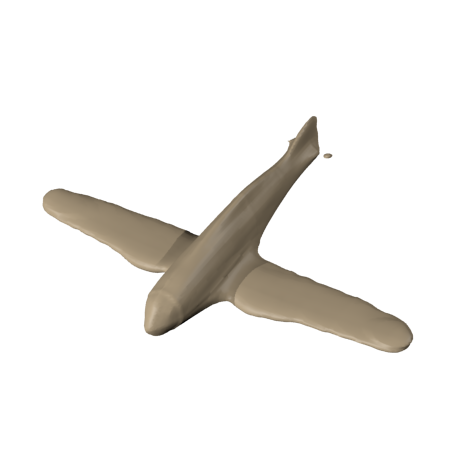}}
 \caption{Influence of overlap on the task of geometry prediction from a single depth map. Without overlap the predictions are much less smooth and the grid pattern induced by the individual voxel blocks is much more visible.}
 \label{fig:overlap}
\end{figure}

\vspace{2mm}
\noindent \textbf{Feature Cropping.}
At this point we assume that we have given the feature block $\mathcal{F}^{\ell,s}$. The goal is to generate the output for the child node corresponding to a specific octant $\mathcal{O}$. The feature block $\mathcal{F}^{\ell,s}$ contains information to generate the output for all the child nodes. In order to only process the information relevant for the prediction of the child node corresponding to $\mathcal{O}$ we extract a $((b/2 + 2p)^3 \times c)$ region out of the four dimensional tensor $\mathcal{F}^{\ell,s}$ spatially centered around $\mathcal{O}$. An illustration of this process is given in Fig.\ \ref{fig:cropUpsampling}. If neighboring octants are processed, the extracted feature channels will have some overlap, which helps to generate a smoother output (see Fig.\ \ref{fig:overlap}). 

\vspace{2mm}
\noindent \textbf{Upsampling.} 
The upsampling module takes input from the cropping module and predicts a new $(b+2p)^3$ feature block $\mathcal{F}^{\ell+1,r}$ via up-convolutional and convolutional layers.

\begin{figure}
 \centering
 \includegraphics[width=0.31\linewidth]{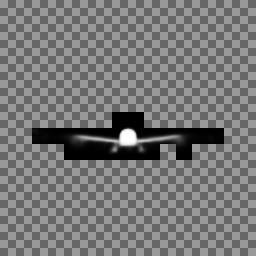}
 \includegraphics[width=0.31\linewidth]{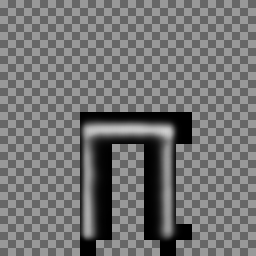}
 \includegraphics[width=0.31\linewidth]{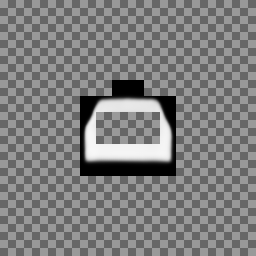}
 \caption{Responses at the highest resolution, the checkered areas indicate not predicted at that resolution, (left) slice through airplane, (middle) slice through front legs of a chair, (right) slice through a car. }
 \label{fig:slices}
\end{figure}

\vspace{2mm}
\noindent \textbf{Output Generation.}
The output network takes the prediction of the feature block $\mathcal{F}^{\ell+1,r}$ from the upsampling module and generates the voxel block $\mathcal{B}^{\ell+1,r}$. This is done using a sequence of convolutional layers. The supervision is given in form of the three ground truth labels for the voxel block $\mathcal{B}^{\ell+1,r}$, \ie there is supervision for each level of the tree. Once the output is generated the child nodes for the next level get generated. The decision on whether to add a child node and hence a higher resolution prediction is based on the boundary prediction in the corresponding octant of the voxel block $\mathcal{B}^{\ell+1,r}$. We compute the maximum boundary prediction response of the corresponding octant $\mathcal{O}'$
\begin{equation}
    C_{\mathcal{O}'}^{\ell+1,r} =\max_{i,j,k \in \mathcal{O}'} \mathcal{B}^{\ell+1,r}_{i,j,k,2} \enspace,
    \label{eq:childCriterion}
\end{equation}
with label $2$ the \emph{boundary} label. The child node is generated if $C_{\mathcal{O}'}^{\ell+1,r}$ is above a small threshold $\gamma$. The intuition behind this choice is that as soon as there is some evidence that the surface is going through a specific block we should predict a higher resolution output. On the other hand if the prediction on a specific level is very certain that there is no boundary within that subtree, a higher resolution prediction is not necessary. For the classes aeroplane, chair and car in practice only around 5 to 10\% of the voxels are predicted at high resolution, in average (\cf Fig.\ \ref{fig:slices}), which shows that our approach effectively reduces the number of voxels that need to be predicted.

In the first level of the tree, which predicts the root node from the shape code $\mathcal{C}$, a small decoding network directly predicts the first feature block without a cropping module in between. Likewise, in the deepest level of the tree, no explicit feature block is needed. Therefore the output is directly generated from the cropped features of the previous level. Also note that the output and upsampling modules have their individual filters at each level of the tree. Furthermore, thanks to this architecture all the convolutions and up-convolutions are standard layers and no special versions for the octree are required. Detailed layer configurations can be found in the appendix.

\subsection{Efficient Training with Subsampling}

As we have seen above, only predicting voxels on the boundary at each level largely reduces the complexity of the prediction task. In the beginning of the training, this is not the case because the boundaries are not yet predicted correctly. Moreover, even once the training has advanced enough and the boundaries are mostly placed correctly, an evaluation of the complete tree is still slow for training (approx. $0.7s$ for only a forward pass).

Therefore, we propose to utilize a subsampling of the child nodes during training. This is possible thanks to our hierarchical structure of the prediction. The tree gets traversed in a depth first manner. Each time the boundary label is present in an octant according to Eq.~\ref{eq:childCriterion} the child node is traversed with a certain probability $\rho$. Different schedules for $\rho$ can be used. We experimented with both, having a fixed probability $\rho$ or start with a very low $\rho$ and gradually increase it during training. The first version leads to a simpler training procedure and was hence used in our experiments. Thanks to the depth first traversal the memory consumption (weights and filter responses) grows linearly with the number of levels if the same upsampling and output architecture is used for each level.

In order to be able to train with mini-batches we sum up all the gradients during traversal of the tree and only do a gradient step as soon as we have done a forward and backward traversal of the subsampled tree for the training examples of the whole mini-batch. As loss function we use Cross-Entropy.

\section{Experiments}
\label{sec:exp}

For our evaluation we use the synthetic dataset ShapeNetCore \cite{shapenet2015}, which is commonly used to evaluate gemetry prediction networks. It is composed of Computer Aided Design (CAD) models of objects which are organized in categories. We use three categories for our evaluation, aeroplanes, chairs and cars. Our system is implemented in Torch\footnote{http://torch.ch/} and we use Adam \cite{kingma2015adam} for stochastic optimization, with a mini-batch size of 4.

\subsection{Voxelization of the Ground Truth}
\label{sec:voxelization}
In a prepossessing step we voxelize the ground truth data. A common approach is to consider all the voxels which intersect the ground truth mesh as occupied space and then fill in the interior by labeling all voxels which are not reachable through free space from the boundary as occupied space. While this works well for the commonly used $32^3$ resolution it does not directly generalize to high resolutions. The reason is that the CAD models are generally not watertight meshes and with increasing resolution the risk that a hole in the mesh prevents filling the interior increases. In our case we are voxelizing the models at a resolution of $256^3$. In order to have filled interiors of the objects at high resolution we utilize a multi-scale approach. We first fill the interior at $32^3$ resolution. Then erode the occupied space by one voxel. We then fix every high resolution voxel which falls within this eroded low resolution occupied space as occupied space and also fix every high resolution voxel which falls within the original low resolution free space as free space. Furthermore, we fix all the high resolution voxels which intersect the original mesh surface as occupied space. To determine the ground truth label for the remaining high resolution voxels we run a graph-cut based regularization with a small smoothness term and a preference to keep the unlabeled voxels as free space.

Given the high resolution voxels at $256^3$ resolution we can now build the ground truth for all the remaining levels of the pyramid. At each level of the pyramid we label the voxels which contain the boundary at highest resolution as boundary and the other labels either free or occupied space, our coarsest resolution is $16^3$.

\subsection{Baselines}
We consider two baselines. Both of the baselines use the state-of-the-art coarse resolution voxel prediction framework from Sec.\ \ref{sec:voxPred}. For both baselines we use a uniform prediction resolution of $32^3$ (except for one evaluation where we use $64^3$ resolution), they differ in the way the ground truth is computed. The first baseline follows the standard approach of labeling all voxels which intersect the ground truth mesh surface as occupied space and then fill in the interior. This can be achieved by downsampling our high resolution ground truth and label all low resolution voxels which contain at least one high resolution occupied space voxel as occupied space and all the other ones as free space. We call this baseline ``Low Resolution Hard" (LR Hard). The other baseline uses a soft assignment for the low resolution, the label is given by the ratio of high resolution free to occupied space voxels that the low resolution voxel contains. Therefore these labels can have fractional assignments. We call this baseline ``Low Resolution Soft" (LR Soft). Note that the baseline LR Soft makes use of the high resolution voxelization but the baseline LR Hard is equivalent to voxelizing at low resolution. As our goal is prediction of high resolution geometry we trininearly upsample the raw classification output of the baselines from $32^3$ to $256^3$ and conduct the evaluation at high resolution.

\begin{figure}
\centering
\includegraphics[width=0.9\linewidth]{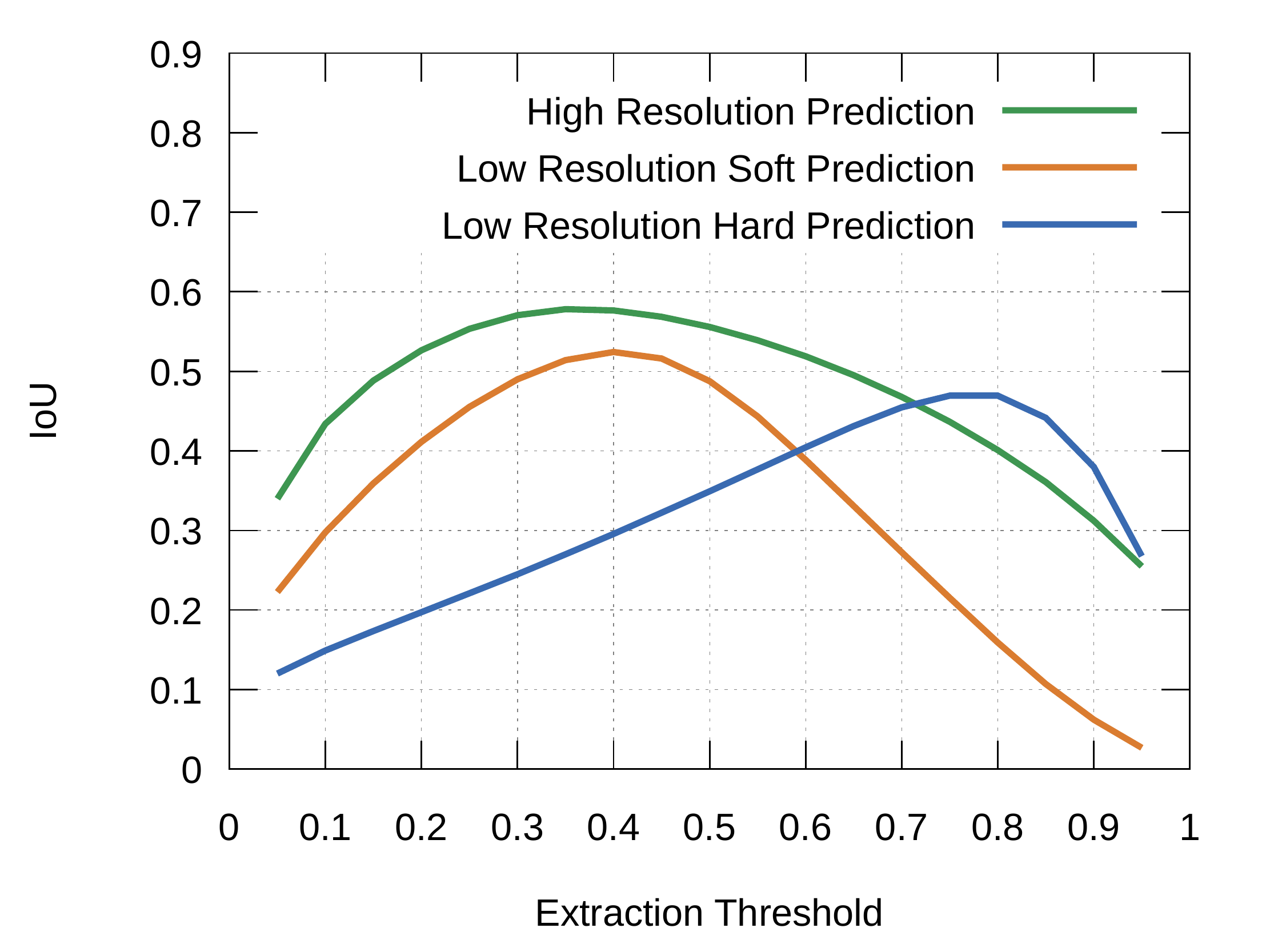} \\
\includegraphics[width=0.9\linewidth]{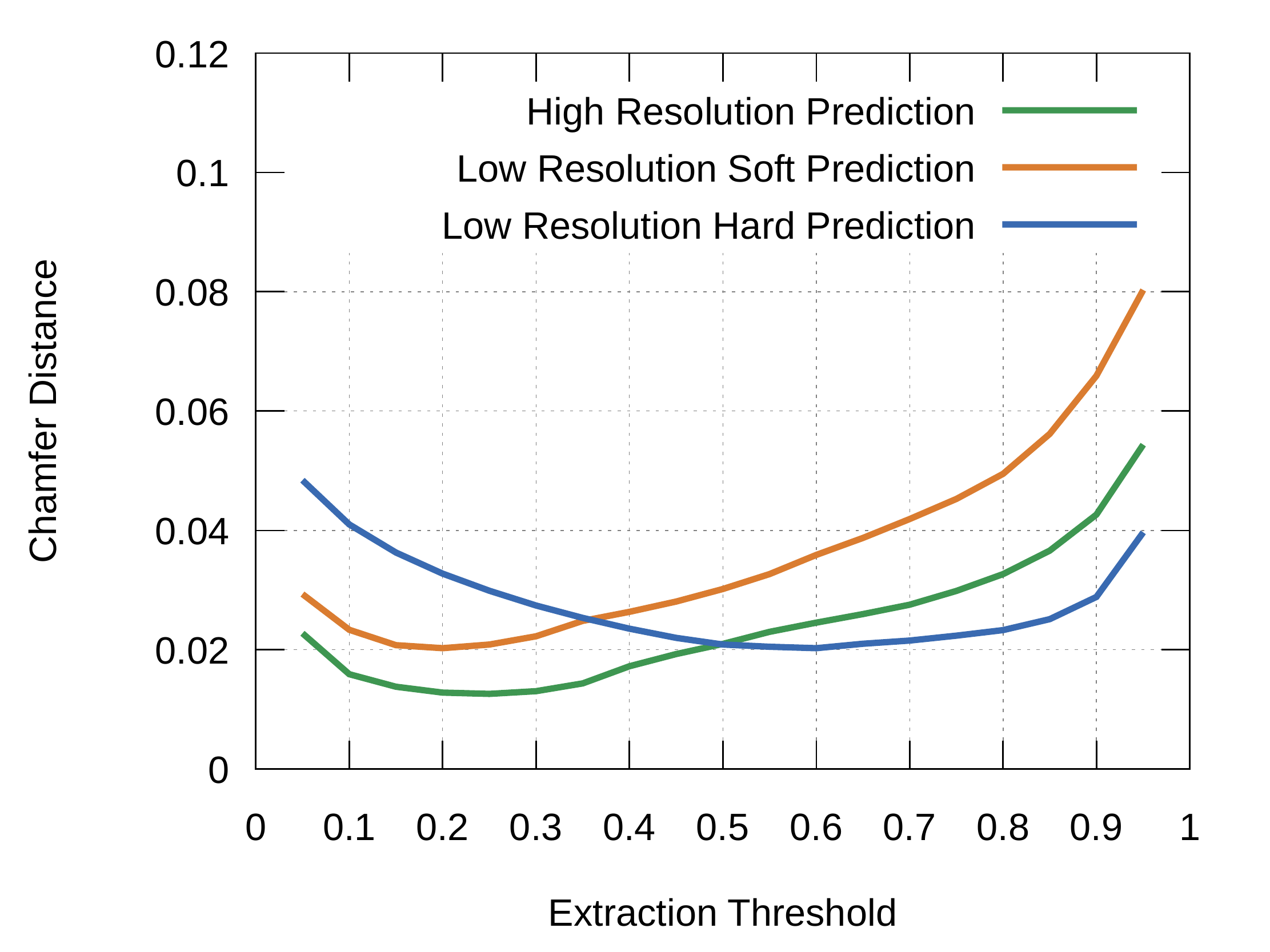}
\caption{Evaluation on the validation set at different extraction thresholds for the class aeroplane at a resolution of $256^3$. For IoU higher is better and for CD lower is better.}
\label{fig:theshEvalAero}
\end{figure}

\renewcommand{\arraystretch}{1.2}
\setlength{\tabcolsep}{10pt}
\begin{table}[]
    \centering
    \small
    \begin{tabular}{lcccc}
        \toprule
        Metric & Method & Car & Chair & Aero \\
         \cmidrule(lr){1-1}
         \cmidrule(lr){2-2}
         \cmidrule(lr){3-5}
        \multirow{3}{*}{IoU} & LR Hard & 0.649 & 0.368 & 0.458 \\
        & LR Soft & 0.679 & 0.360 & 0.518 \\
        & HSP & \textbf{0.701} &\textbf{0.378} & \textbf{0.561} \\
        \midrule
        \multirow{3}{*}{CD} & LR Hard & 0.0142 & 0.0275 & 0.0190 \\
        & LR Soft & 0.0122 & 0.0321& 0.0190  \\
        & HSP & \textbf{0.0108} & \textbf{0.0266} & \textbf{0.0121}  \\
        \bottomrule
    \end{tabular}
    \caption{Quantitative analysis for the task of predicting geometry from 
    RGB input.}
    \label{tab:rgbPredictionQuant}
\end{table}

\begin{figure*}[h]
 \includegraphics[width=0.115\linewidth, trim={0.5cm 0.1cm 0.5cm 0.1cm}, clip]{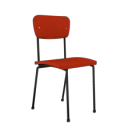}
 \includegraphics[width=0.115\linewidth, trim={4cm 2.5cm 4cm 2.5cm}, clip]{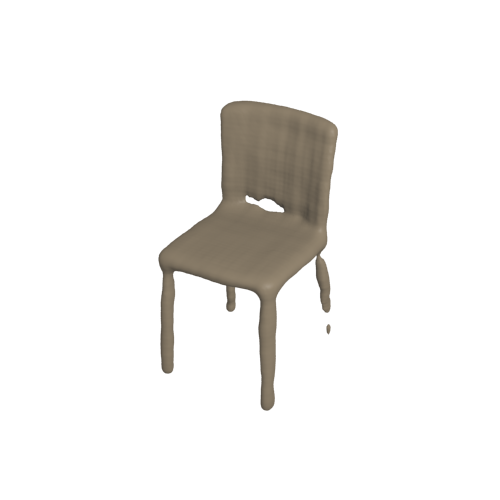}
 \includegraphics[width=0.115\linewidth, trim={4cm 2.5cm 4cm 2.5cm}, clip]{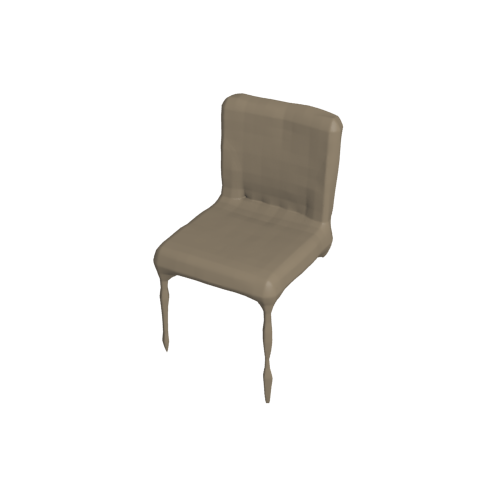}
 \includegraphics[width=0.115\linewidth, trim={4cm 2.5cm 4cm 2.5cm}, clip]{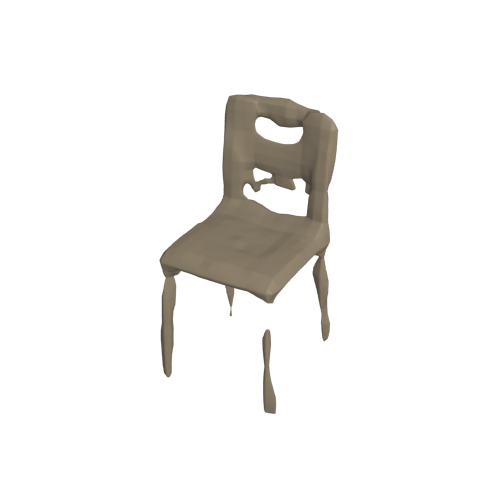} \hfill
 \includegraphics[width=0.115\linewidth, trim={0.5cm 0.1cm 0.5cm 0.1cm}, clip]{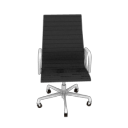}
 \includegraphics[width=0.115\linewidth, trim={4cm 2.5cm 4cm 2.5cm}, clip]{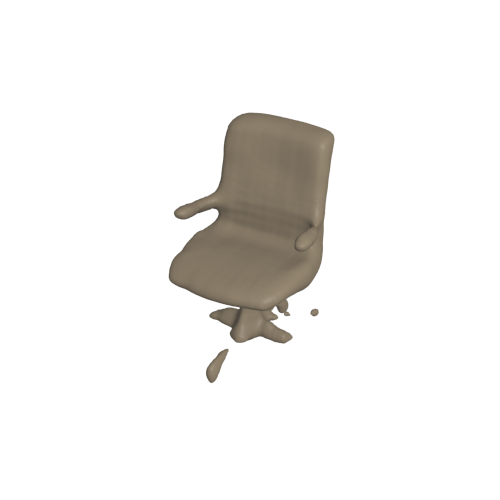}
 \includegraphics[width=0.115\linewidth, trim={4cm 2.5cm 4cm 2.5cm}, clip]{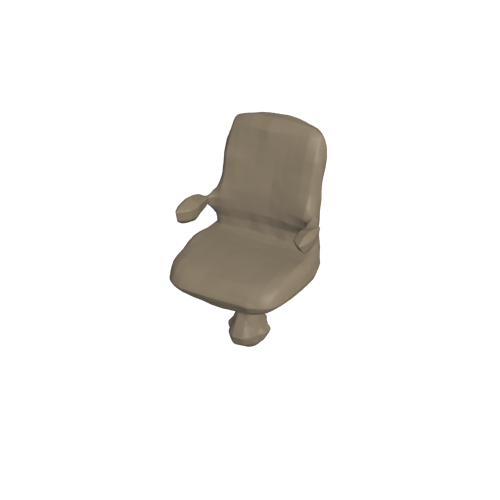}
 \includegraphics[width=0.115\linewidth, trim={4cm 2.5cm 4cm 2.5cm}, clip]{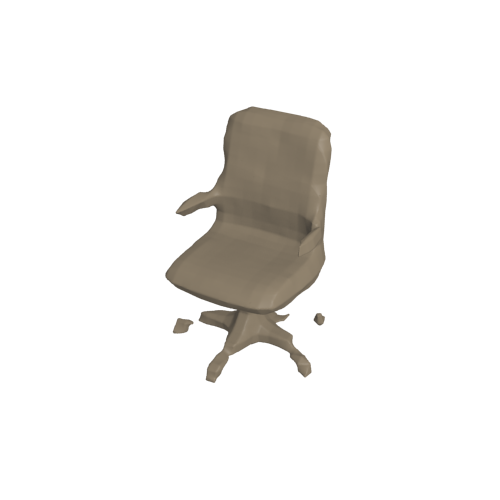} \\
 \includegraphics[width=0.115\linewidth, trim={0.5cm 0.1cm 0.5cm 0.1cm}, clip]{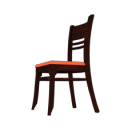}
 \includegraphics[width=0.115\linewidth, trim={4cm 2.5cm 4cm 2.5cm}, clip]{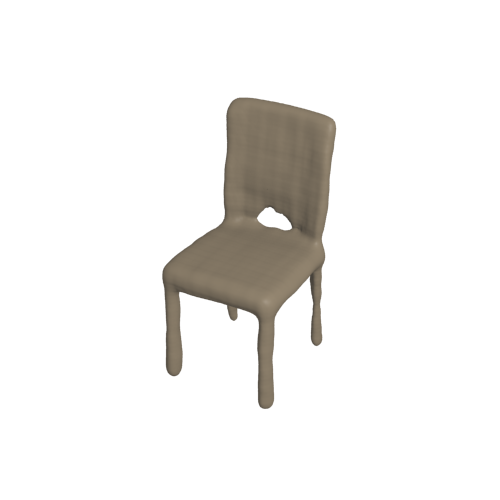}
 \includegraphics[width=0.115\linewidth, trim={4cm 2.5cm 4cm 2.5cm}, clip]{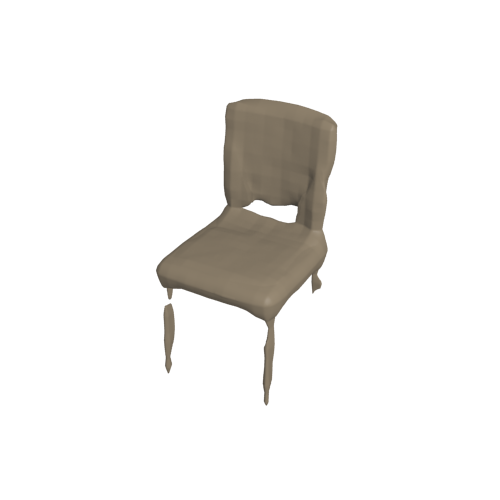}
 \includegraphics[width=0.115\linewidth, trim={4cm 2.5cm 4cm 2.5cm}, clip]{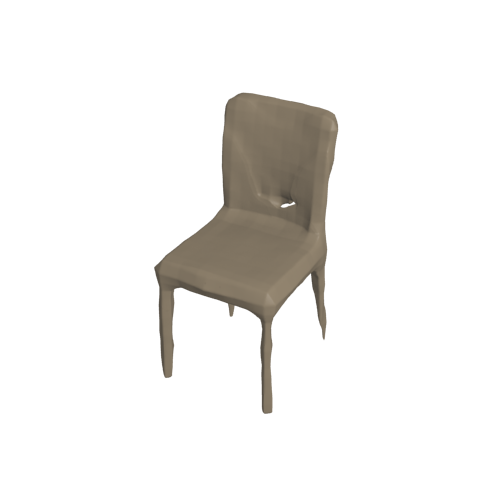} \hfill
 \includegraphics[width=0.115\linewidth, trim={0.5cm 0.1cm 0.5cm 0.1cm}, clip]{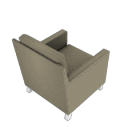}
 \includegraphics[width=0.115\linewidth, trim={4cm 2.5cm 4cm 2.5cm}, clip]{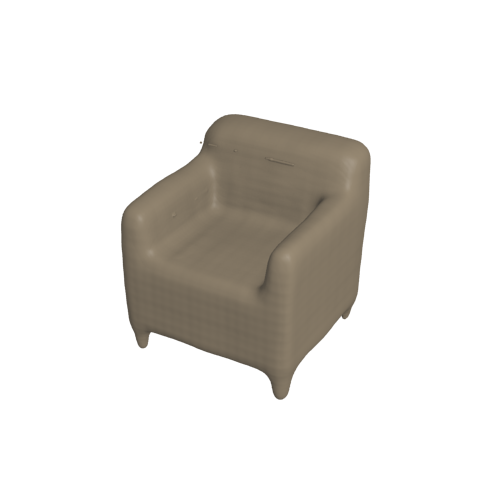}
 \includegraphics[width=0.115\linewidth, trim={4cm 2.5cm 4cm 2.5cm}, clip]{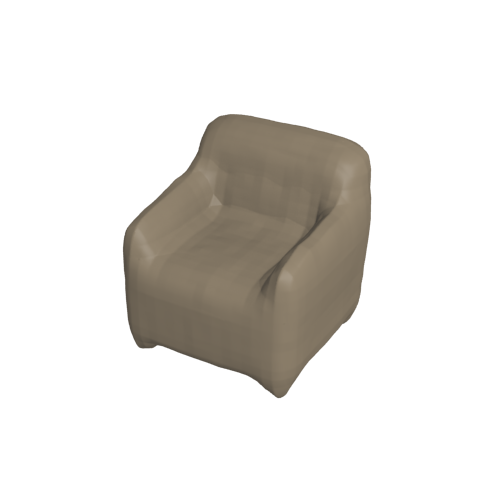}
 \includegraphics[width=0.115\linewidth, trim={4cm 2.5cm 4cm 2.5cm}, clip]{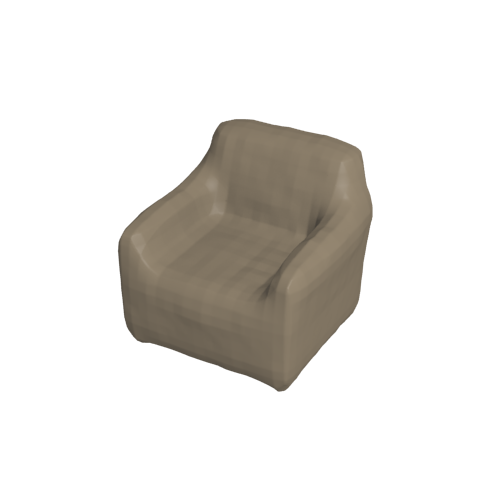}
 \caption{Results for the category chairs. Each block from left to right, input image, our proposed HSP, LR Soft, LR Hard.}
 \label{fig:chairQualitative}
\end{figure*}

\begin{figure*}[h]
 \centering
 \includegraphics[width=0.20\linewidth, trim={0.4cm 1.1cm 0.4cm 1.1cm}, clip]{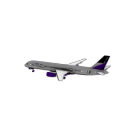}
 \includegraphics[width=0.20\linewidth, trim={4cm 5.5cm 4cm 5.5cm}, clip]{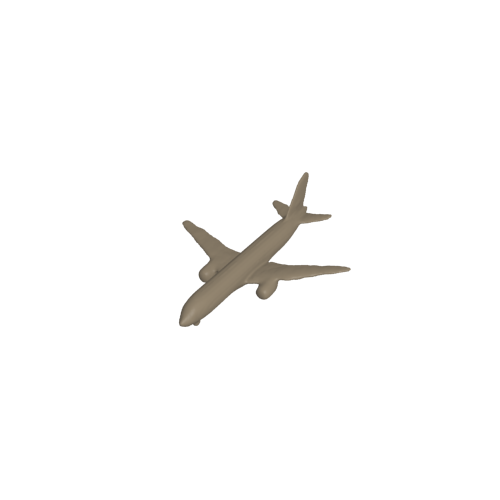}
 \includegraphics[width=0.20\linewidth, trim={4cm 5.5cm 4cm 5.5cm}, clip]{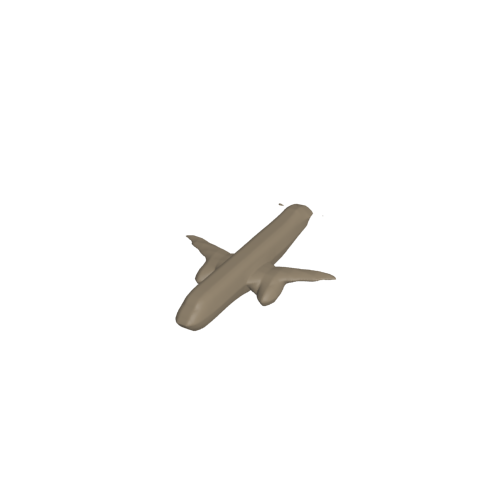}
 \includegraphics[width=0.20\linewidth, trim={4cm 5.5cm 4cm 5.5cm}, clip]{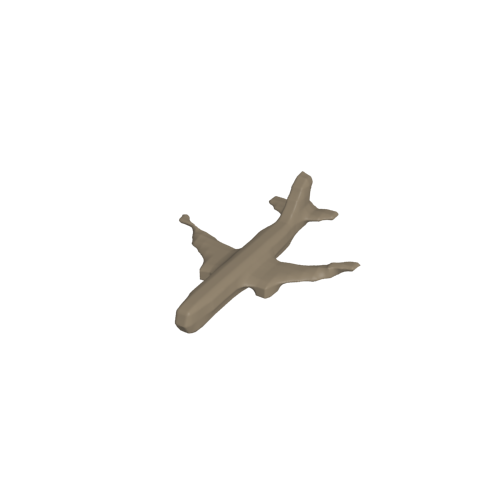} \vspace{-0.5cm} \\
 \includegraphics[width=0.20\linewidth, trim={0.4cm 1.1cm 0.4cm 1.1cm}, clip]{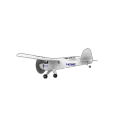}
 \includegraphics[width=0.20\linewidth, trim={4cm 5.5cm 4cm 5.5cm}, clip]{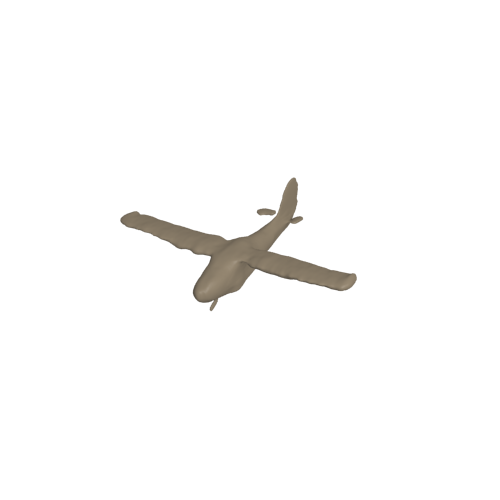}
 \includegraphics[width=0.20\linewidth, trim={4cm 5.5cm 4cm 5.5cm}, clip]{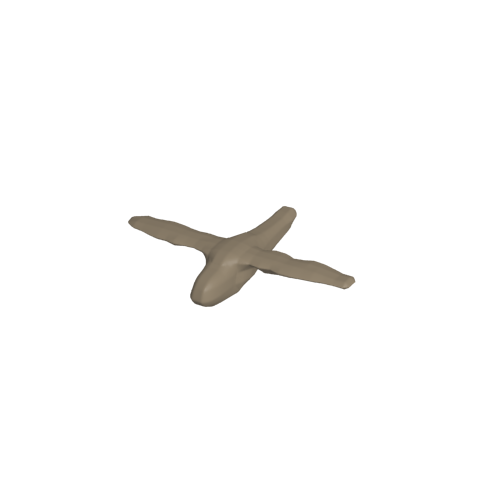}
 \includegraphics[width=0.20\linewidth, trim={4cm 5.5cm 4cm 5.5cm}, clip]{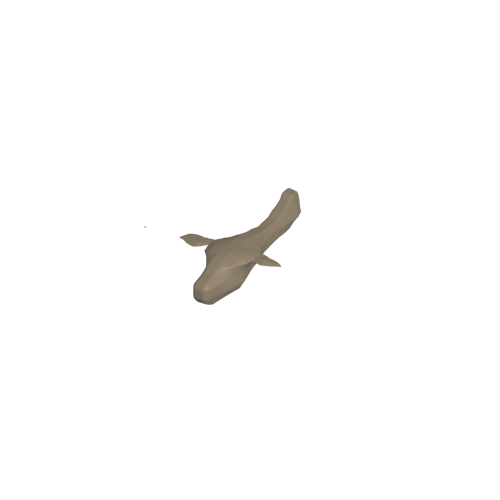}   \vspace{-0.5cm}  \\
 \includegraphics[width=0.20\linewidth, trim={0.4cm 1.1cm 0.4cm 1.1cm}, clip]{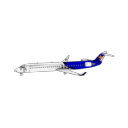}
 \includegraphics[width=0.20\linewidth, trim={4cm 5.5cm 4cm 5.5cm}, clip]{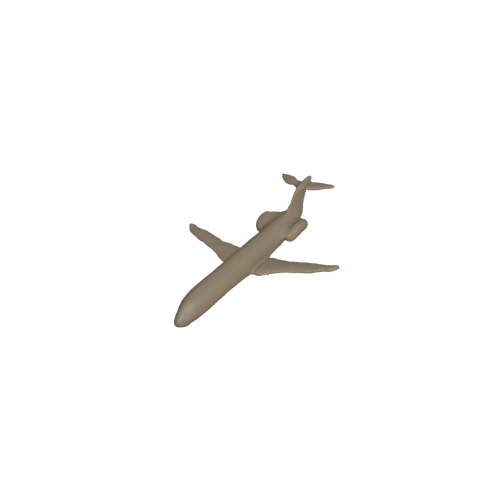}
 \includegraphics[width=0.20\linewidth, trim={4cm 5.5cm 4cm 5.5cm}, clip]{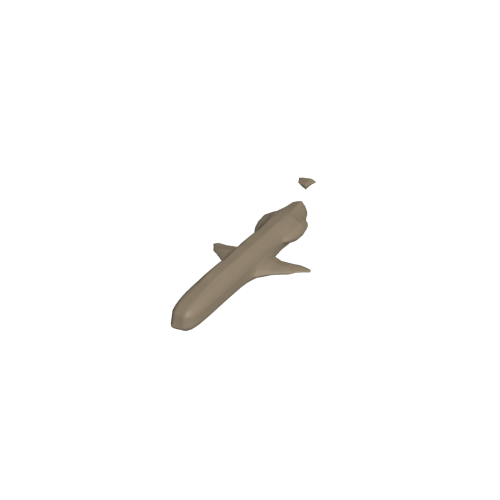}
 \includegraphics[width=0.20\linewidth, trim={4cm 5.5cm 4cm 5.5cm}, clip]{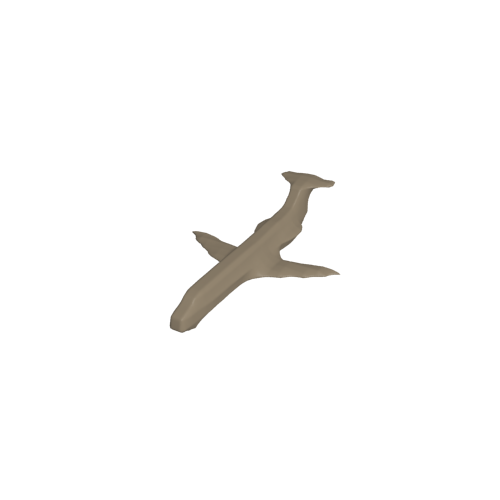}   \vspace{-0.5cm}  \\
 \includegraphics[width=0.20\linewidth, trim={0.4cm 1.1cm 0.4cm 1.1cm}, clip]{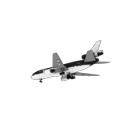}
 \includegraphics[width=0.20\linewidth, trim={4cm 5.5cm 4cm 5.5cm}, clip]{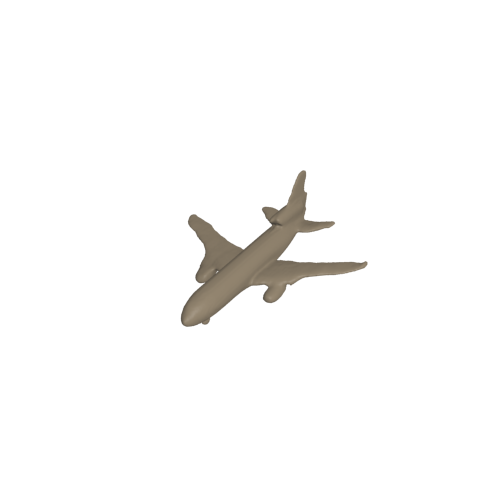}
 \includegraphics[width=0.20\linewidth, trim={4cm 5.5cm 4cm 5.5cm}, clip]{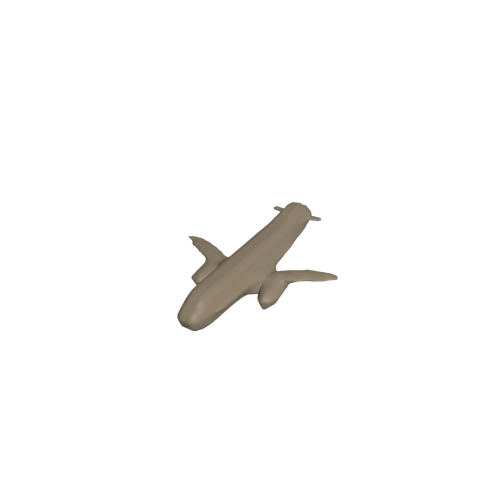}
 \includegraphics[width=0.20\linewidth, trim={4cm 5.5cm 4cm 5.5cm}, clip]{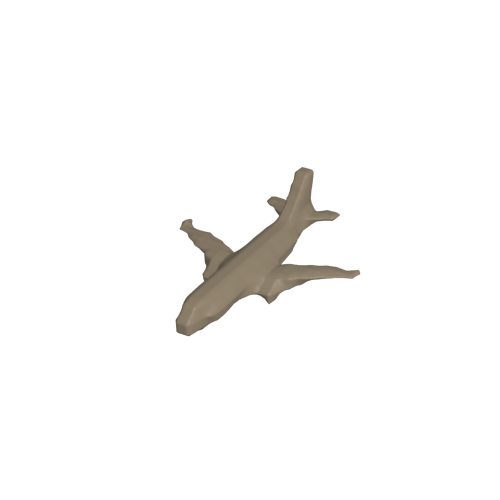}  \\
 \caption{Results for the category aeroplanes. From left to right, input image, our proposed HSP, LR Soft, LR Hard.}
 \label{fig:aeroQualitative}
\end{figure*}

\begin{figure*}[h]
 \includegraphics[width=0.115\linewidth,  trim={0.325cm 0.6cm 0.325cm 0.6cm}, clip]{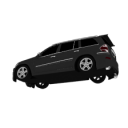}
 \includegraphics[width=0.115\linewidth,  trim={4cm 4.5cm 4cm 4.5cm}, clip]{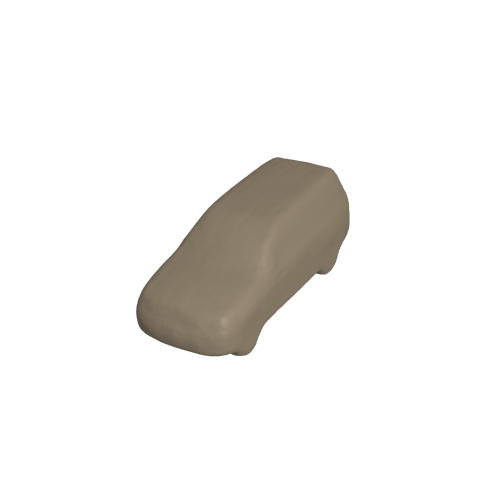}
 \includegraphics[width=0.115\linewidth,  trim={4cm 4.5cm 4cm 4.5cm}, clip]{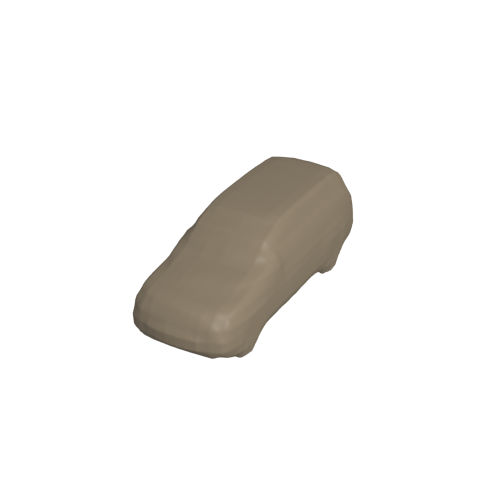}
 \includegraphics[width=0.115\linewidth,  trim={4cm 4.5cm 4cm 4.5cm}, clip]{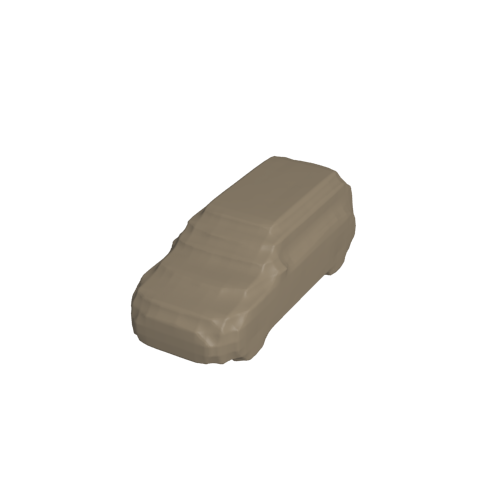} \hfill
 \includegraphics[width=0.115\linewidth,  trim={0.325cm 0.6cm 0.325cm 0.6cm}, clip]{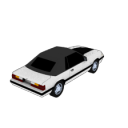} 
 \includegraphics[width=0.115\linewidth,  trim={4cm 4.5cm 4cm 4.5cm}, clip]{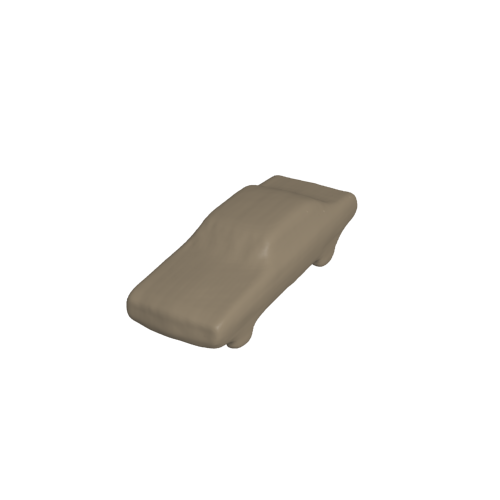}
 \includegraphics[width=0.115\linewidth,  trim={4cm 4.5cm 4cm 4.5cm}, clip]{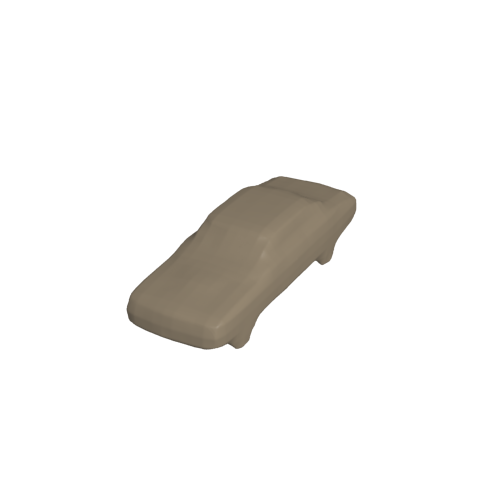}
 \includegraphics[width=0.115\linewidth,  trim={4cm 4.5cm 4cm 4.5cm}, clip]{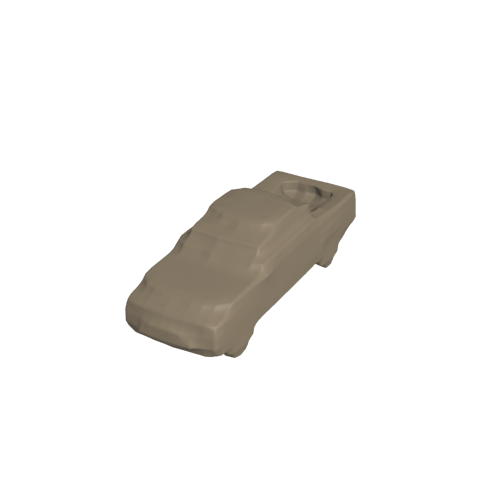} \\
  \includegraphics[width=0.115\linewidth,  trim={0.325cm 0.6cm 0.325cm 0.6cm}, clip]{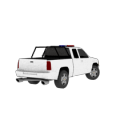}
 \includegraphics[width=0.115\linewidth,  trim={4cm 4.5cm 4cm 4.5cm}, clip]{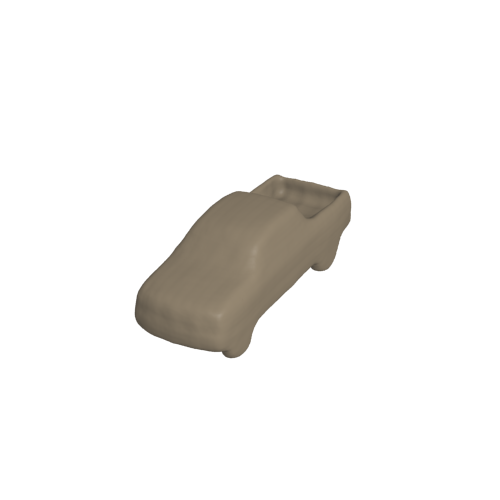}
 \includegraphics[width=0.115\linewidth,  trim={4cm 4.5cm 4cm 4.5cm}, clip]{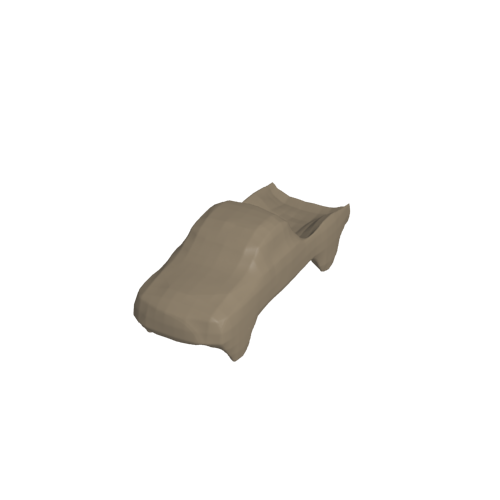}
 \includegraphics[width=0.115\linewidth,  trim={4cm 4.5cm 4cm 4.5cm}, clip]{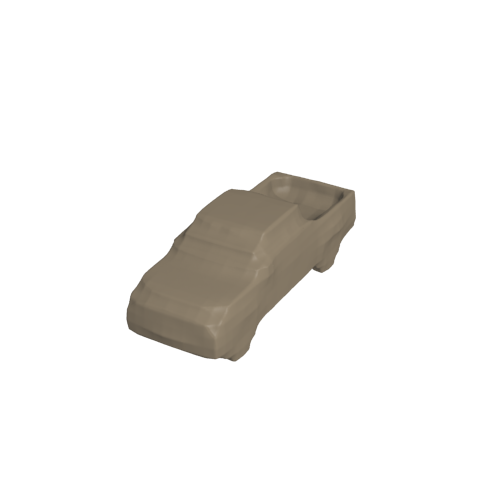} \hfill
 \includegraphics[width=0.115\linewidth,  trim={0.325cm 0.6cm 0.325cm 0.6cm}, clip]{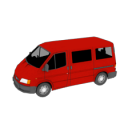} 
 \includegraphics[width=0.115\linewidth,  trim={4cm 4.5cm 4cm 4.5cm}, clip]{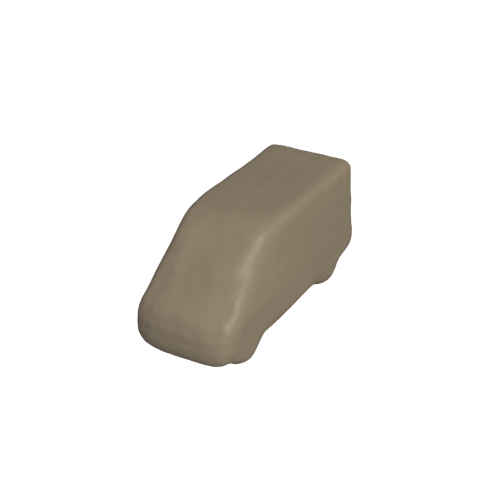}
 \includegraphics[width=0.115\linewidth,  trim={4cm 4.5cm 4cm 4.5cm}, clip]{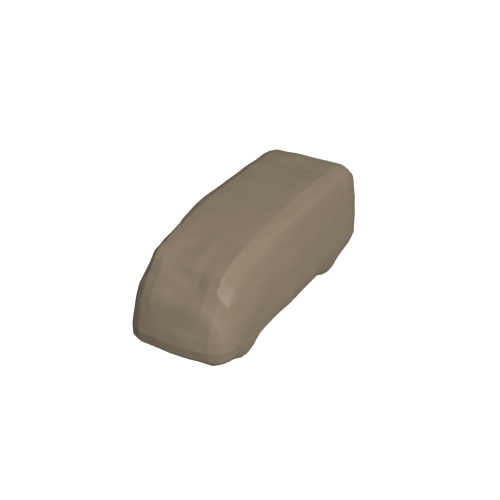}
 \includegraphics[width=0.115\linewidth,  trim={4cm 4.5cm 4cm 4.5cm}, clip]{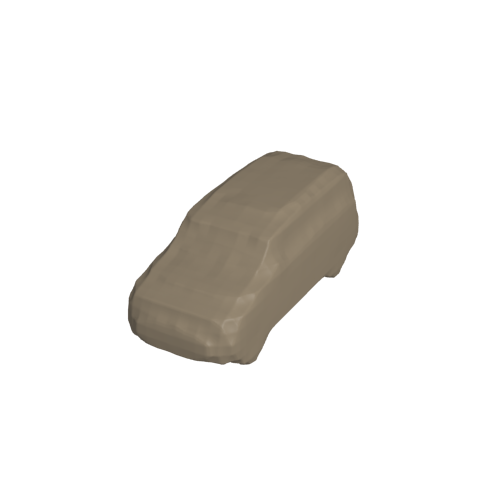} 
 \caption{Results for the category cars. Each block from left to right, input image, our proposed HSP, LR Soft, LR Hard.}
 \label{fig:carQualitative}
\end{figure*}

\begin{figure*}
 \includegraphics[width=0.115\linewidth,  trim={0.325cm 0.6cm 0.325cm 0.6cm}, clip]{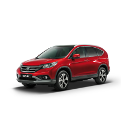}
 \includegraphics[width=0.115\linewidth,  trim={4cm 4.5cm 4cm 4.5cm}, clip]{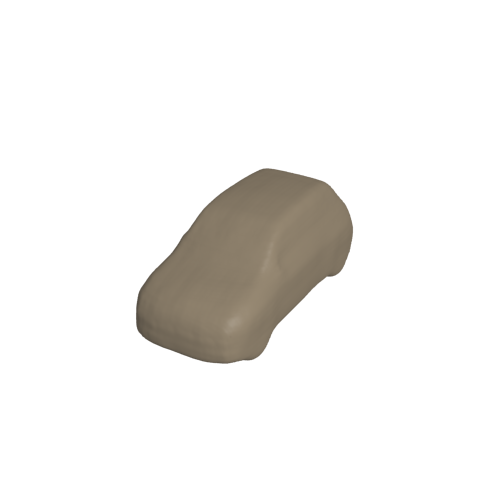}
 \includegraphics[width=0.115\linewidth,  trim={0.325cm 0.6cm 0.325cm 0.6cm}, clip]{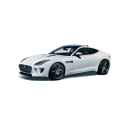}
 \includegraphics[width=0.115\linewidth,  trim={4cm 4.5cm 4cm 4.5cm}, clip]{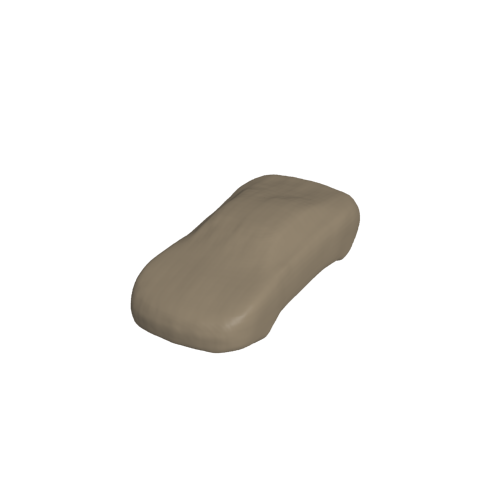} \hfill
 \includegraphics[width=0.115\linewidth]{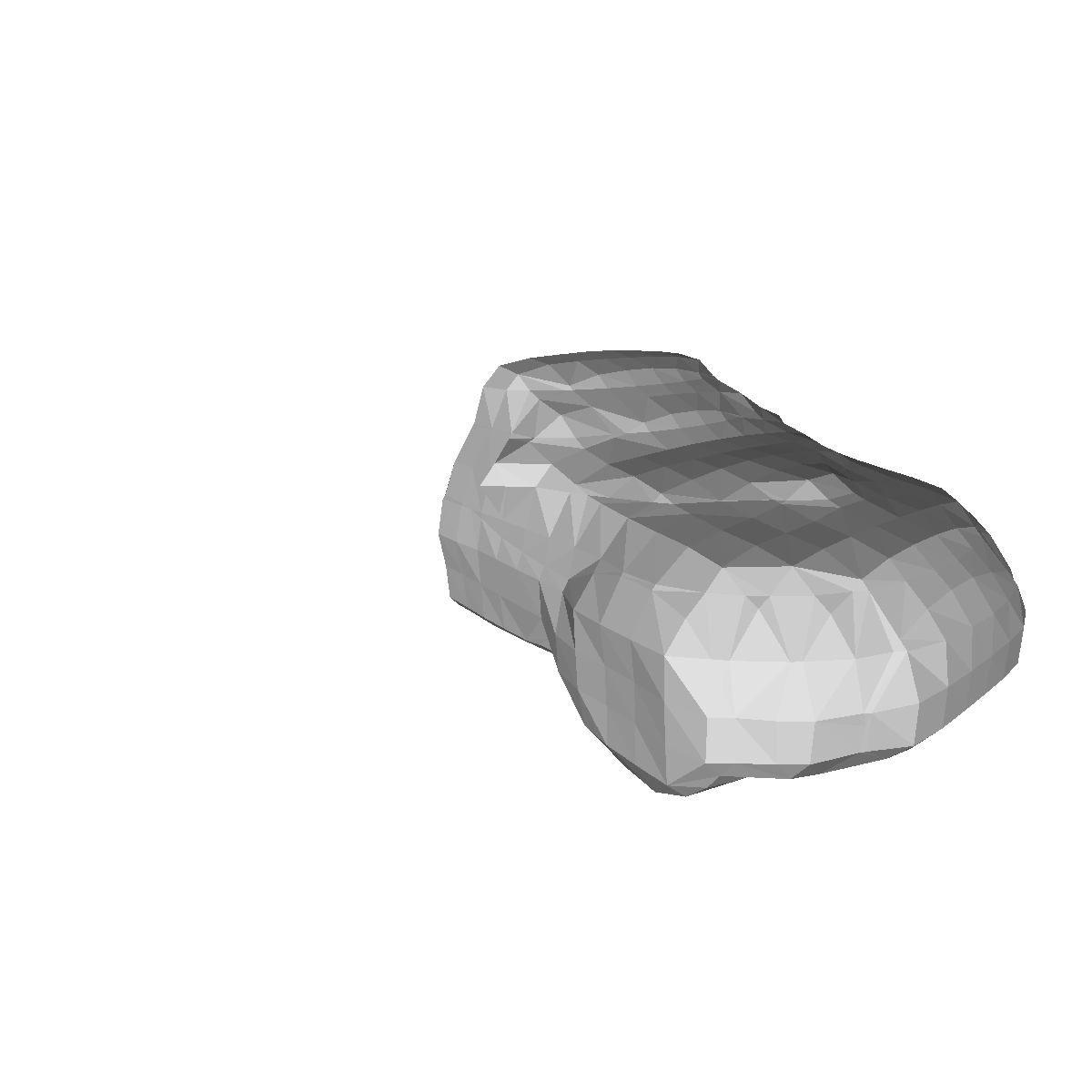}
 \includegraphics[width=0.115\linewidth]{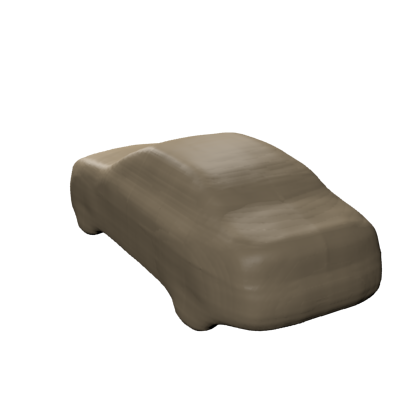}
 \includegraphics[width=0.115\linewidth]{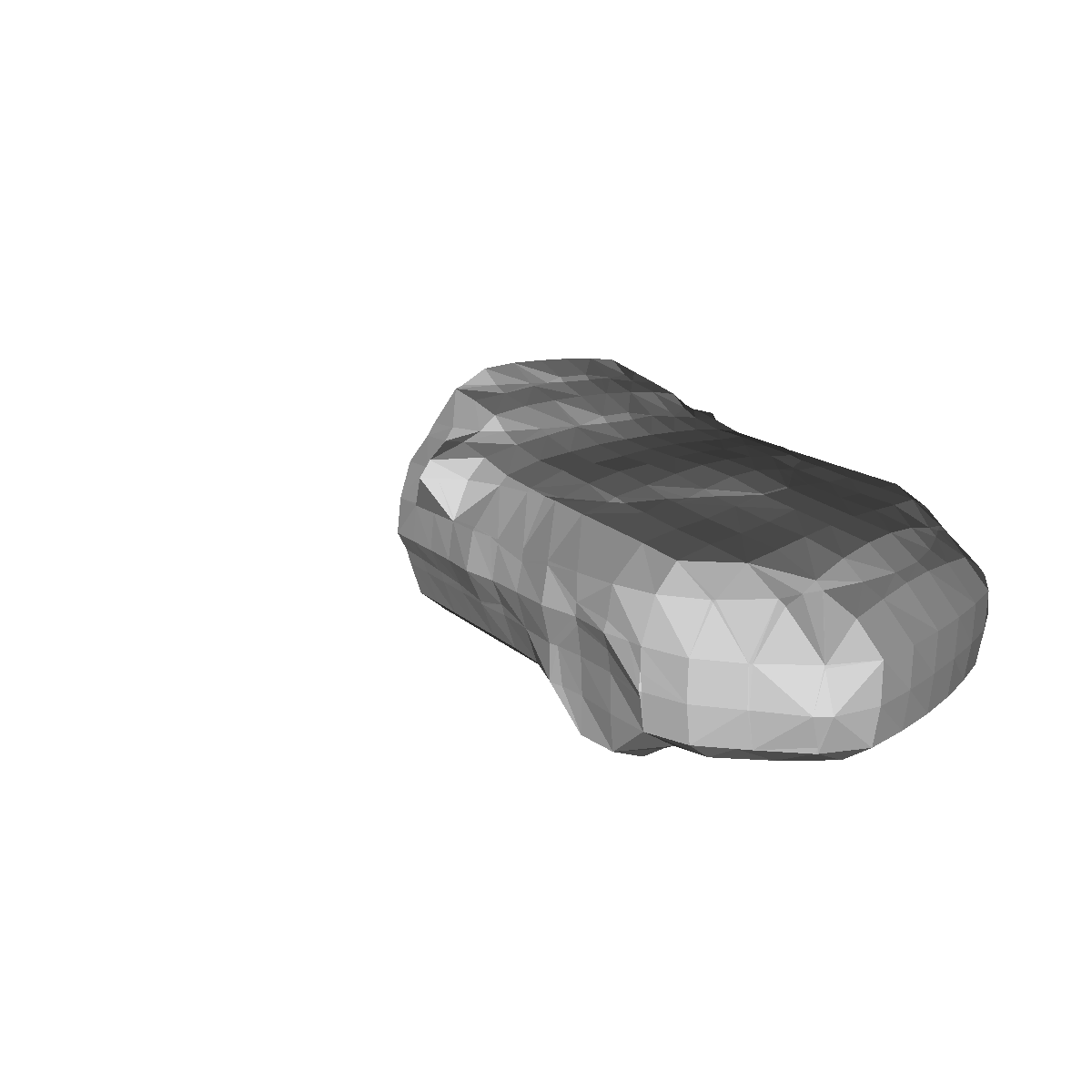}
 \includegraphics[width=0.115\linewidth]{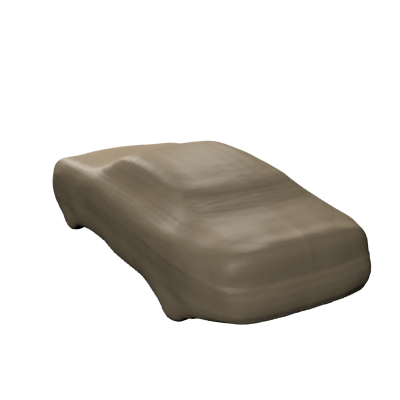}
 \caption{Additional applications of our method, (left) reconstruction from real world rgb images, (right) geometry prediction from partial low resolution input volumes. \vspace{-0.1cm}}
 \label{fig:addRes}
\end{figure*}

\subsection{Input Data}
To obtain the RGB/Depth images which are used as input for training and testing our CNNs, we render the CAD models in the ShapeNet dataset using Blender\footnote{http://www.blender.org}. For each CAD model, we render 10 images from random viewpoints obtained via uniformly sampling azimuth from $[0,360)$ degrees and elevation from $[-20,30]$ degrees. We also use random lighting variations to render the RGB images.

We also use partial voxel grids as input. In practice such an input can arise when multiple depth maps of an object are available but they are all from the same side of the object. To imitate this task we use the ground truth of the LR Soft baseline and randomly zero the data in half of the voxel grid. The network then learns to predict the complete high resolution geometry from the partial low resolution input.

\subsection{Quantitative Evaluation}
\label{sec:quantEval}

We conducted a quantitative evaluation on the task of predicting high resolution geometry from a single RGB image. We assign each of the 3D models from the dataset randomly to the train, validation or test split with probabilities 70\%, 10\% and 20\%, respectively. We trained category specific networks for our method Hierarchical Surface Prediction (HSP) and both baselines for the three categories, aeroplanes, chairs and cars. To quantify the accuracy we use use the Intersection over Union (IoU) score and the symmetric Chamfer Distance (CD). We compute the measures for each model and average over the dataset. To compute the measures we binarize the non-binary predictions with a suitable threshold. In order to determine the best threshold for each measure and category we test an exhaustive set of thresholds using the validation set. The plots given in Fig.\ \ref{fig:theshEvalAero} show that the choice of a good threshold has a very drastic influence to the error measure. As a first experiment, we study predictions at a coarse resolution of $64^3$. For this experiment both the baseline and HSP are trained to predict $64^3$ resolution and the evaluation is conducted at the same resolution. We demonstrate that our hierarchical prediction performs similar to the LR hard baseline \ie the performance of HSP does not degrade compared to uniform prediction at the same resolution. For the baseline LR Hard we get an IoU of 43.47\%  and 43.12\% for HSP, respectively on the validation set for the class chair. However, unlike HSP, the uniform prediction baselines cannot scale to finer resolutions and we now empirically show the benefits of being able to predict at finer resolutions.
The evaluation is conducted  at $256^3$ resolution. We compute the IoU and Chamfer distance for each of the models in the test set using the per category and per error measure best threshold determined using the validation set. The results are given in Table \ref{tab:rgbPredictionQuant}. We outperform the baselines for all categories and evaluation metrics.

\subsection{Qualitative Evaluation}
In Figs. \ref{fig:chairQualitative},\ref{fig:aeroQualitative} and \ref{fig:carQualitative} we show example results on the task of geometry prediction from a single color image for the three categories. The results are selected to demonstrate the variablility of shapes within a category. Random results are shown in the appendix. The results are predicted with the same networks which we used in the quantitative analysis above. The meshes are extracted with marching cubes \cite{lorensen1987marching} using the same thresholds as for the quantitative analysis using the Chamfer Distance. For chair the LR Hard baseline produces visually pleasant results. However, this same baseline does not achieve detailed results on the categories aeroplane and car. Due to the fractional labels the LR Soft baseline often misses thin structures. The results generated with our proposed HSP have high quality surfaces for all the three categories.

We also downloaded two images of cars with a white background from the Internet using Google image search and processed them with our car prediction network, which was trained only on synthetic data. As can be seen in Fig.\ \ref{fig:addRes}, the network is able to generalize to this type of input. Additionally, we trained a network on the category car which takes a partial low resolution voxel grid as input and predicts a high resolution model of the whole object (c.f.\ Fig.\ \ref{fig:addRes}).

\section{Conclusion and Future Work}
\label{sec:conclusion}

In this work, we presented a hierarcical surface prediction framework which facilitates high resolution 3D reconstruction of objects from very little input data. We presented results on the task of predicting geometry from a single color image, a single depth image and from a partial low resolution voxel grid. Our quantitative evaluation shows that the high resolution prediction leads to more accurate results than low resolution baselines. Qualitatively, our predicted surfaces are superior to the low resolution baselines.

In this work we have demonstrated that high resolution geometry prediction is feasible therefore future work will investigate the potential to use prediction based reconstruction methods in multi-view stereo. Furthermore, for almost symmetric objects methods which explicitly take the symmetry into account are an interesting further direction which might improve the quality of prediction based approaches.

\section*{Acknowledgments}
C. H{\"a}ne received funding from the “Early Postdoc.Mobility” fellowship  No. 165245  from  the  Swiss  National  Science Foundation. The project received funding from the Intel/NSF VEC award IIS-153909 and NSF award IIS-1212798.

{\small
\bibliographystyle{ieee}
\bibliography{paper_arxiv}
}

\begin{appendices}
 \section{Detailed CNN Architectures}

In this section we give the detailed layer configurations of our networks, including the baselines.  We use the following abbreviations:
\begin{description}[leftmargin=!,labelwidth=\widthof{\bfseries kW, kH, kD}]
    \setlength{\itemsep}{2pt}
    \setlength{\parskip}{2pt}
    \item [Conv] Convolution
    \item [UpConv] Up-convolution
    \item [FC] Fully connected
    \item [MP] Max Pooling 
    \item [LR] Leaky ReLU with negative slope of 0.2
    \item [R] ReLU
    \item [DO] Dropout with probability 0.5
    \item [RS] Reshape
    \item [kW, kH, kD] Kernel sizes in the three dimensions
    \item [sW, sH, sD] Strides in the three dimensions
    \item [oC] Number of output feature channels
    \item [oW, oH, oD] Output sizes in the three dimensions
\end{description}

For both our proposed hierarchical surface prediction and the baselines we use the same encoder. The layers of the encoder for color or depth images are given in Tab.~\ref{tab:encoderRGBDepth}. The encoder for the partial low resolution grids has the same general structure. The decoder for both the baselines is given in Tab.\ \ref{tab:baselineDecoder}. Note that both baselines use the exact same network. In one case the ground truth are hard binary labels and in the other case the ground truth are soft labels which represent the ratio of free/occupied space of the high resolution voxels, \cf main text.

Our proposed hierarchical surface prediction (HSP) uses a set of modules in the decoder. The module which decodes the shape code from the bottle neck to the first feature block is given in Tab.\ \ref{tab:firstDecoder}. As described in the main text each transition from one level to the next is done using an upsampling module Tab.\ \ref{tab:upsampling} and an output module Tab.\ \ref{tab:output}. On the last level the output is generated by the module given in Tab.\ \ref{tab:fullOutput}.

\section{Qualitative Results}

In order to demonstrate the quality of our method we show the reconstruction on randomly selected elements. We show every twentieth element of the validation set of the respective category. Category aeroplanes is depicted in Figs \ref{fig:aero1} and \ref{fig:aero2}, category cars is depicted in Figs \ref{fig:car1}-\ref{fig:car3} and category chairs is depiced in Figs \ref{fig:chair1} and \ref{fig:chair2}.

\begin{table}[t]
\setlength{\tabcolsep}{4pt}
    \centering
    \begin{tabular}{|c|c|c|c|c|c|c|c|c|}
        \hline
        lType & kW & kH & sW & sH & oC & oW & oH\\
        \hline
        \hline
        Input & - & - & - & - & 1/3 & 128 & 128 \\
        \hline
        Conv & 3 & 3 & 1 & 1 & 8 & 128 & 128 \\
        \hline
        MP + LR & 2 & 2 & 2 & 2 & 8 & 64 & 64 \\
        \hline
        Conv & 3 & 3 & 1 & 1 & 16 & 64 & 64 \\
        \hline
        MP + LR & 2 & 2 & 2 & 2 & 16 & 32 & 32 \\
        \hline
        Conv & 3 & 3 & 1 & 1 & 32 & 32 & 32 \\
        \hline
        MP + LR & 2 & 2 & 2 & 2 & 32 & 16 & 16 \\
        \hline
        Conv & 3 & 3 & 1 & 1 & 64 & 16 & 16 \\
        \hline
        MP + LR & 2 & 2 & 2 & 2 & 64 & 8 & 8 \\
        \hline
        Conv & 3 & 3 & 1 & 1 &  128 & 8 & 8 \\
        \hline
        MP + LR & 2 & 2 & 2 & 2 & 128 & 4 & 4 \\
        \hline 
        Conv & 3 & 3 & 1 & 1 & 128 & 4 & 4 \\
        \hline
        MP + LR & 2 & 2 & 2 & 2 & 128 & 2 & 2 \\
        \hline
        Conv & 3 & 3 & 1 & 1 & 128 & 2 & 2 \\
        \hline
        MP + LR & 2 & 2 & 2 & 2 & 128 & 1 & 1 \\
        \hline
        DO & 1 & 1 & 1 & 1 & 128 & 1 & 1 \\
        \hline
        FC + LR & 1 & 1 & 1 & 1 & 128 & 1 & 1 \\
        \hline
        FC + LR & 1 & 1 & 1 & 1 & 128 & 1 & 1 \\
        \hline
    \end{tabular}
    \caption{Color/Depth Encoder}
    \label{tab:encoderRGBDepth}
\end{table}

\begin{table}[t]
\setlength{\tabcolsep}{2.25pt}
    \centering
    \begin{tabular}{|c|c|c|c|c|c|c|c|c|c|c|c|}
    \hline
        lType & kW & kH & kD & sW & sH & sD & oC & oW & oH & oD \\
        \hline
        FC + LR & 1 & 1 & 1 & 1 & 1 & 1 & 1024 & 1 & 1 & 1 \\
        \hline
        RS & - & - & - & - & - & - &  128 & 2 & 2 & 2 \\
        \hline
        Conv + R & 3 & 3 & 3 & 1 & 1 & 1 & 128 & 2 & 2 & 2 \\
        \hline
        UpConv + R & 4 & 4 & 4 & 2 & 2 & 2 & 128 & 4 & 4 & 4 \\
        \hline
        Conv + R & 3 & 3 & 3 & 1 & 1 & 1 & 128 & 4 & 4 & 4 \\
        \hline
        UpConv +  R & 4 & 4 & 4 & 2 & 2 & 2 & 128 & 8 & 8 & 8 \\
        \hline
        Conv + R & 3 & 3 & 3 & 1 & 1 & 1 & 64 & 8 & 8 & 8 \\
        \hline
        UpConv + R & 4 & 4 & 4 & 2 & 2 & 2 & 32 & 16 & 16 & 16 \\
        \hline
        Conv + R & 3 & 3 & 3 & 1 & 1 & 1 & 16 & 16 & 16 & 16 \\
        \hline
        UpConv + R & 4 & 4 & 4 & 2 & 2 & 2 & 8 & 32 & 32 & 32 \\
        \hline
        Conv & 3 & 3 & 3 & 1 & 1 & 1 & 1 & 32 & 32 & 32 \\
        \hline
    \end{tabular}
    \caption{Baseline Decoder}
    \label{tab:baselineDecoder}
\end{table}

\begin{table}[t]
    \setlength{\tabcolsep}{2.25pt}
    \centering
    \begin{tabular}{|c|c|c|c|c|c|c|c|c|c|c|c|}
    \hline
        lType & kW & kH & kD & sW & sH & sD & oC & oW & oH & oD \\
        \hline
        FC + LR & 1 & 1 & 1 & 1 & 1 & 1 & 1024 & 1 & 1 & 1 \\
        \hline
        RS & - & - & - & - & - & - &  128 & 2 & 2 & 2 \\
        \hline
        Conv + R & 3 & 3 & 3 & 1 & 1 & 1 & 128 & 2 & 2 & 2 \\
        \hline
        UpConv + R & 4 & 4 & 4 & 2 & 2 & 2 & 128 & 4 & 4 & 4 \\
        \hline
        Conv + R & 3 & 3 & 3 & 1 & 1 & 1 & 128 & 4 & 4 & 4 \\
        \hline
        UpConv + R & 4 & 4 & 4 & 2 & 2 & 2 & 128 & 8 & 8 & 8 \\
        \hline 
        Conv + R & 3 & 3 & 3 & 1 & 1 & 1 & 64 & 6 & 6 & 6 \\
        \hline
        UpConv + R & 4 & 4 & 4 & 2 & 2 & 2 & 64 & 12 & 12 & 12 \\
        \hline
        Conv + R & 3 & 3 & 3 & 1 & 1 & 1 & 32 & 12 & 12 & 12 \\
        \hline
        UpConv + R & 4 & 4 & 4 & 2 & 2 & 2 & 32 & 22 & 22 & 22 \\
        \hline
        Conv + R & 3 & 3 & 3 & 1 & 1 & 1 & 32 & 20 & 20 & 20 \\
        \hline
    \end{tabular}
    \caption{Decoder module, bottleneck to feature block $\mathcal{F}^{1,1}$}
    \label{tab:firstDecoder}
\end{table}

\begin{table}[t]
   \setlength{\tabcolsep}{2.25pt}
    \centering
    \begin{tabular}{|c|c|c|c|c|c|c|c|c|c|c|c|}
    \hline
        lType & kW & kH & kD & sW & sH & sD & oC & oW & oH & oD \\
        \hline
        UpConv + R & 4 & 4 & 4 & 2 & 2 & 2 & 32 & 22 & 22 & 22 \\
        \hline
        Conv + R & 3 & 3 & 3 & 1 & 1 & 1 & 32 & 20 & 20 & 20 \\
        \hline
    \end{tabular}
    \caption{Upsampling module}
    \label{tab:upsampling}
\end{table}

\begin{table}[t]
   \setlength{\tabcolsep}{2.25pt}
    \centering
    \begin{tabular}{|c|c|c|c|c|c|c|c|c|c|c|c|}
    \hline
        lType & kW & kH & kD & sW & sH & sD & oC & oW & oH & oD \\
        \hline
        Conv + R & 3 & 3 & 3 & 1 & 1 & 1 & 16 & 18 & 18 & 18 \\
        \hline
        Conv + R & 3 & 3 & 3 & 1 & 1 & 1 & 8 & 16 & 16 & 16 \\
        \hline
        Conv + R & 3 & 3 & 3 & 1 & 1 & 1 & 3 & 16 & 16 & 16 \\
        \hline
    \end{tabular}
    \caption{Output module}
    \label{tab:output}
\end{table}

\begin{table}[t]
    \setlength{\tabcolsep}{2.25pt}
    \centering
    \begin{tabular}{|c|c|c|c|c|c|c|c|c|c|c|c|}
    \hline
        lType & kW & kH & kD & sW & sH & sD & oC & oW & oH & oD \\
        \hline
        UpConv + R & 4 & 4 & 4 & 2 & 2 & 2 & 16 & 20 & 20 & 20 \\
        \hline
        Conv + R & 3 & 3 & 3 & 1 & 1 & 1 & 8 & 18 & 18 & 18 \\
        \hline
        Conv + R & 3 & 3 & 3 & 1 & 1 & 1 & 1 & 16 & 16 & 16 \\
        \hline
    \end{tabular}
    \caption{Level $L$ output module}
    \label{tab:fullOutput}
\end{table}

\pgfmathsetmacro{\trimI}{0.35}
\pgfmathsetmacro{\trimR}{3}

\begin{figure*}

    \includegraphics[width=0.115\linewidth, trim={\trimI cm \trimI cm \trimI cm \trimI cm}, clip]{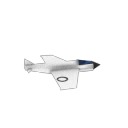}
    \includegraphics[width=0.115\linewidth, trim={\trimR cm \trimR cm \trimR cm \trimR cm}, clip]{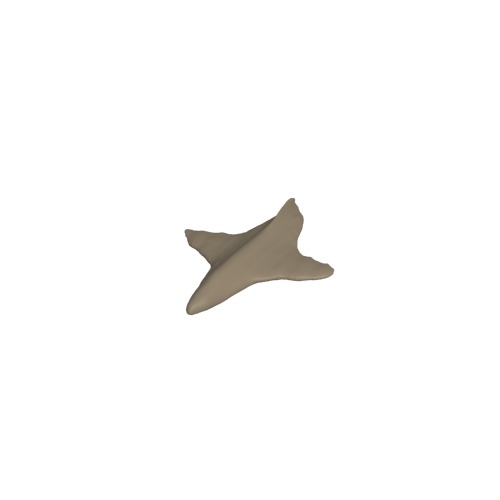}
    \includegraphics[width=0.115\linewidth, trim={\trimR cm \trimR cm \trimR cm \trimR cm}, clip]{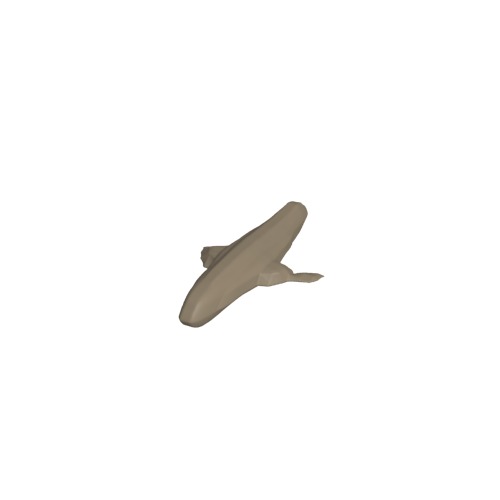}
    \includegraphics[width=0.115\linewidth, trim={\trimR cm \trimR cm \trimR cm \trimR cm}, clip]{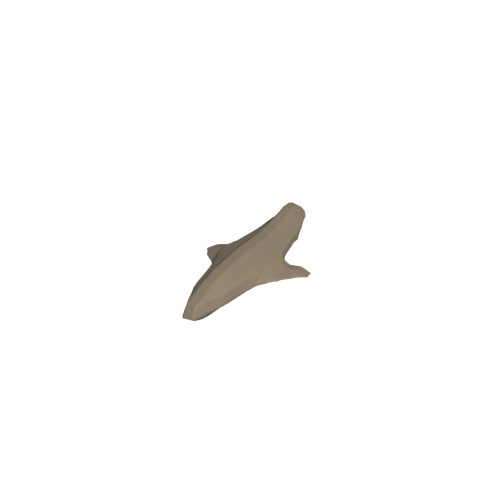}\hfill
    \includegraphics[width=0.115\linewidth, trim={\trimI cm \trimI cm \trimI cm \trimI cm}, clip]{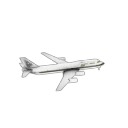}
    \includegraphics[width=0.115\linewidth, trim={\trimR cm \trimR cm \trimR cm \trimR cm}, clip]{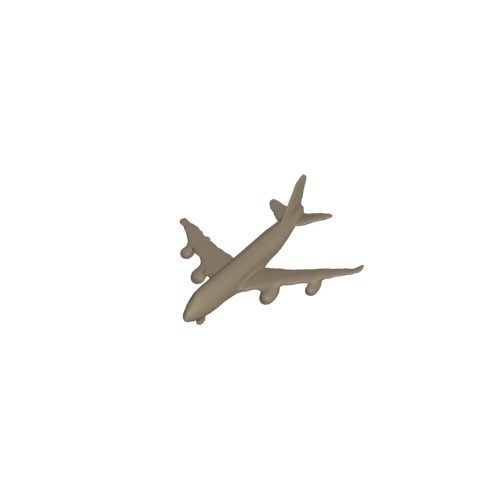}
    \includegraphics[width=0.115\linewidth, trim={\trimR cm \trimR cm \trimR cm \trimR cm}, clip]{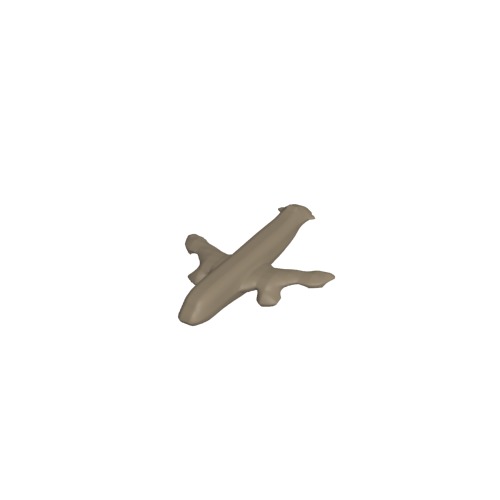}
    \includegraphics[width=0.115\linewidth, trim={\trimR cm \trimR cm \trimR cm \trimR cm}, clip]{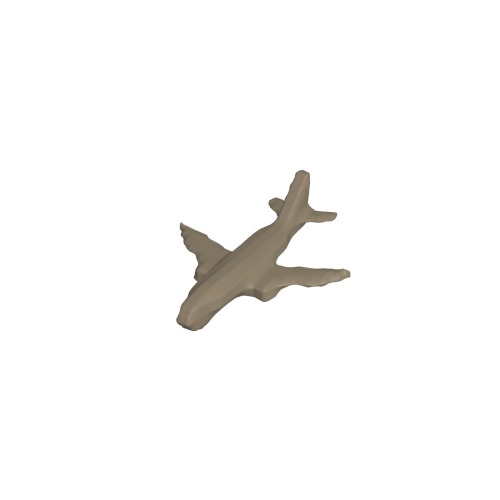}\\
    \includegraphics[width=0.115\linewidth, trim={\trimI cm \trimI cm \trimI cm \trimI cm}, clip]{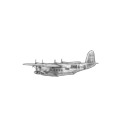}
    \includegraphics[width=0.115\linewidth, trim={\trimR cm \trimR cm \trimR cm \trimR cm}, clip]{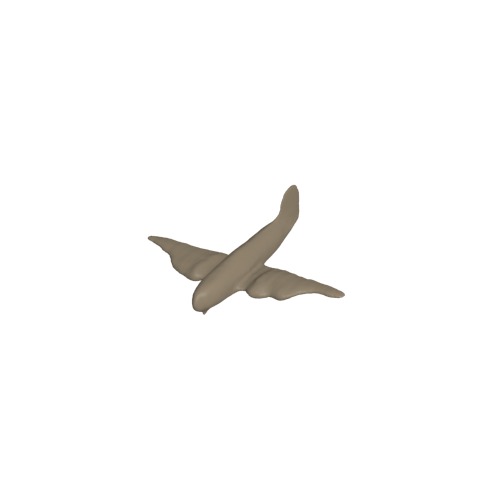}
    \includegraphics[width=0.115\linewidth, trim={\trimR cm \trimR cm \trimR cm \trimR cm}, clip]{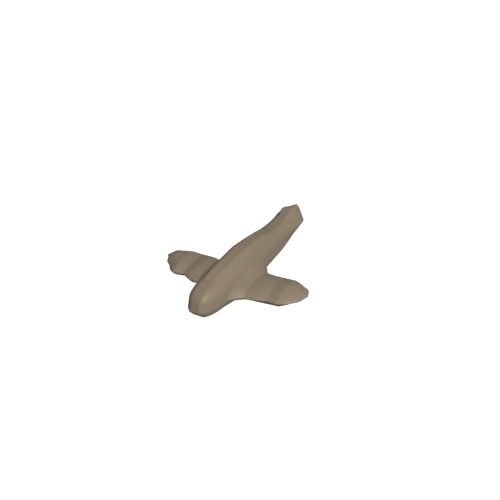}
    \includegraphics[width=0.115\linewidth, trim={\trimR cm \trimR cm \trimR cm \trimR cm}, clip]{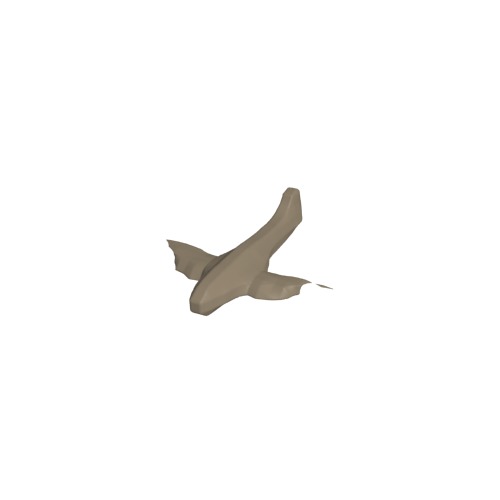}\hfill
    \includegraphics[width=0.115\linewidth, trim={\trimI cm \trimI cm \trimI cm \trimI cm}, clip]{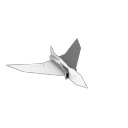}
    \includegraphics[width=0.115\linewidth, trim={\trimR cm \trimR cm \trimR cm \trimR cm}, clip]{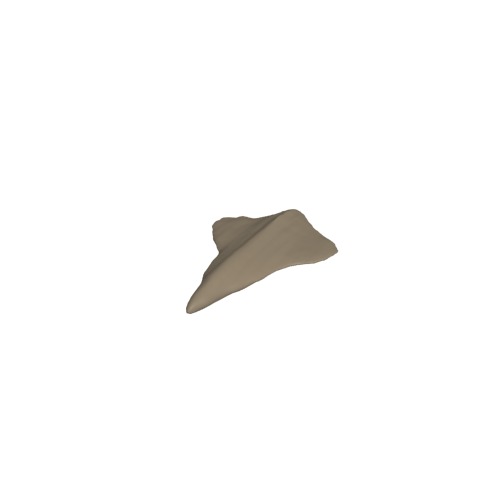}
    \includegraphics[width=0.115\linewidth, trim={\trimR cm \trimR cm \trimR cm \trimR cm}, clip]{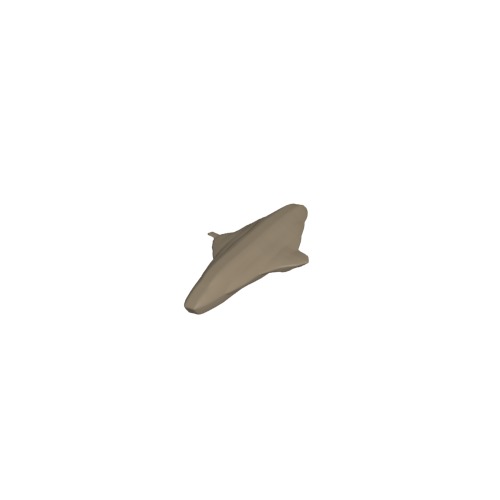}
    \includegraphics[width=0.115\linewidth, trim={\trimR cm \trimR cm \trimR cm \trimR cm}, clip]{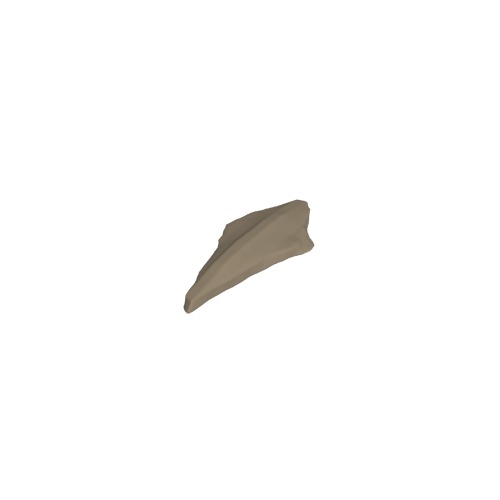}\\
    \includegraphics[width=0.115\linewidth, trim={\trimI cm \trimI cm \trimI cm \trimI cm}, clip]{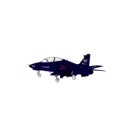}
    \includegraphics[width=0.115\linewidth, trim={\trimR cm \trimR cm \trimR cm \trimR cm}, clip]{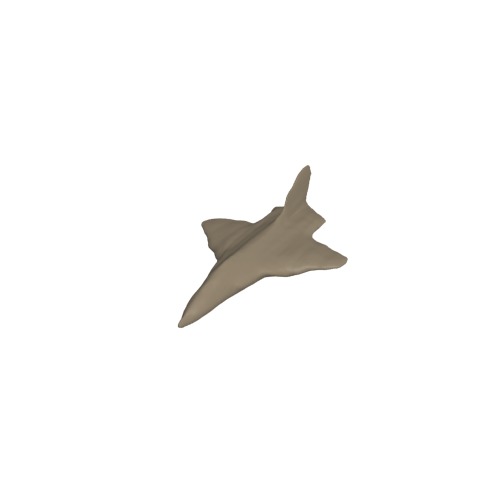}
    \includegraphics[width=0.115\linewidth, trim={\trimR cm \trimR cm \trimR cm \trimR cm}, clip]{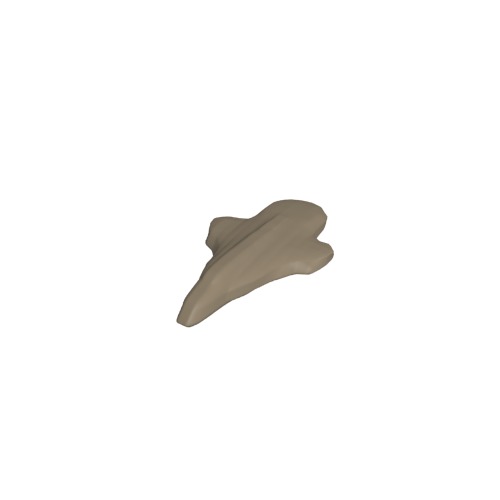}
    \includegraphics[width=0.115\linewidth, trim={\trimR cm \trimR cm \trimR cm \trimR cm}, clip]{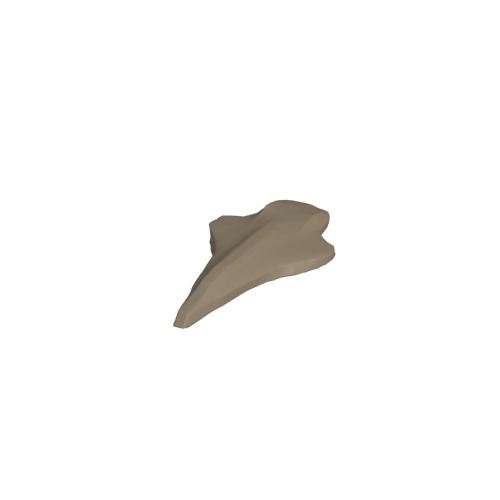}\hfill
    \includegraphics[width=0.115\linewidth, trim={\trimI cm \trimI cm \trimI cm \trimI cm}, clip]{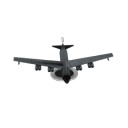}
    \includegraphics[width=0.115\linewidth, trim={\trimR cm \trimR cm \trimR cm \trimR cm}, clip]{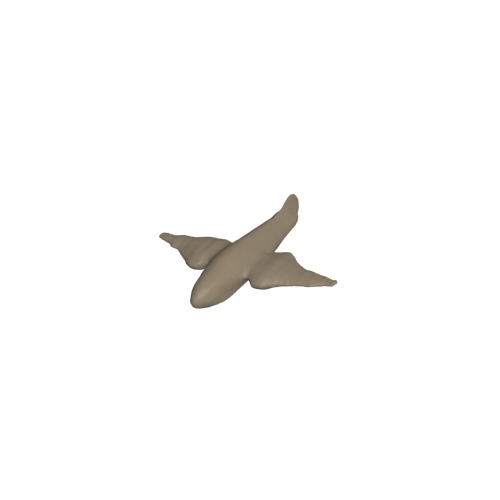}
    \includegraphics[width=0.115\linewidth, trim={\trimR cm \trimR cm \trimR cm \trimR cm}, clip]{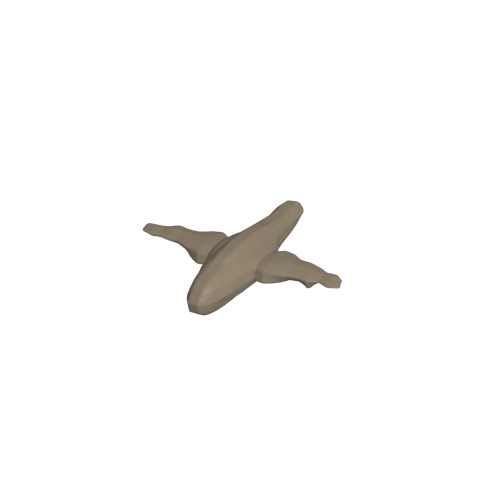}
    \includegraphics[width=0.115\linewidth, trim={\trimR cm \trimR cm \trimR cm \trimR cm}, clip]{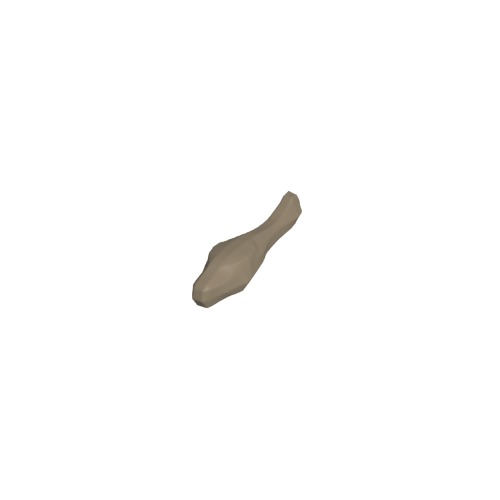}\\ 
    \includegraphics[width=0.115\linewidth, trim={\trimI cm \trimI cm \trimI cm \trimI cm}, clip]{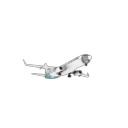}
    \includegraphics[width=0.115\linewidth, trim={\trimR cm \trimR cm \trimR cm \trimR cm}, clip]{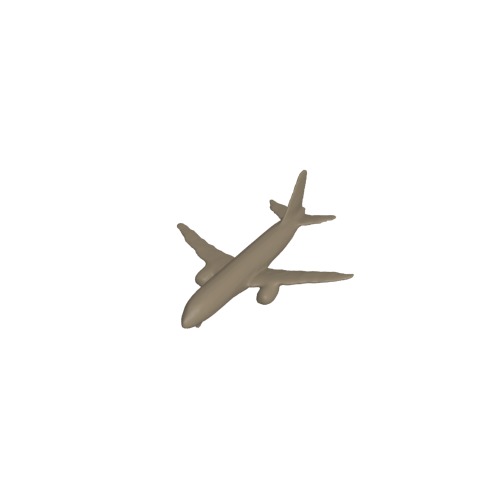}
    \includegraphics[width=0.115\linewidth, trim={\trimR cm \trimR cm \trimR cm \trimR cm}, clip]{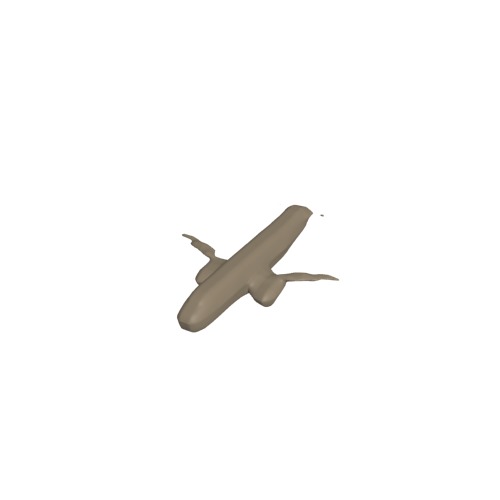}
    \includegraphics[width=0.115\linewidth, trim={\trimR cm \trimR cm \trimR cm \trimR cm}, clip]{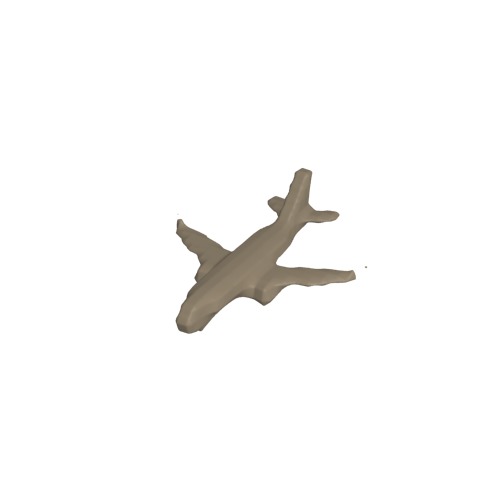}\hfill
    \includegraphics[width=0.115\linewidth, trim={\trimI cm \trimI cm \trimI cm \trimI cm}, clip]{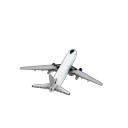}
    \includegraphics[width=0.115\linewidth, trim={\trimR cm \trimR cm \trimR cm \trimR cm}, clip]{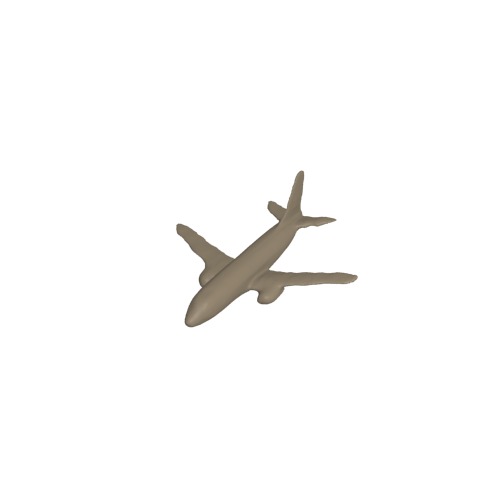}
    \includegraphics[width=0.115\linewidth, trim={\trimR cm \trimR cm \trimR cm \trimR cm}, clip]{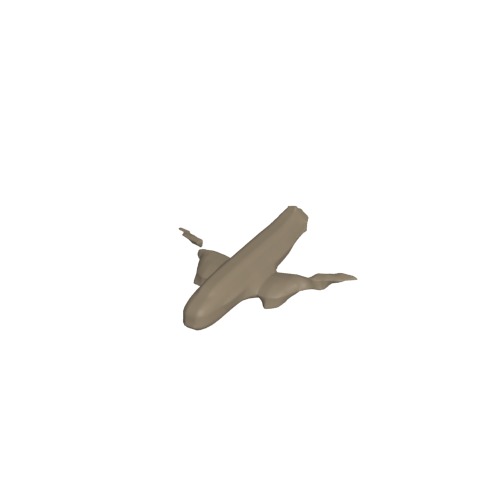}
    \includegraphics[width=0.115\linewidth, trim={\trimR cm \trimR cm \trimR cm \trimR cm}, clip]{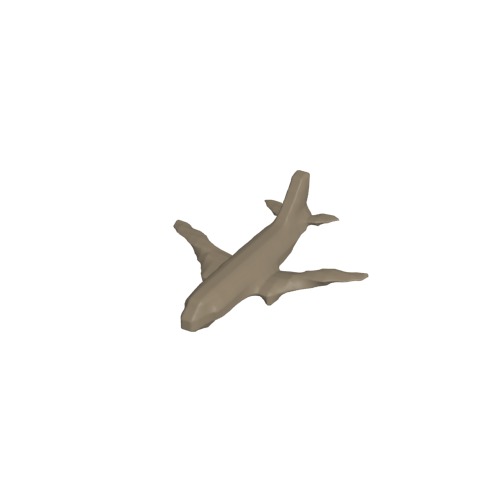}\\   
    \includegraphics[width=0.115\linewidth, trim={\trimI cm \trimI cm \trimI cm \trimI cm}, clip]{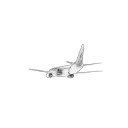}
    \includegraphics[width=0.115\linewidth, trim={\trimR cm \trimR cm \trimR cm \trimR cm}, clip]{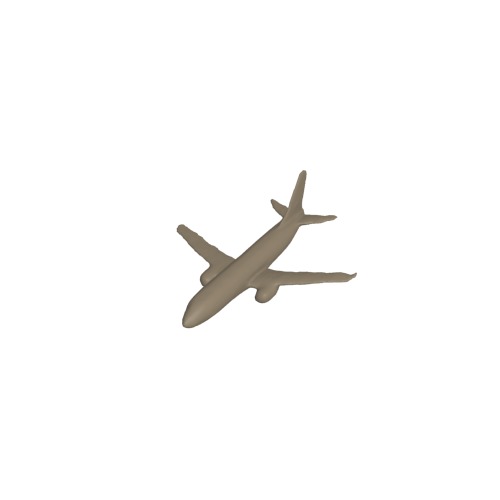}
    \includegraphics[width=0.115\linewidth, trim={\trimR cm \trimR cm \trimR cm \trimR cm}, clip]{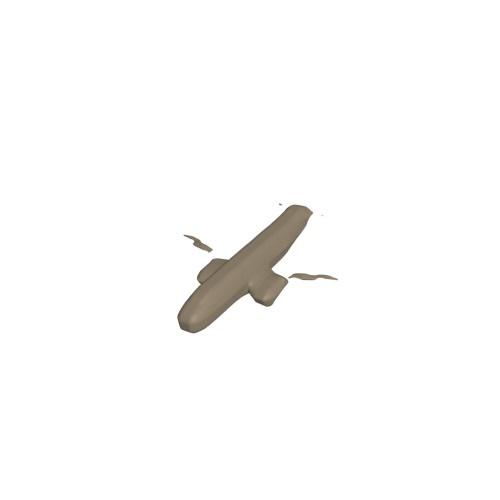}
    \includegraphics[width=0.115\linewidth, trim={\trimR cm \trimR cm \trimR cm \trimR cm}, clip]{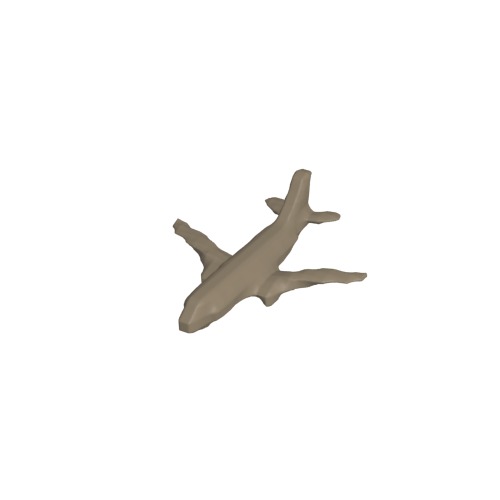}\hfill
    \includegraphics[width=0.115\linewidth, trim={\trimI cm \trimI cm \trimI cm \trimI cm}, clip]{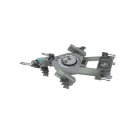}
    \includegraphics[width=0.115\linewidth, trim={\trimR cm \trimR cm \trimR cm \trimR cm}, clip]{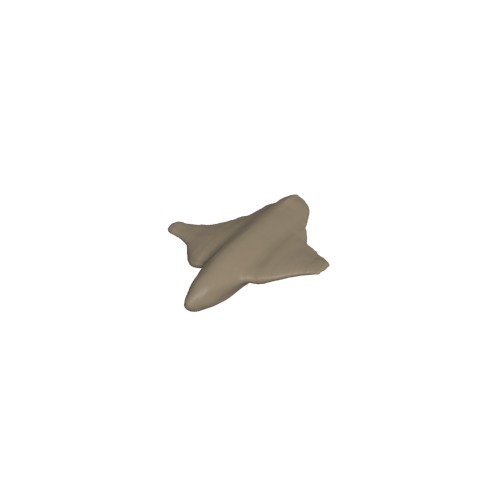}
    \includegraphics[width=0.115\linewidth, trim={\trimR cm \trimR cm \trimR cm \trimR cm}, clip]{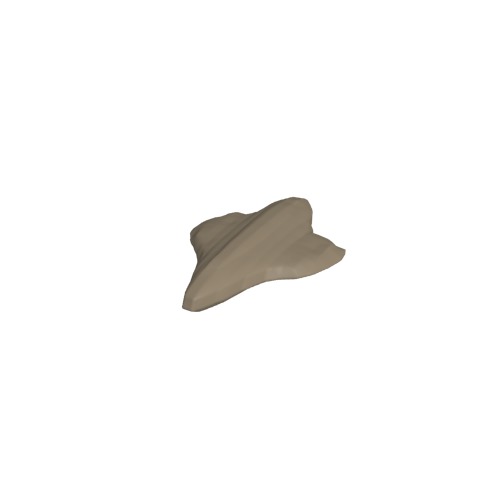}
    \includegraphics[width=0.115\linewidth, trim={\trimR cm \trimR cm \trimR cm \trimR cm}, clip]{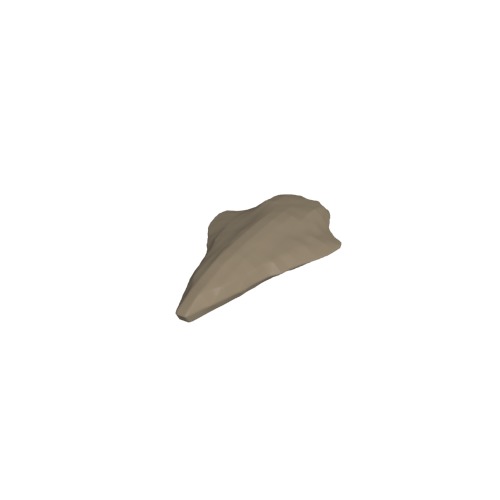}\\
    \includegraphics[width=0.115\linewidth, trim={\trimI cm \trimI cm \trimI cm \trimI cm}, clip]{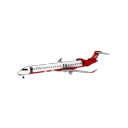}
    \includegraphics[width=0.115\linewidth, trim={\trimR cm \trimR cm \trimR cm \trimR cm}, clip]{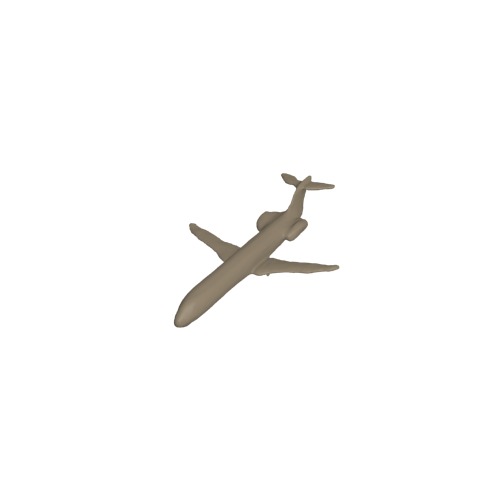}
    \includegraphics[width=0.115\linewidth, trim={\trimR cm \trimR cm \trimR cm \trimR cm}, clip]{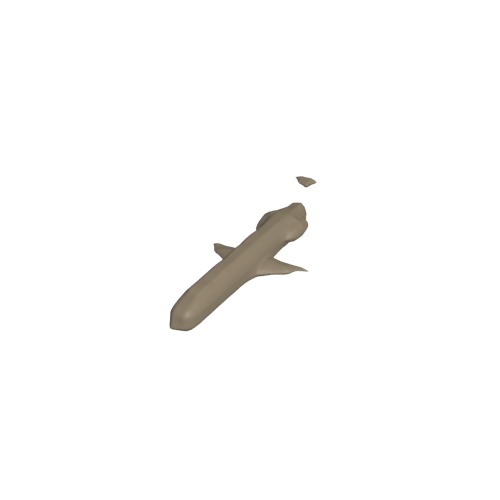}
    \includegraphics[width=0.115\linewidth, trim={\trimR cm \trimR cm \trimR cm \trimR cm}, clip]{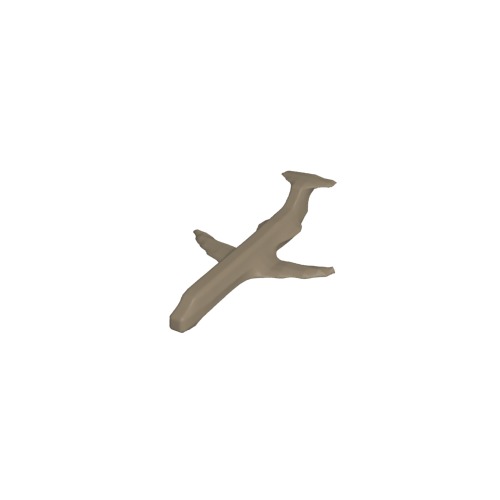}\hfill
    \includegraphics[width=0.115\linewidth, trim={\trimI cm \trimI cm \trimI cm \trimI cm}, clip]{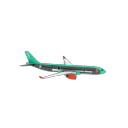}
    \includegraphics[width=0.115\linewidth, trim={\trimR cm \trimR cm \trimR cm \trimR cm}, clip]{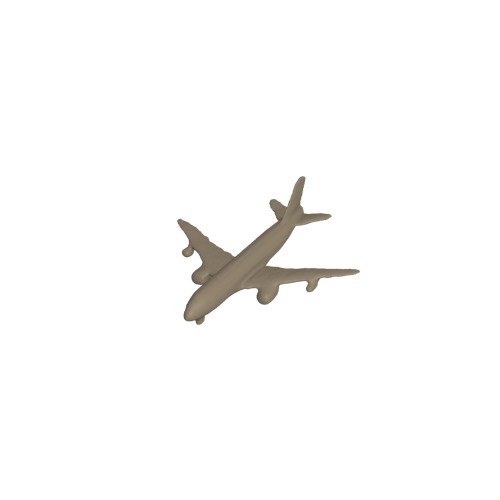}
    \includegraphics[width=0.115\linewidth, trim={\trimR cm \trimR cm \trimR cm \trimR cm}, clip]{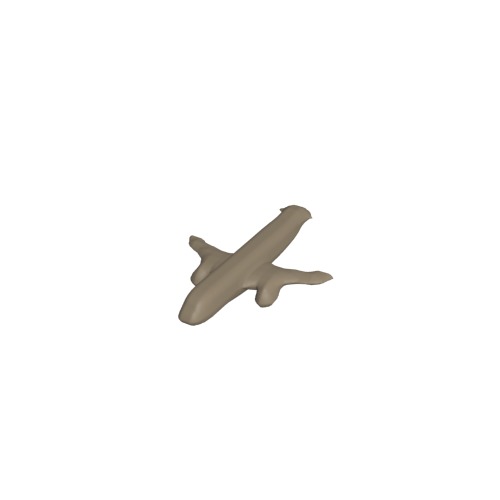}
    \includegraphics[width=0.115\linewidth, trim={\trimR cm \trimR cm \trimR cm \trimR cm}, clip]{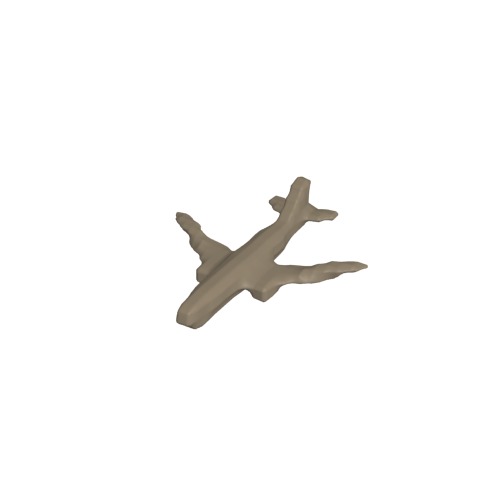}\\
    \includegraphics[width=0.115\linewidth, trim={\trimI cm \trimI cm \trimI cm \trimI cm}, clip]{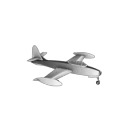}
    \includegraphics[width=0.115\linewidth, trim={\trimR cm \trimR cm \trimR cm \trimR cm}, clip]{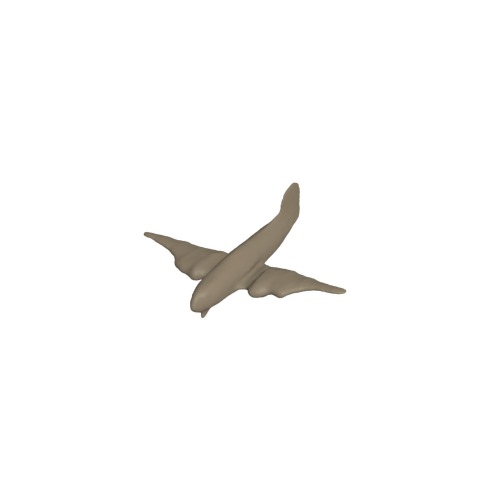}
    \includegraphics[width=0.115\linewidth, trim={\trimR cm \trimR cm \trimR cm \trimR cm}, clip]{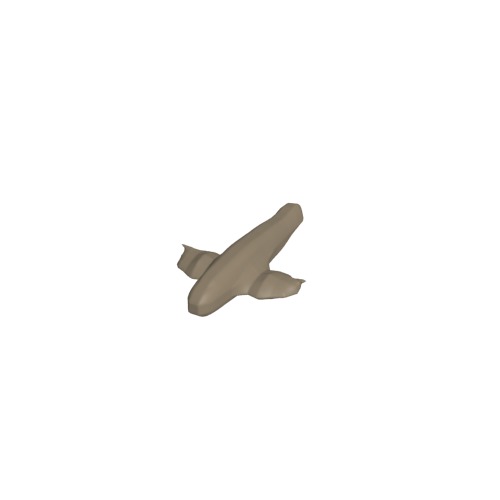}
    \includegraphics[width=0.115\linewidth, trim={\trimR cm \trimR cm \trimR cm \trimR cm}, clip]{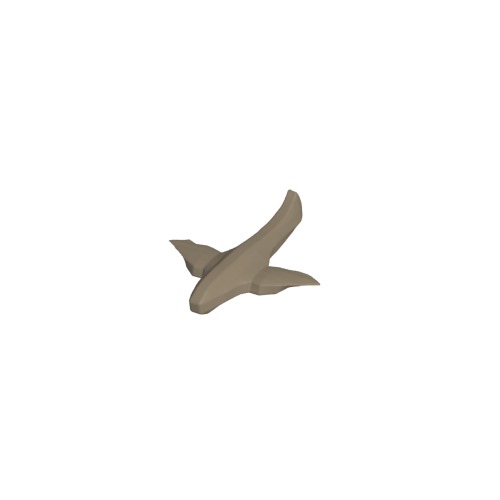}\hfill
    \includegraphics[width=0.115\linewidth, trim={\trimI cm \trimI cm \trimI cm \trimI cm}, clip]{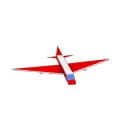}
    \includegraphics[width=0.115\linewidth, trim={\trimR cm \trimR cm \trimR cm \trimR cm}, clip]{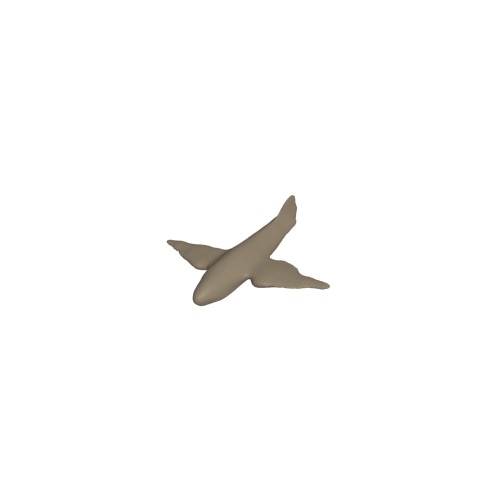}
    \includegraphics[width=0.115\linewidth, trim={\trimR cm \trimR cm \trimR cm \trimR cm}, clip]{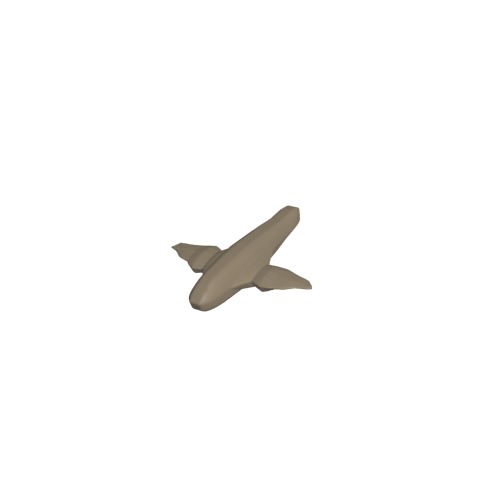}
    \includegraphics[width=0.115\linewidth, trim={\trimR cm \trimR cm \trimR cm \trimR cm}, clip]{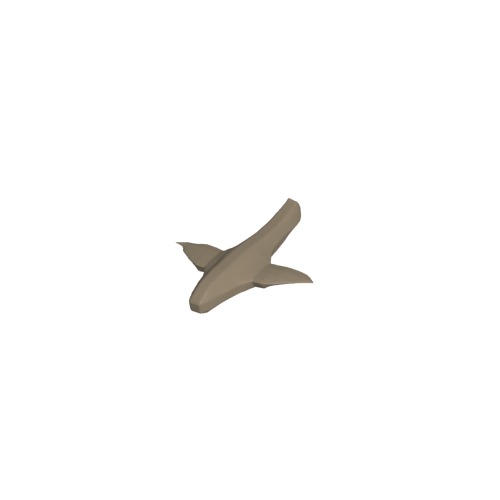}\\
    \caption{Results for the category aeroplane. Each block from left to right, input image, our proposed HSP, LR Soft, LR Hard.}
    \label{fig:aero1}
\end{figure*}

  \begin{figure*}
    \includegraphics[width=0.115\linewidth, trim={\trimI cm \trimI cm \trimI cm \trimI cm}, clip]{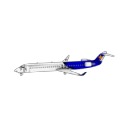}
    \includegraphics[width=0.115\linewidth, trim={\trimR cm \trimR cm \trimR cm \trimR cm}, clip]{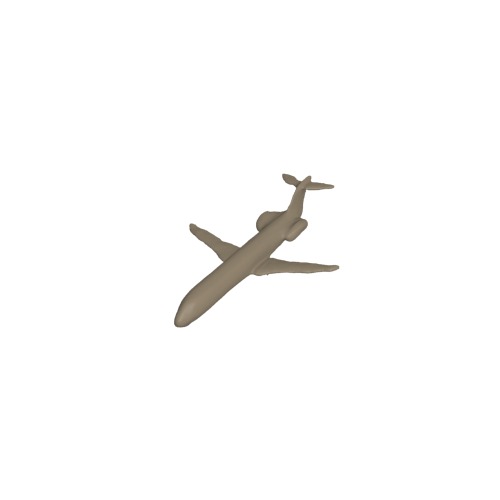}
    \includegraphics[width=0.115\linewidth, trim={\trimR cm \trimR cm \trimR cm \trimR cm}, clip]{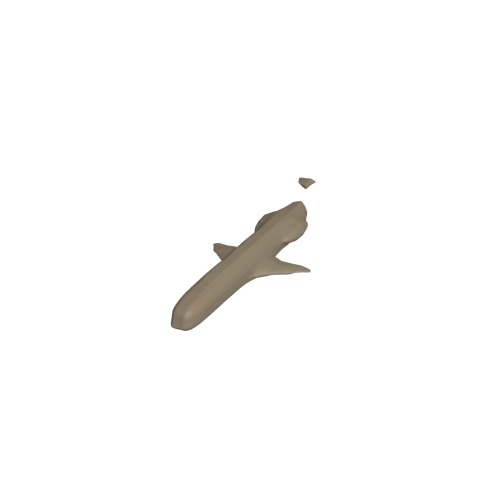}
    \includegraphics[width=0.115\linewidth, trim={\trimR cm \trimR cm \trimR cm \trimR cm}, clip]{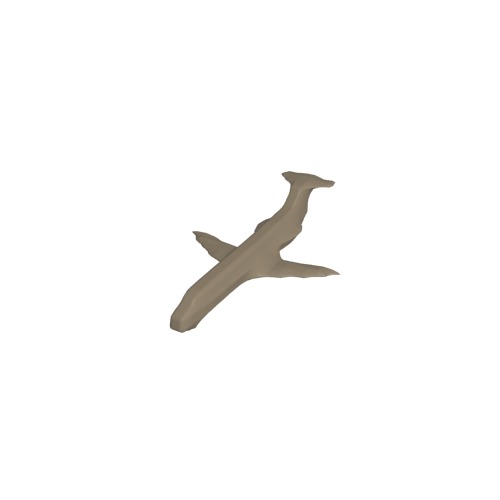}\hfill
    \includegraphics[width=0.115\linewidth, trim={\trimI cm \trimI cm \trimI cm \trimI cm}, clip]{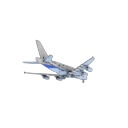}
    \includegraphics[width=0.115\linewidth, trim={\trimR cm \trimR cm \trimR cm \trimR cm}, clip]{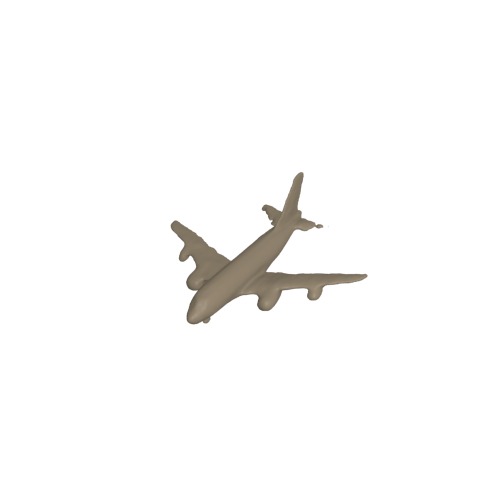}
    \includegraphics[width=0.115\linewidth, trim={\trimR cm \trimR cm \trimR cm \trimR cm}, clip]{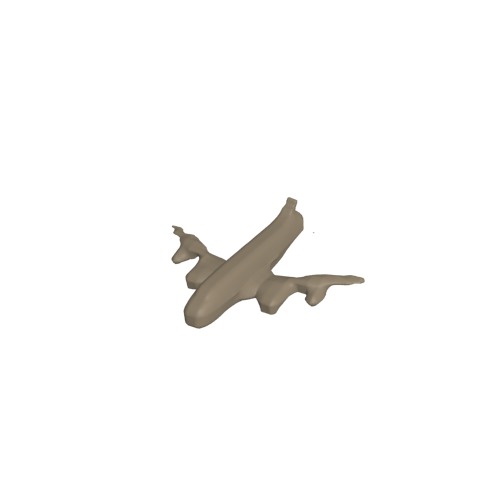}
    \includegraphics[width=0.115\linewidth, trim={\trimR cm \trimR cm \trimR cm \trimR cm}, clip]{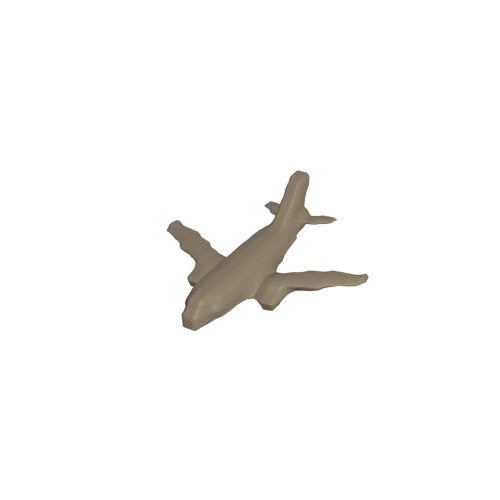} \\
    \includegraphics[width=0.115\linewidth, trim={\trimI cm \trimI cm \trimI cm \trimI cm}, clip]{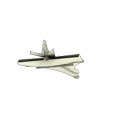}
    \includegraphics[width=0.115\linewidth, trim={\trimR cm \trimR cm \trimR cm \trimR cm}, clip]{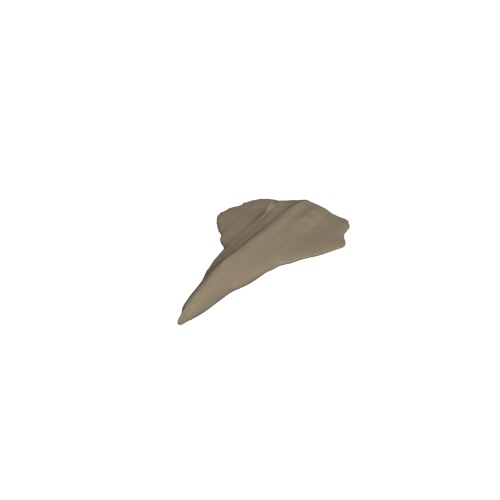}
    \includegraphics[width=0.115\linewidth, trim={\trimR cm \trimR cm \trimR cm \trimR cm}, clip]{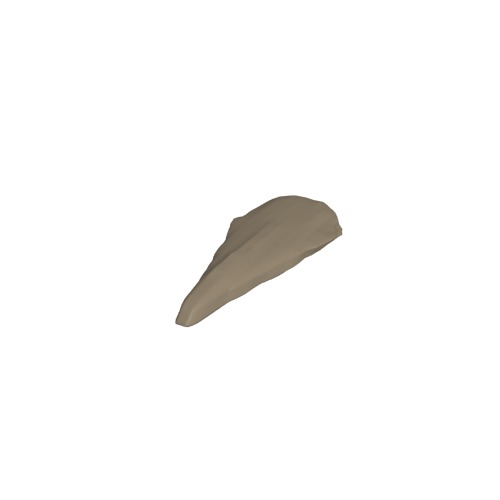}
    \includegraphics[width=0.115\linewidth, trim={\trimR cm \trimR cm \trimR cm \trimR cm}, clip]{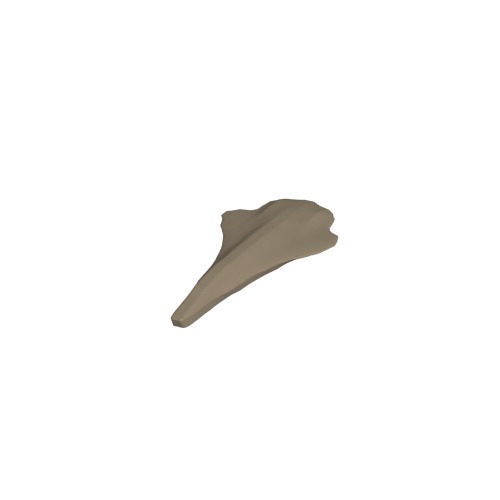}\hfill
    \includegraphics[width=0.115\linewidth, trim={\trimI cm \trimI cm \trimI cm \trimI cm}, clip]{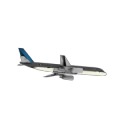}
    \includegraphics[width=0.115\linewidth, trim={\trimR cm \trimR cm \trimR cm \trimR cm}, clip]{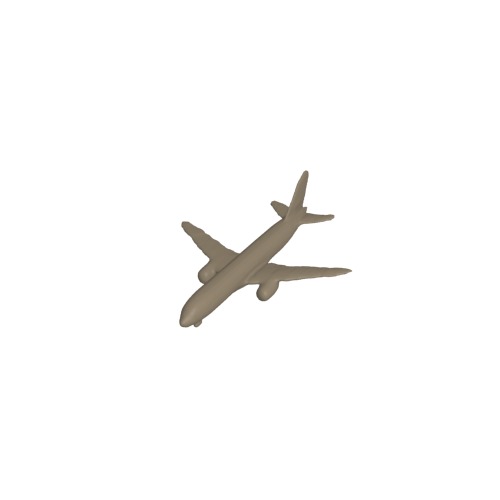}
    \includegraphics[width=0.115\linewidth, trim={\trimR cm \trimR cm \trimR cm \trimR cm}, clip]{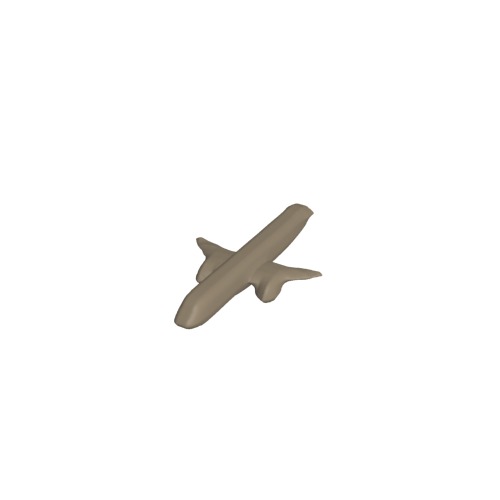}
    \includegraphics[width=0.115\linewidth, trim={\trimR cm \trimR cm \trimR cm \trimR cm}, clip]{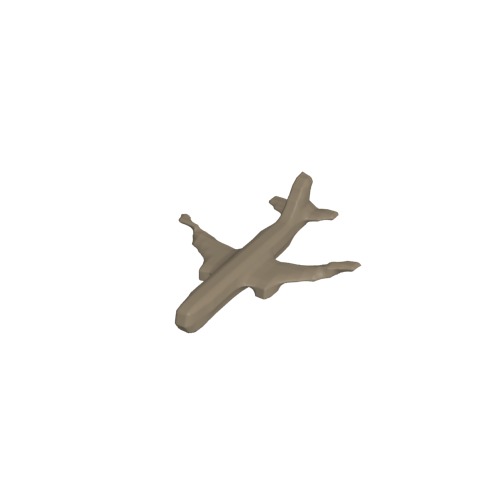} \\
    \includegraphics[width=0.115\linewidth, trim={\trimI cm \trimI cm \trimI cm \trimI cm}, clip]{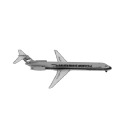}
    \includegraphics[width=0.115\linewidth, trim={\trimR cm \trimR cm \trimR cm \trimR cm}, clip]{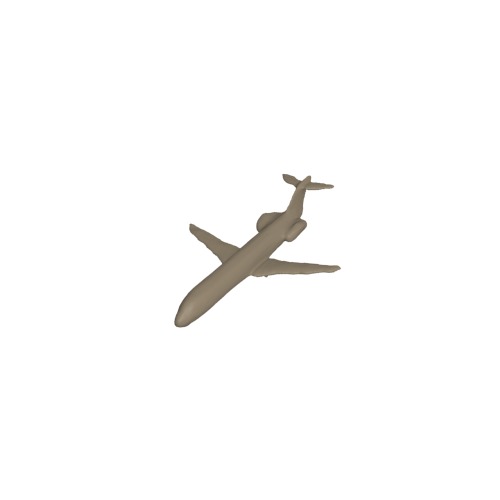}
    \includegraphics[width=0.115\linewidth, trim={\trimR cm \trimR cm \trimR cm \trimR cm}, clip]{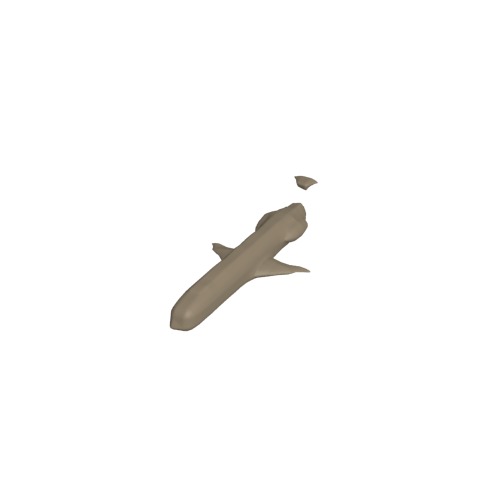}
    \includegraphics[width=0.115\linewidth, trim={\trimR cm \trimR cm \trimR cm \trimR cm}, clip]{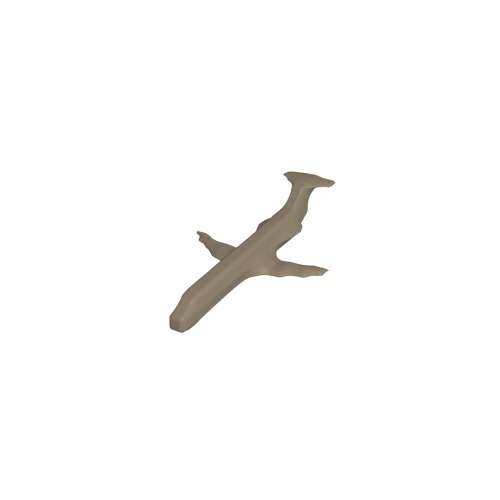}\hfill
    \includegraphics[width=0.115\linewidth, trim={\trimI cm \trimI cm \trimI cm \trimI cm}, clip]{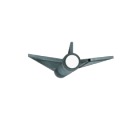}
    \includegraphics[width=0.115\linewidth, trim={\trimR cm \trimR cm \trimR cm \trimR cm}, clip]{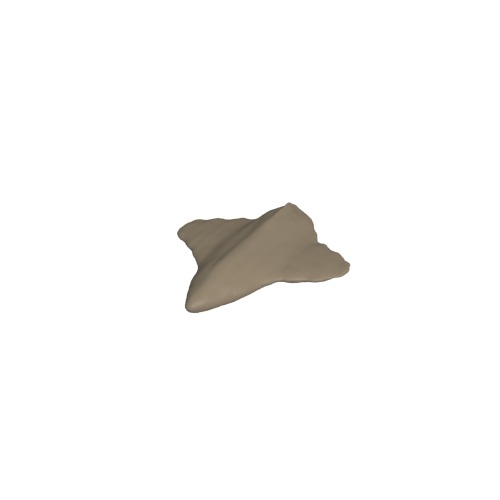}
    \includegraphics[width=0.115\linewidth, trim={\trimR cm \trimR cm \trimR cm \trimR cm}, clip]{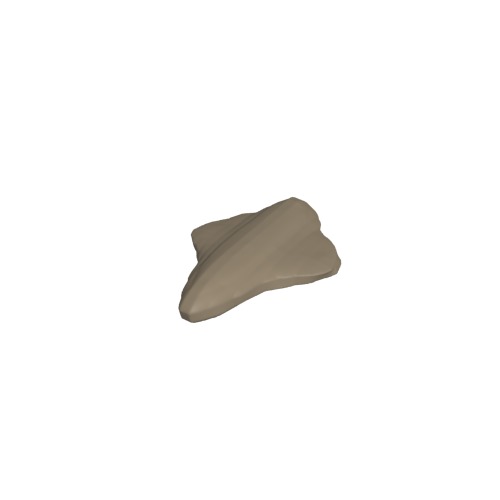}
    \includegraphics[width=0.115\linewidth, trim={\trimR cm \trimR cm \trimR cm \trimR cm}, clip]{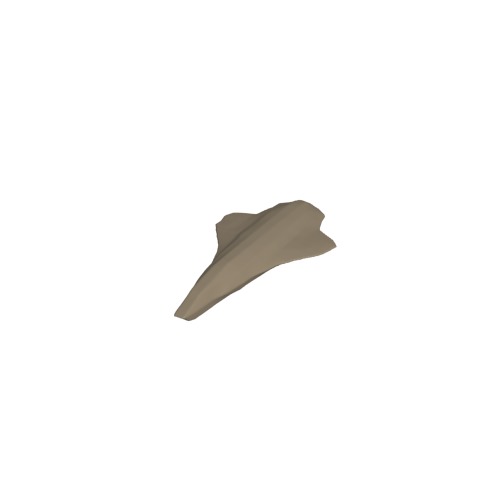}\\
    \includegraphics[width=0.115\linewidth, trim={\trimI cm \trimI cm \trimI cm \trimI cm}, clip]{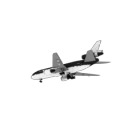}
    \includegraphics[width=0.115\linewidth, trim={\trimR cm \trimR cm \trimR cm \trimR cm}, clip]{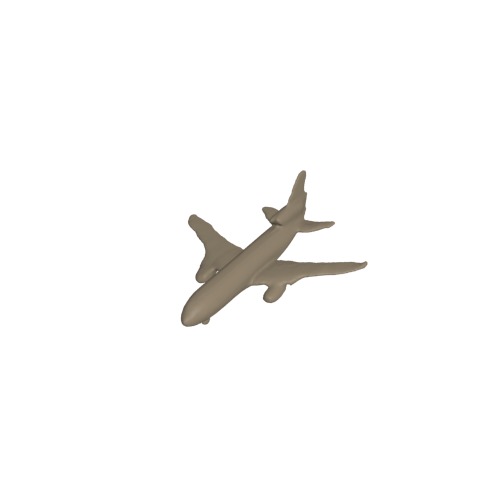}
    \includegraphics[width=0.115\linewidth, trim={\trimR cm \trimR cm \trimR cm \trimR cm}, clip]{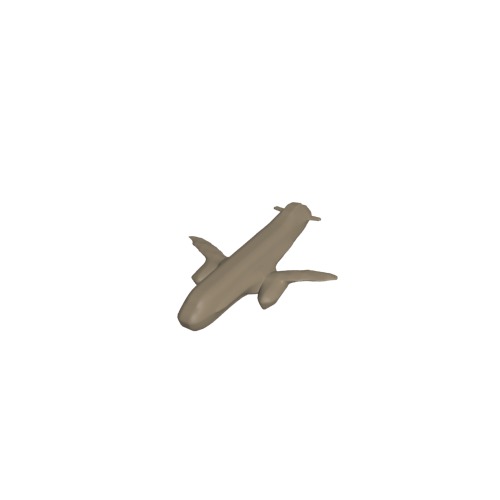}
    \includegraphics[width=0.115\linewidth, trim={\trimR cm \trimR cm \trimR cm \trimR cm}, clip]{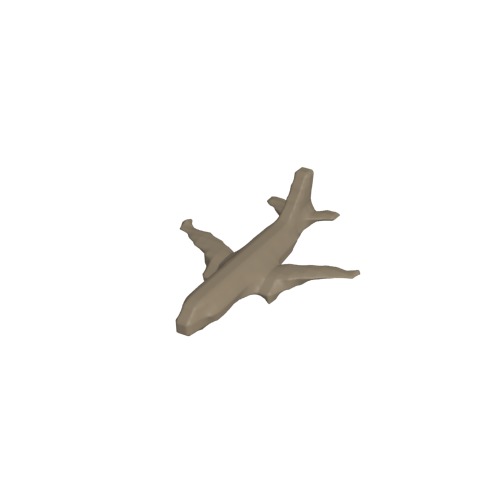}
    \caption{Results for the category aeroplane continued. Each block from left to right, input image, our proposed HSP, LR Soft, LR Hard.}
    \label{fig:aero2}
  \end{figure*}

  \begin{figure*}
    \includegraphics[width=0.115\linewidth, trim={\trimI cm \trimI cm \trimI cm \trimI cm}, clip]{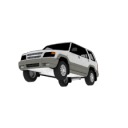}
    \includegraphics[width=0.115\linewidth, trim={\trimR cm \trimR cm \trimR cm \trimR cm}, clip]{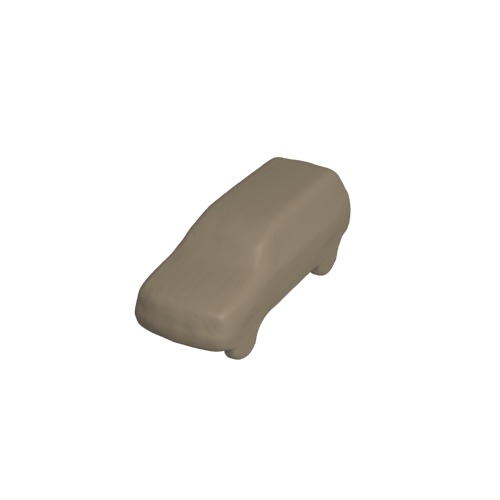}
    \includegraphics[width=0.115\linewidth, trim={\trimR cm \trimR cm \trimR cm \trimR cm}, clip]{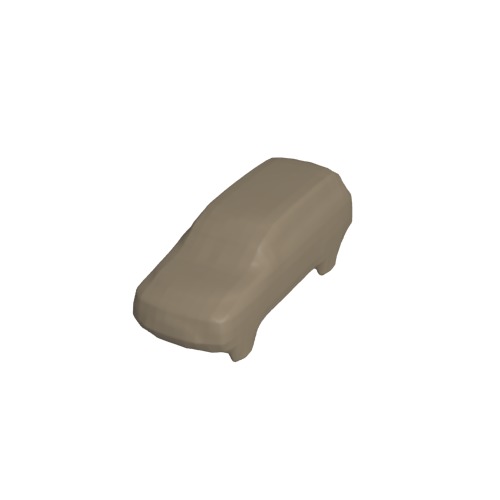}
    \includegraphics[width=0.115\linewidth, trim={\trimR cm \trimR cm \trimR cm \trimR cm}, clip]{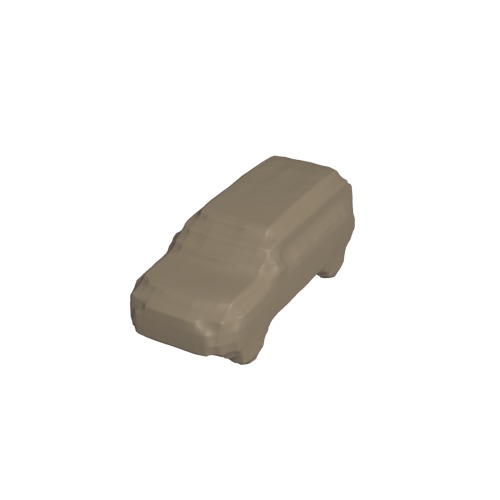}\hfill
    \includegraphics[width=0.115\linewidth, trim={\trimI cm \trimI cm \trimI cm \trimI cm}, clip]{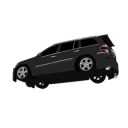}
    \includegraphics[width=0.115\linewidth, trim={\trimR cm \trimR cm \trimR cm \trimR cm}, clip]{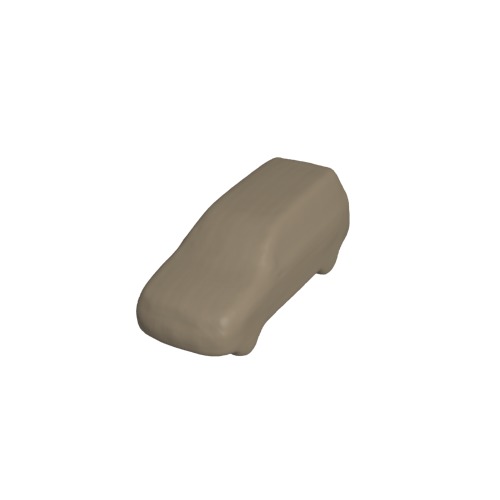}
    \includegraphics[width=0.115\linewidth, trim={\trimR cm \trimR cm \trimR cm \trimR cm}, clip]{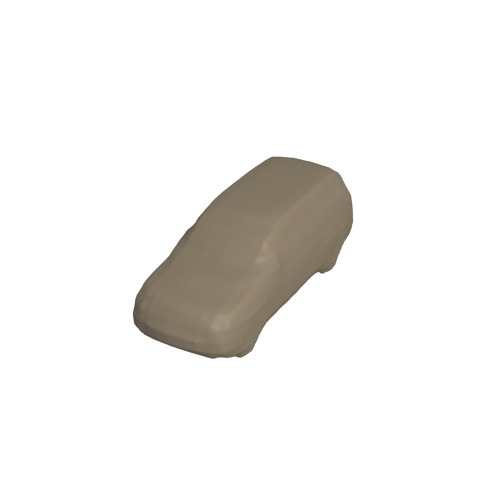}
    \includegraphics[width=0.115\linewidth, trim={\trimR cm \trimR cm \trimR cm \trimR cm}, clip]{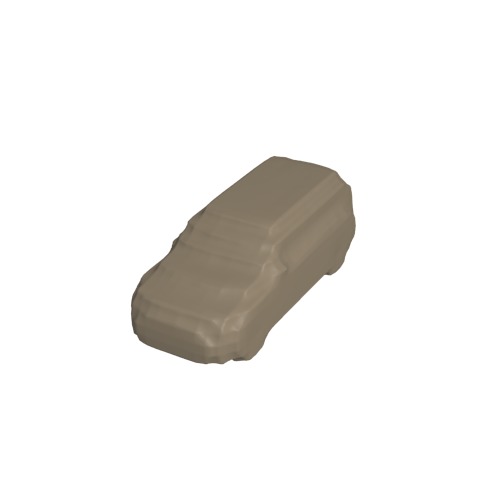}\\
    \includegraphics[width=0.115\linewidth, trim={\trimI cm \trimI cm \trimI cm \trimI cm}, clip]{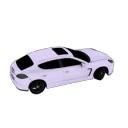}
    \includegraphics[width=0.115\linewidth, trim={\trimR cm \trimR cm \trimR cm \trimR cm}, clip]{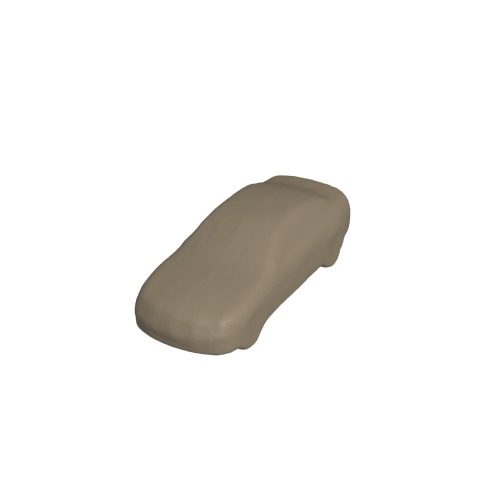}
    \includegraphics[width=0.115\linewidth, trim={\trimR cm \trimR cm \trimR cm \trimR cm}, clip]{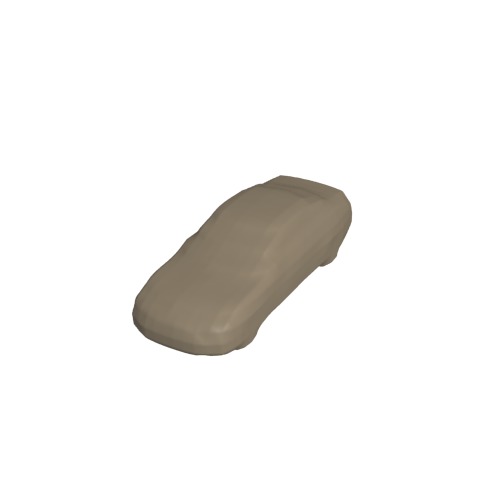}
    \includegraphics[width=0.115\linewidth, trim={\trimR cm \trimR cm \trimR cm \trimR cm}, clip]{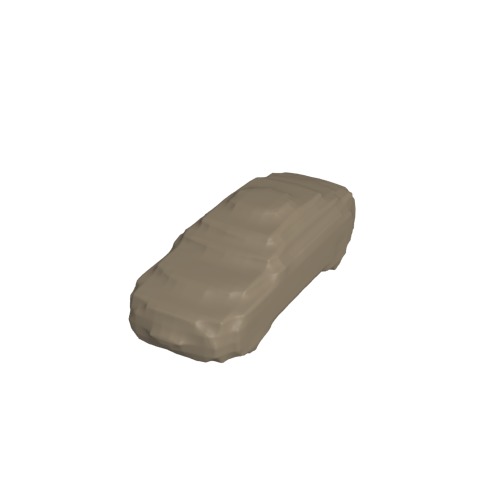}\hfill
    \includegraphics[width=0.115\linewidth, trim={\trimI cm \trimI cm \trimI cm \trimI cm}, clip]{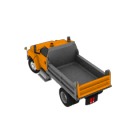}
    \includegraphics[width=0.115\linewidth, trim={\trimR cm \trimR cm \trimR cm \trimR cm}, clip]{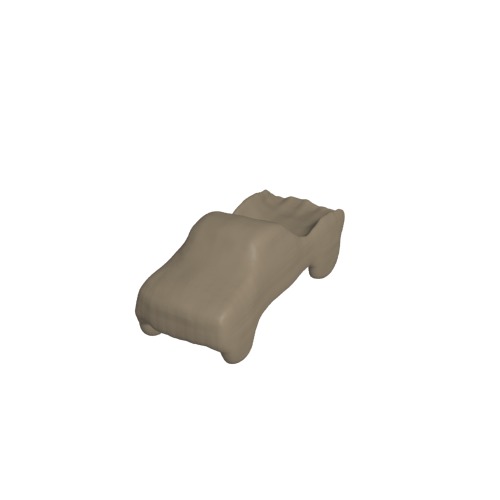}
    \includegraphics[width=0.115\linewidth, trim={\trimR cm \trimR cm \trimR cm \trimR cm}, clip]{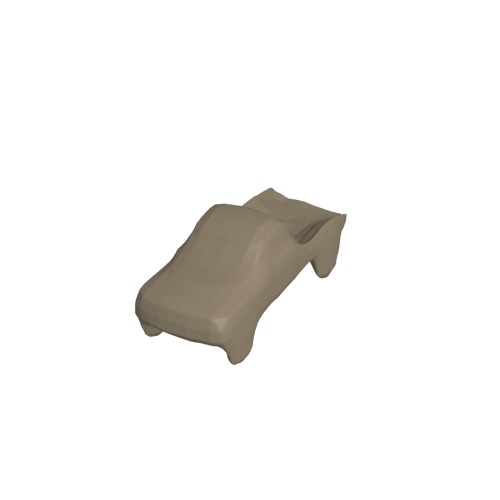}
    \includegraphics[width=0.115\linewidth, trim={\trimR cm \trimR cm \trimR cm \trimR cm}, clip]{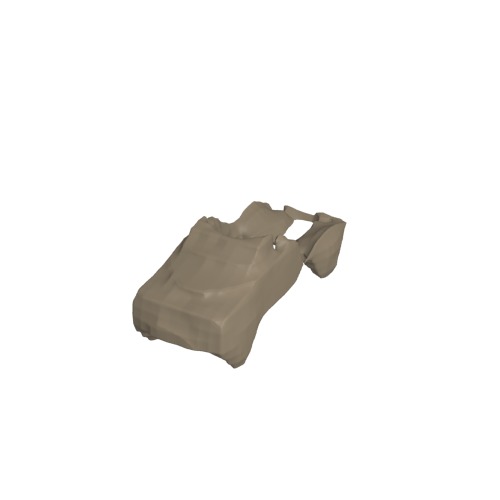}\\
    \includegraphics[width=0.115\linewidth, trim={\trimI cm \trimI cm \trimI cm \trimI cm}, clip]{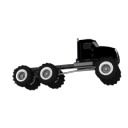}
    \includegraphics[width=0.115\linewidth, trim={\trimR cm \trimR cm \trimR cm \trimR cm}, clip]{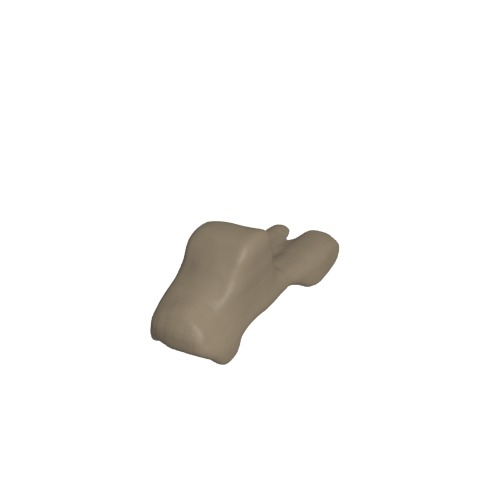}
    \includegraphics[width=0.115\linewidth, trim={\trimR cm \trimR cm \trimR cm \trimR cm}, clip]{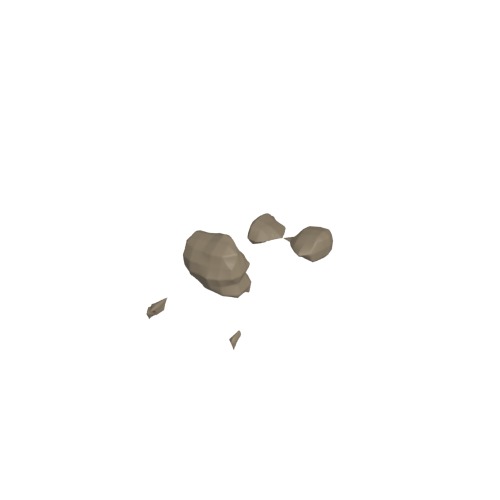}
    \includegraphics[width=0.115\linewidth, trim={\trimR cm \trimR cm \trimR cm \trimR cm}, clip]{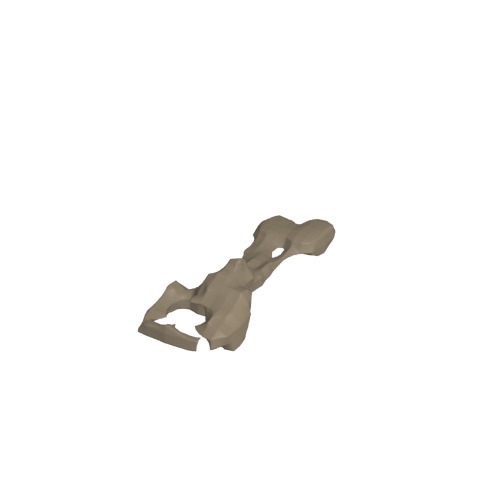}\hfill
    \includegraphics[width=0.115\linewidth, trim={\trimI cm \trimI cm \trimI cm \trimI cm}, clip]{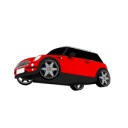}
    \includegraphics[width=0.115\linewidth, trim={\trimR cm \trimR cm \trimR cm \trimR cm}, clip]{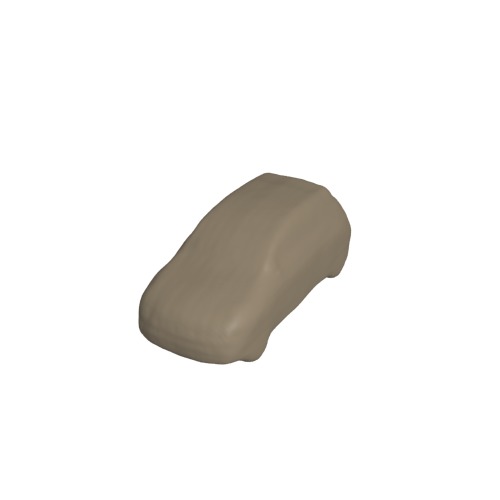}
    \includegraphics[width=0.115\linewidth, trim={\trimR cm \trimR cm \trimR cm \trimR cm}, clip]{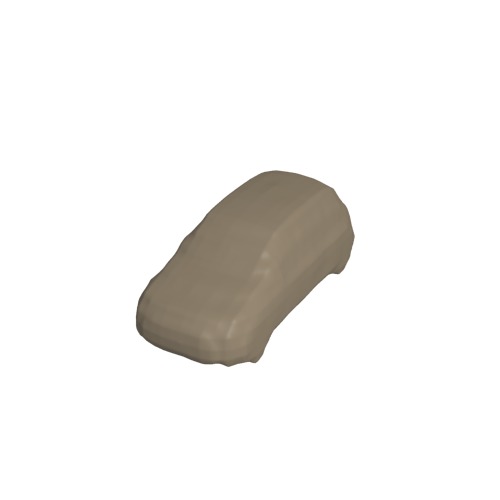}
    \includegraphics[width=0.115\linewidth, trim={\trimR cm \trimR cm \trimR cm \trimR cm}, clip]{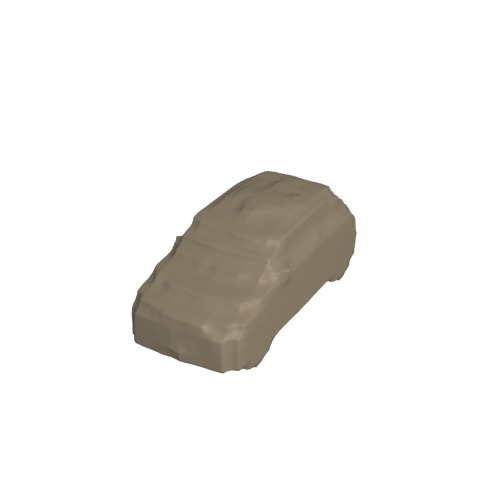}\\
    \includegraphics[width=0.115\linewidth, trim={\trimI cm \trimI cm \trimI cm \trimI cm}, clip]{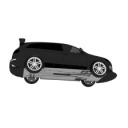}
    \includegraphics[width=0.115\linewidth, trim={\trimR cm \trimR cm \trimR cm \trimR cm}, clip]{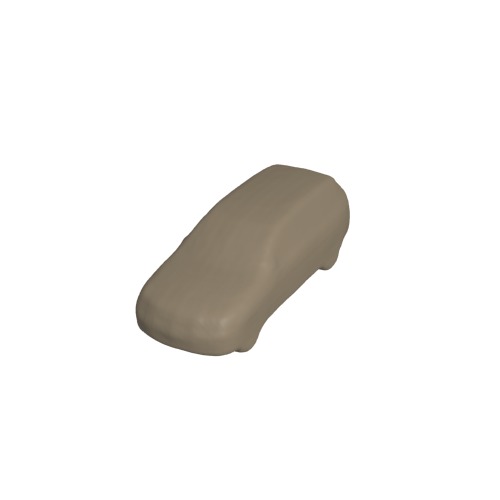}
    \includegraphics[width=0.115\linewidth, trim={\trimR cm \trimR cm \trimR cm \trimR cm}, clip]{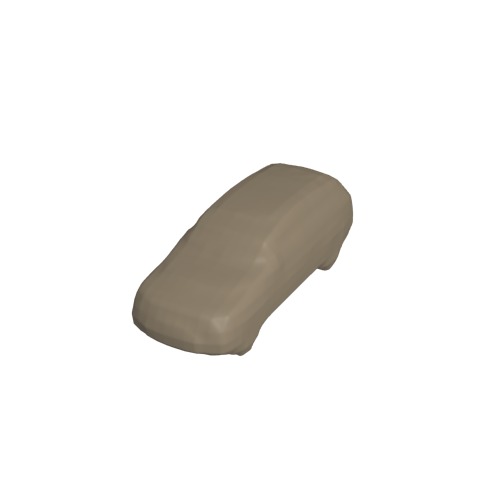}
    \includegraphics[width=0.115\linewidth, trim={\trimR cm \trimR cm \trimR cm \trimR cm}, clip]{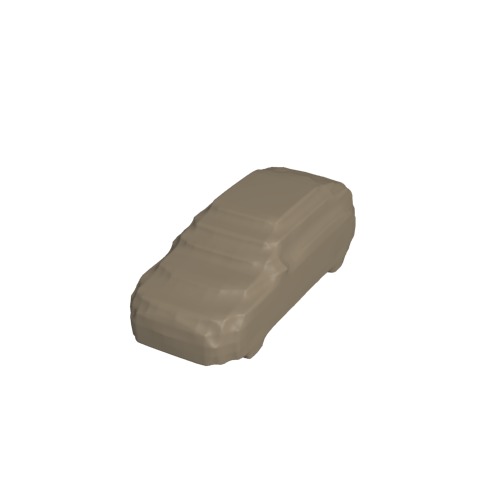}\hfill
    \includegraphics[width=0.115\linewidth, trim={\trimI cm \trimI cm \trimI cm \trimI cm}, clip]{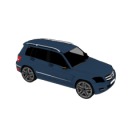}
    \includegraphics[width=0.115\linewidth, trim={\trimR cm \trimR cm \trimR cm \trimR cm}, clip]{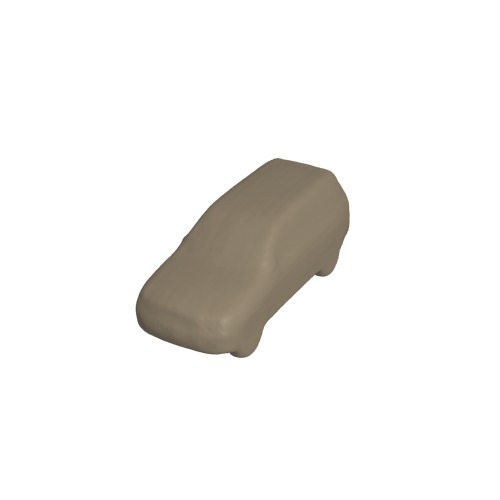}
    \includegraphics[width=0.115\linewidth, trim={\trimR cm \trimR cm \trimR cm \trimR cm}, clip]{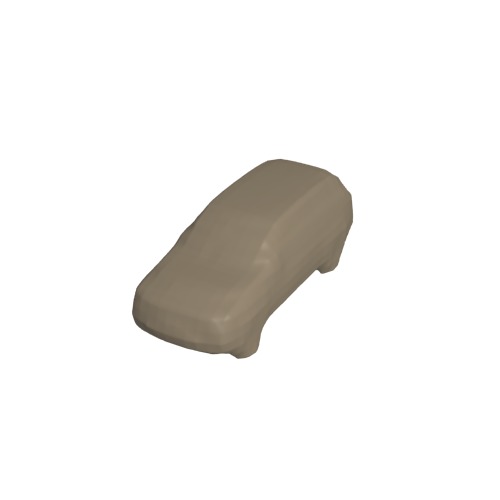}
    \includegraphics[width=0.115\linewidth, trim={\trimR cm \trimR cm \trimR cm \trimR cm}, clip]{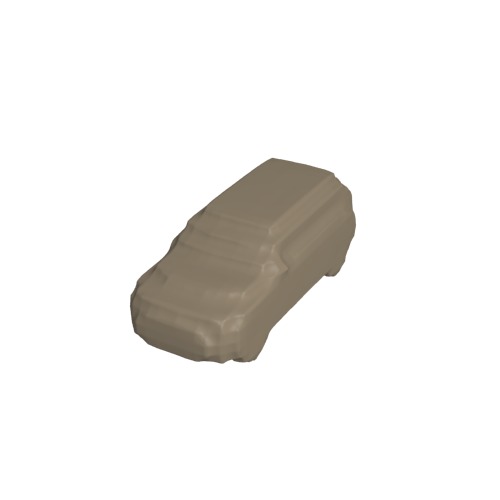}\\
    \includegraphics[width=0.115\linewidth, trim={\trimI cm \trimI cm \trimI cm \trimI cm}, clip]{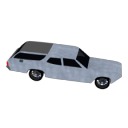}
    \includegraphics[width=0.115\linewidth, trim={\trimR cm \trimR cm \trimR cm \trimR cm}, clip]{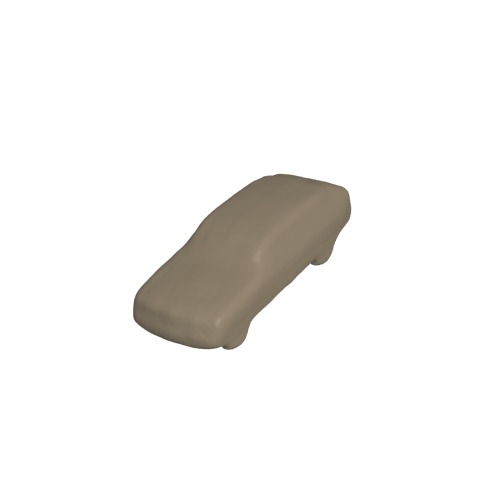}
    \includegraphics[width=0.115\linewidth, trim={\trimR cm \trimR cm \trimR cm \trimR cm}, clip]{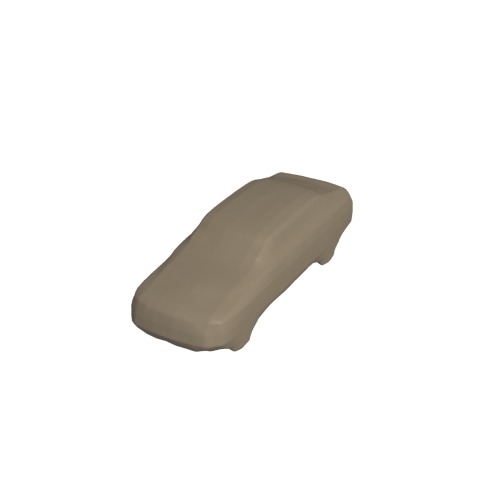}
    \includegraphics[width=0.115\linewidth, trim={\trimR cm \trimR cm \trimR cm \trimR cm}, clip]{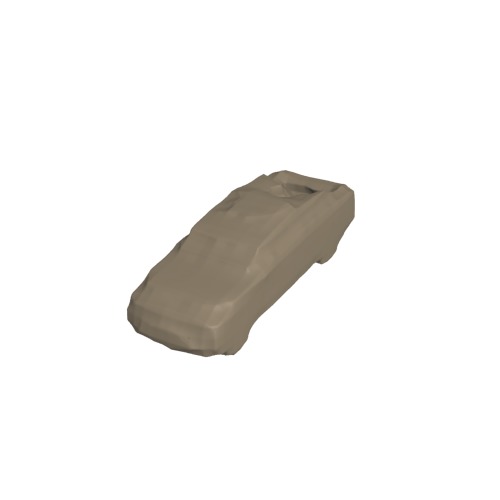}\hfill
    \includegraphics[width=0.115\linewidth, trim={\trimI cm \trimI cm \trimI cm \trimI cm}, clip]{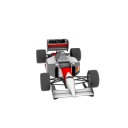}
    \includegraphics[width=0.115\linewidth, trim={\trimR cm \trimR cm \trimR cm \trimR cm}, clip]{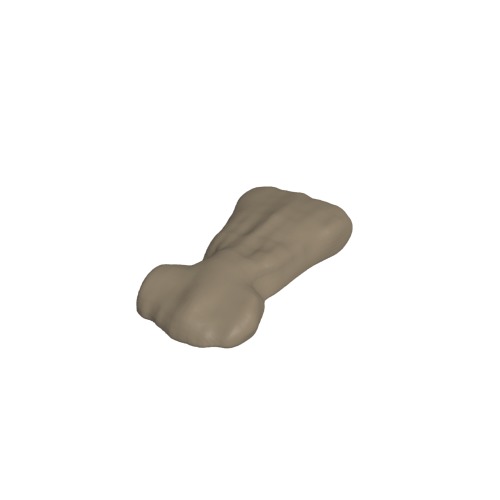}
    \includegraphics[width=0.115\linewidth, trim={\trimR cm \trimR cm \trimR cm \trimR cm}, clip]{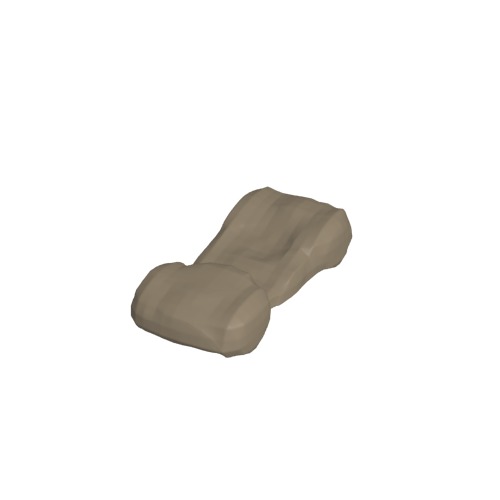}
    \includegraphics[width=0.115\linewidth, trim={\trimR cm \trimR cm \trimR cm \trimR cm}, clip]{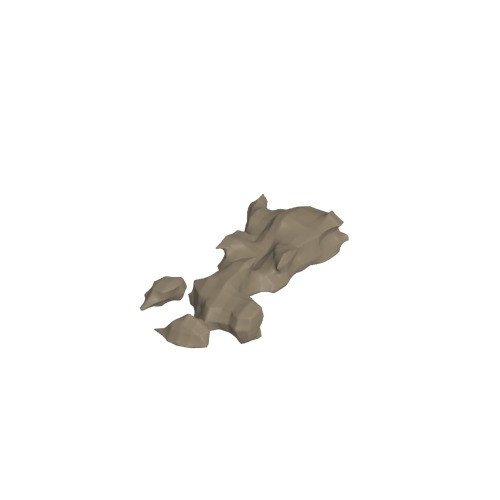}\\
    \includegraphics[width=0.115\linewidth, trim={\trimI cm \trimI cm \trimI cm \trimI cm}, clip]{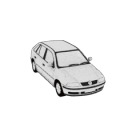}
    \includegraphics[width=0.115\linewidth, trim={\trimR cm \trimR cm \trimR cm \trimR cm}, clip]{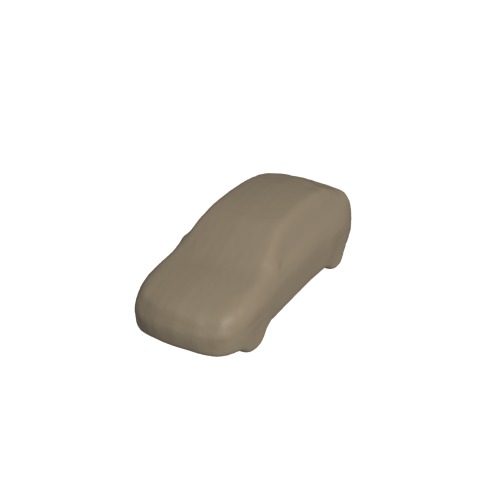}
    \includegraphics[width=0.115\linewidth, trim={\trimR cm \trimR cm \trimR cm \trimR cm}, clip]{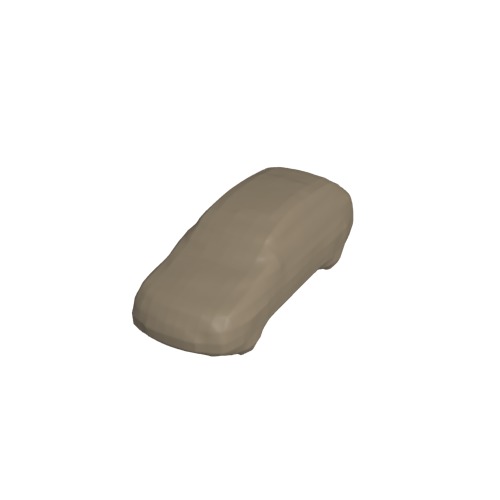}
    \includegraphics[width=0.115\linewidth, trim={\trimR cm \trimR cm \trimR cm \trimR cm}, clip]{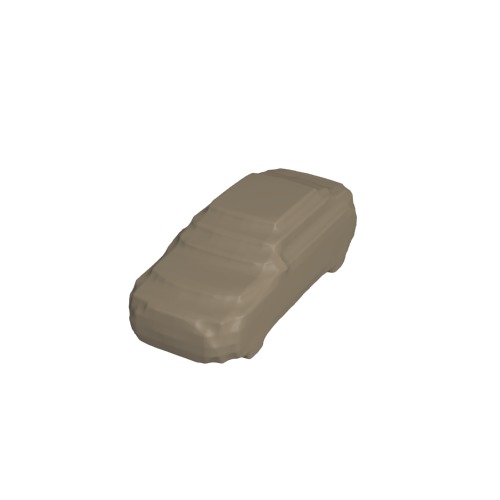}\hfill
    \includegraphics[width=0.115\linewidth, trim={\trimI cm \trimI cm \trimI cm \trimI cm}, clip]{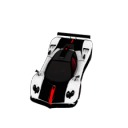}
    \includegraphics[width=0.115\linewidth, trim={\trimR cm \trimR cm \trimR cm \trimR cm}, clip]{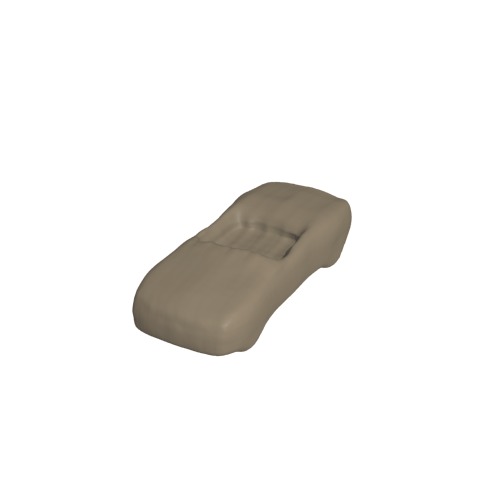}
    \includegraphics[width=0.115\linewidth, trim={\trimR cm \trimR cm \trimR cm \trimR cm}, clip]{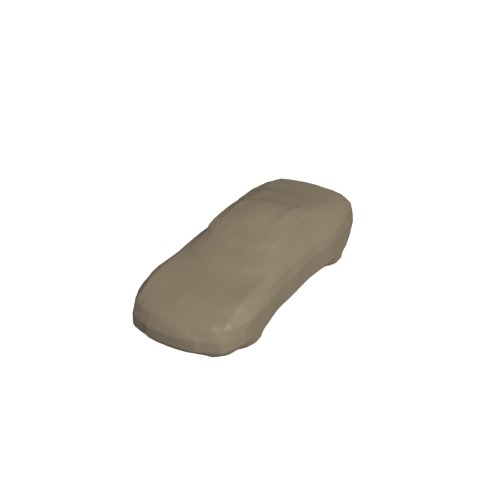}
    \includegraphics[width=0.115\linewidth, trim={\trimR cm \trimR cm \trimR cm \trimR cm}, clip]{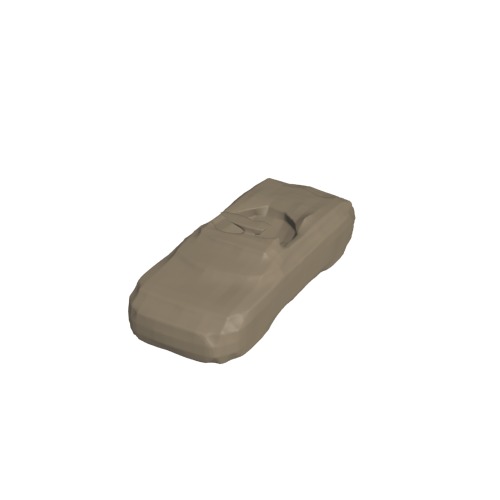}
    \caption{Results for the category car. Each block from left to right, input image, our proposed HSP, LR Soft, LR Hard.}
    \label{fig:car1}
  \end{figure*}
  \begin{figure*}
    \includegraphics[width=0.115\linewidth, trim={\trimI cm \trimI cm \trimI cm \trimI cm}, clip]{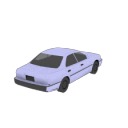}
    \includegraphics[width=0.115\linewidth, trim={\trimR cm \trimR cm \trimR cm \trimR cm}, clip]{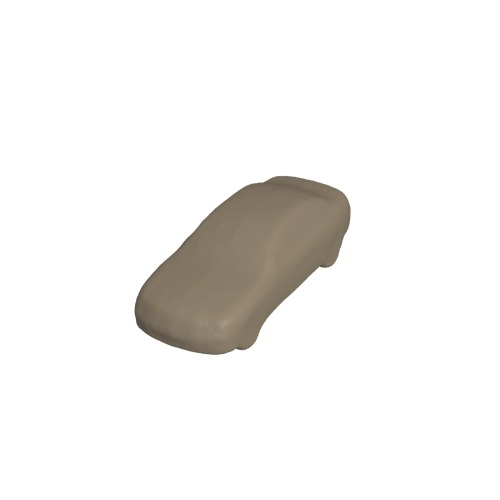}
    \includegraphics[width=0.115\linewidth, trim={\trimR cm \trimR cm \trimR cm \trimR cm}, clip]{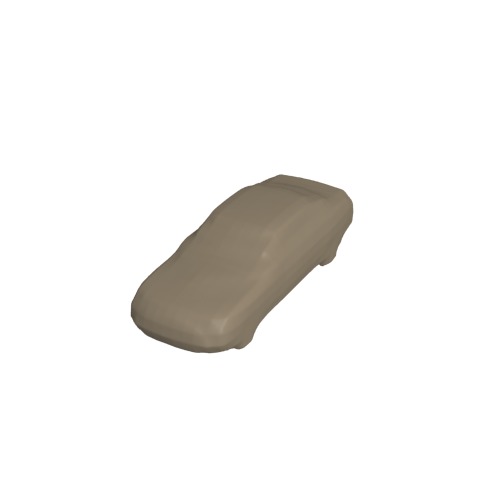}
    \includegraphics[width=0.115\linewidth, trim={\trimR cm \trimR cm \trimR cm \trimR cm}, clip]{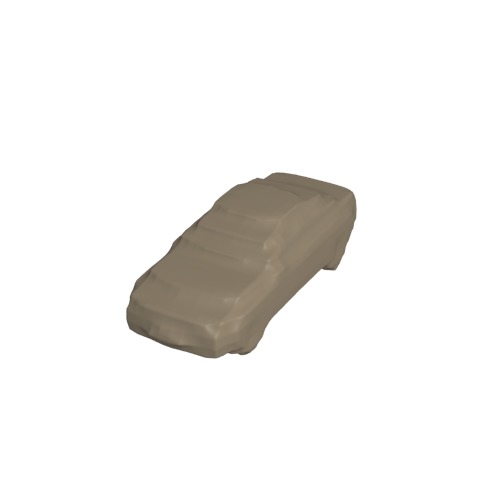}\hfill
    \includegraphics[width=0.115\linewidth, trim={\trimI cm \trimI cm \trimI cm \trimI cm}, clip]{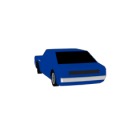}
    \includegraphics[width=0.115\linewidth, trim={\trimR cm \trimR cm \trimR cm \trimR cm}, clip]{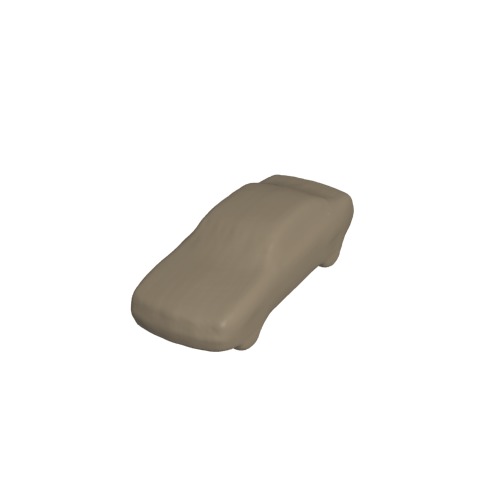}
    \includegraphics[width=0.115\linewidth, trim={\trimR cm \trimR cm \trimR cm \trimR cm}, clip]{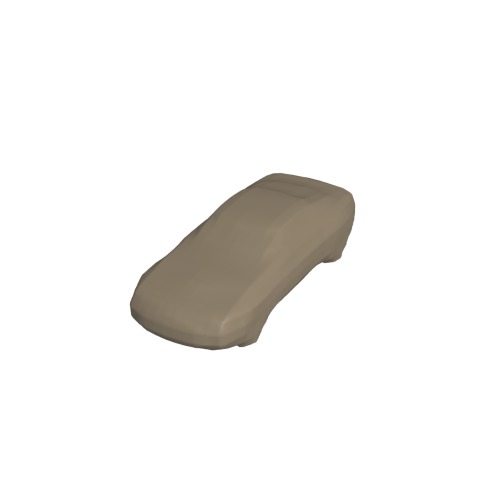}
    \includegraphics[width=0.115\linewidth, trim={\trimR cm \trimR cm \trimR cm \trimR cm}, clip]{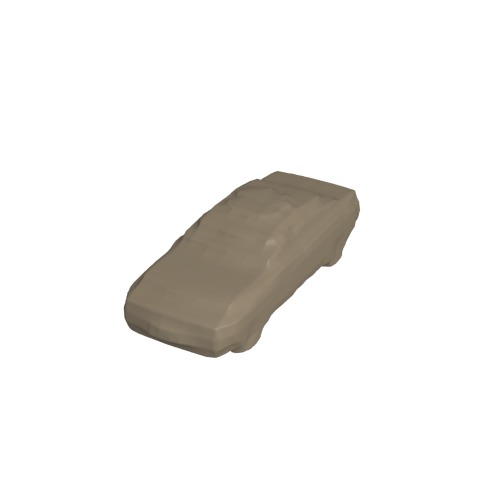} \\
      \includegraphics[width=0.115\linewidth, trim={\trimI cm \trimI cm \trimI cm \trimI cm}, clip]{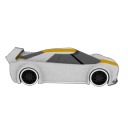}
    \includegraphics[width=0.115\linewidth, trim={\trimR cm \trimR cm \trimR cm \trimR cm}, clip]{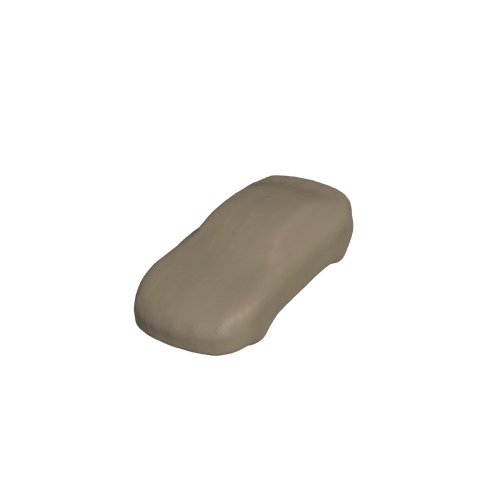}
    \includegraphics[width=0.115\linewidth, trim={\trimR cm \trimR cm \trimR cm \trimR cm}, clip]{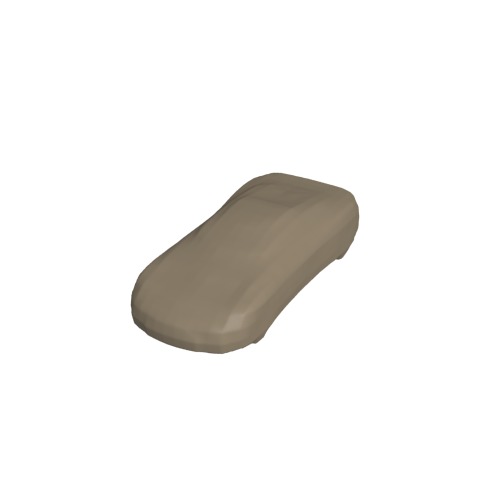}
    \includegraphics[width=0.115\linewidth, trim={\trimR cm \trimR cm \trimR cm \trimR cm}, clip]{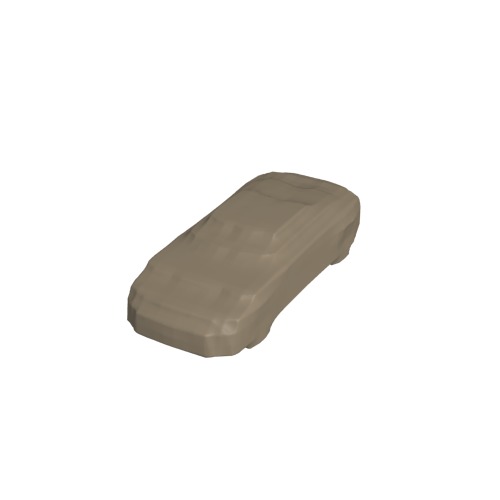}\hfill
    \includegraphics[width=0.115\linewidth, trim={\trimI cm \trimI cm \trimI cm \trimI cm}, clip]{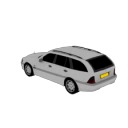}
    \includegraphics[width=0.115\linewidth, trim={\trimR cm \trimR cm \trimR cm \trimR cm}, clip]{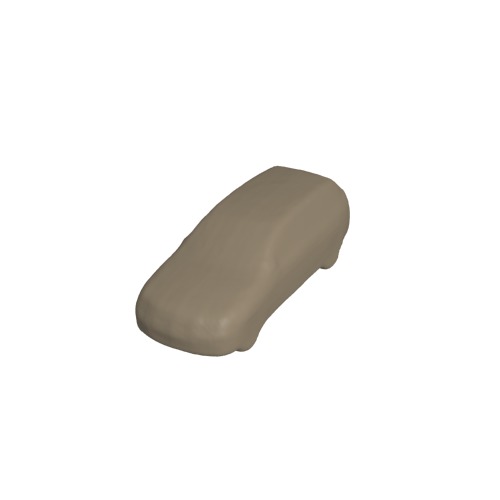}
    \includegraphics[width=0.115\linewidth, trim={\trimR cm \trimR cm \trimR cm \trimR cm}, clip]{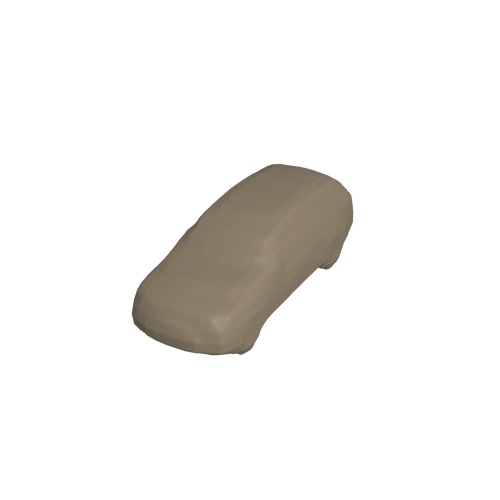}
    \includegraphics[width=0.115\linewidth, trim={\trimR cm \trimR cm \trimR cm \trimR cm}, clip]{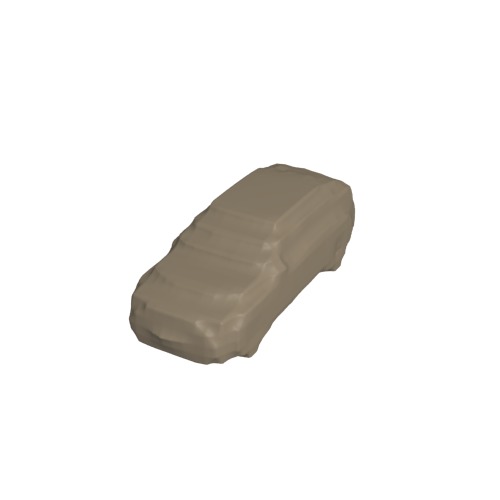} \\
    \includegraphics[width=0.115\linewidth, trim={\trimI cm \trimI cm \trimI cm \trimI cm}, clip]{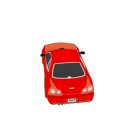}
    \includegraphics[width=0.115\linewidth, trim={\trimR cm \trimR cm \trimR cm \trimR cm}, clip]{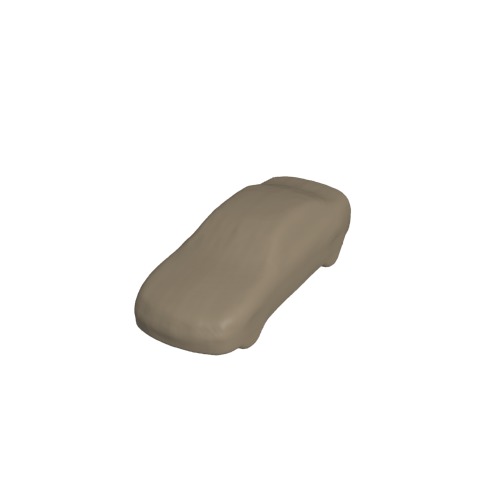}
    \includegraphics[width=0.115\linewidth, trim={\trimR cm \trimR cm \trimR cm \trimR cm}, clip]{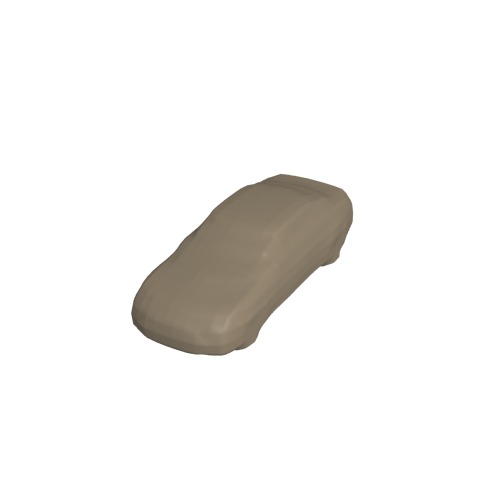}
    \includegraphics[width=0.115\linewidth, trim={\trimR cm \trimR cm \trimR cm \trimR cm}, clip]{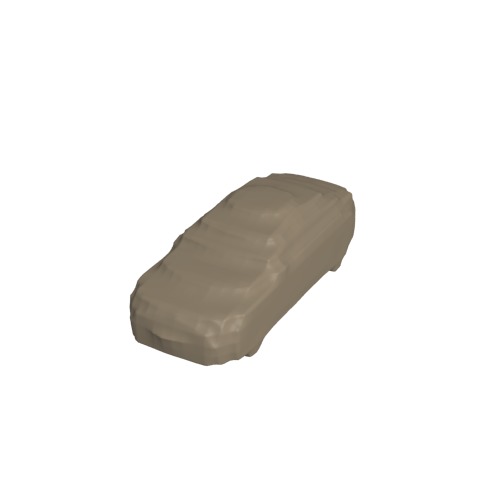}\hfill
    \includegraphics[width=0.115\linewidth, trim={\trimI cm \trimI cm \trimI cm \trimI cm}, clip]{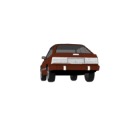}
    \includegraphics[width=0.115\linewidth, trim={\trimR cm \trimR cm \trimR cm \trimR cm}, clip]{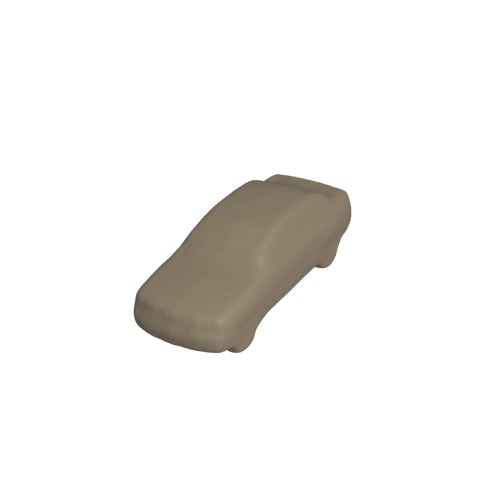}
    \includegraphics[width=0.115\linewidth, trim={\trimR cm \trimR cm \trimR cm \trimR cm}, clip]{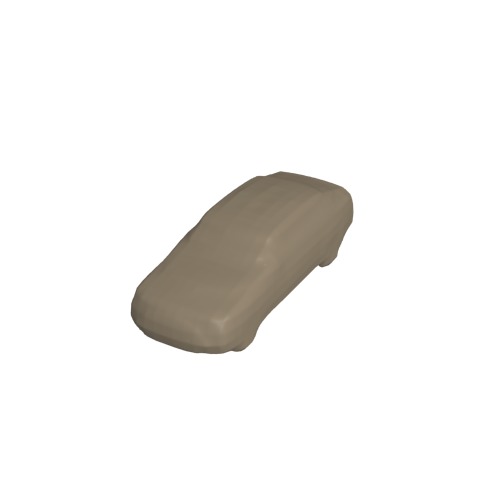}
    \includegraphics[width=0.115\linewidth, trim={\trimR cm \trimR cm \trimR cm \trimR cm}, clip]{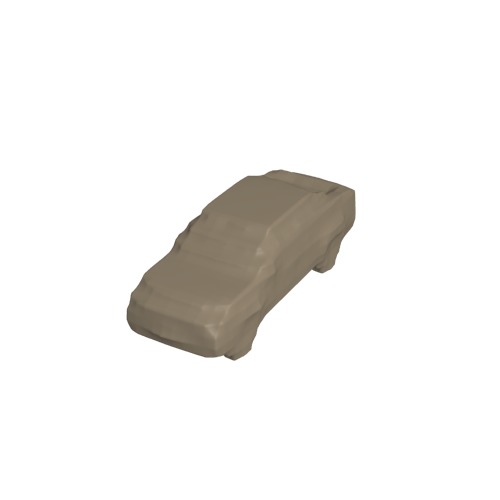}\\
    \includegraphics[width=0.115\linewidth, trim={\trimI cm \trimI cm \trimI cm \trimI cm}, clip]{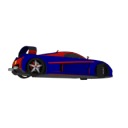}
    \includegraphics[width=0.115\linewidth, trim={\trimR cm \trimR cm \trimR cm \trimR cm}, clip]{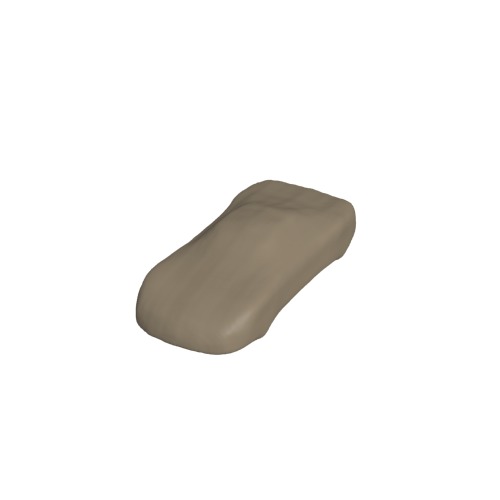}
    \includegraphics[width=0.115\linewidth, trim={\trimR cm \trimR cm \trimR cm \trimR cm}, clip]{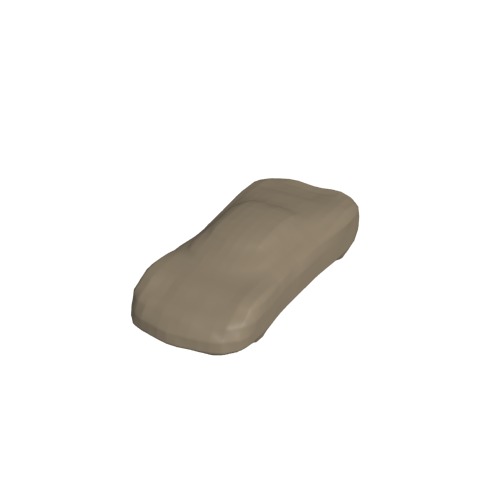}
    \includegraphics[width=0.115\linewidth, trim={\trimR cm \trimR cm \trimR cm \trimR cm}, clip]{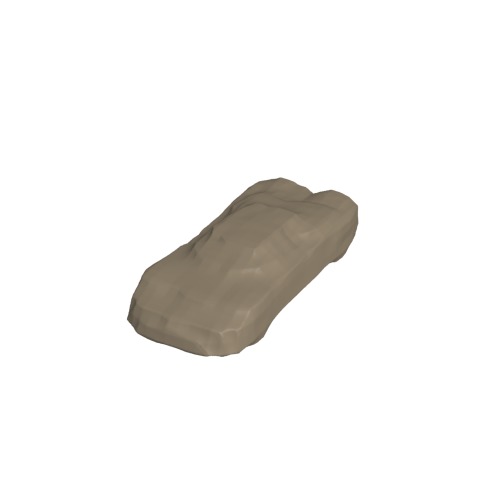}\hfill
    \includegraphics[width=0.115\linewidth, trim={\trimI cm \trimI cm \trimI cm \trimI cm}, clip]{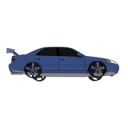}
    \includegraphics[width=0.115\linewidth, trim={\trimR cm \trimR cm \trimR cm \trimR cm}, clip]{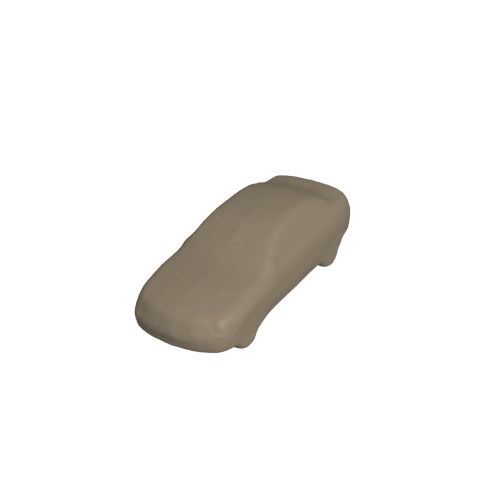}
    \includegraphics[width=0.115\linewidth, trim={\trimR cm \trimR cm \trimR cm \trimR cm}, clip]{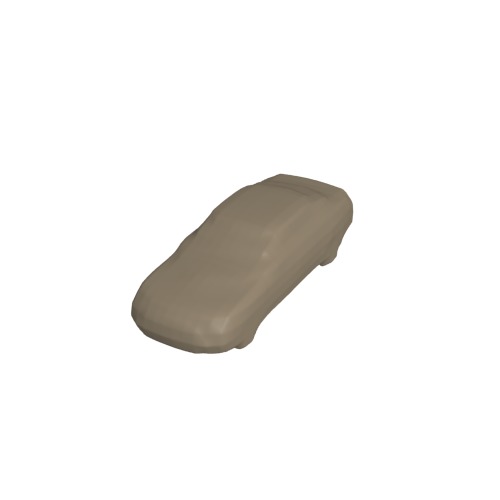}
    \includegraphics[width=0.115\linewidth, trim={\trimR cm \trimR cm \trimR cm \trimR cm}, clip]{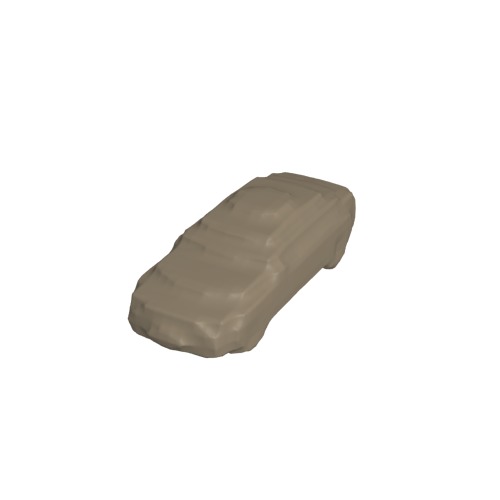}\\
    \includegraphics[width=0.115\linewidth, trim={\trimI cm \trimI cm \trimI cm \trimI cm}, clip]{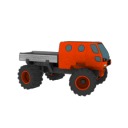}
    \includegraphics[width=0.115\linewidth, trim={\trimR cm \trimR cm \trimR cm \trimR cm}, clip]{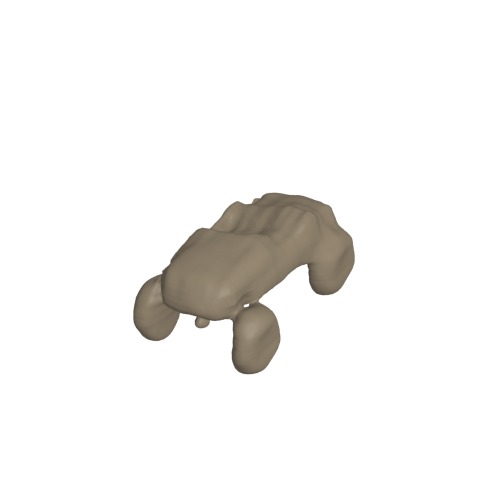}
    \includegraphics[width=0.115\linewidth, trim={\trimR cm \trimR cm \trimR cm \trimR cm}, clip]{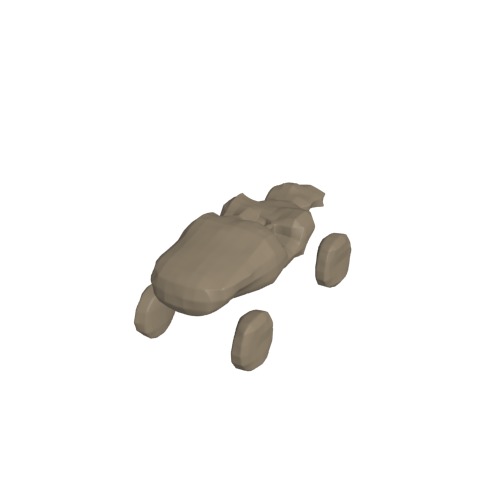}
    \includegraphics[width=0.115\linewidth, trim={\trimR cm \trimR cm \trimR cm \trimR cm}, clip]{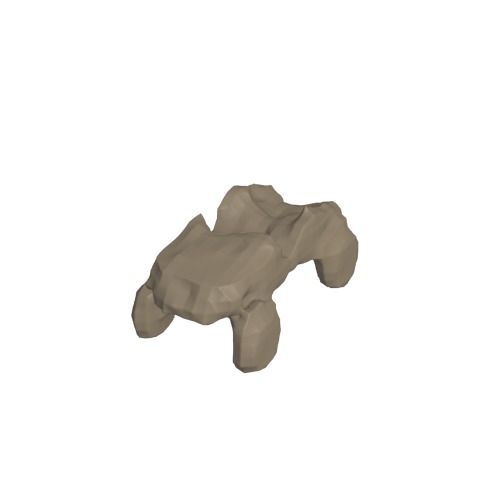}\hfill
    \includegraphics[width=0.115\linewidth, trim={\trimI cm \trimI cm \trimI cm \trimI cm}, clip]{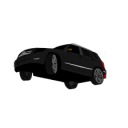}
    \includegraphics[width=0.115\linewidth, trim={\trimR cm \trimR cm \trimR cm \trimR cm}, clip]{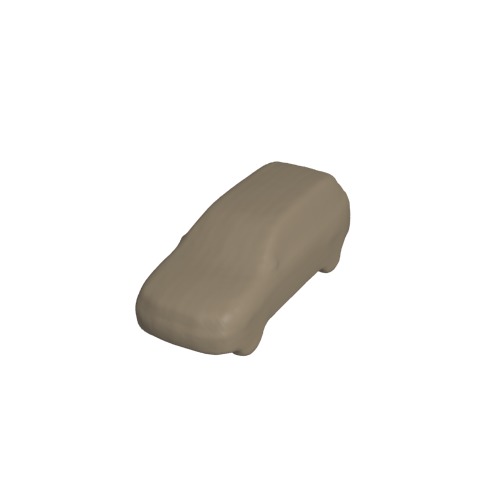}
    \includegraphics[width=0.115\linewidth, trim={\trimR cm \trimR cm \trimR cm \trimR cm}, clip]{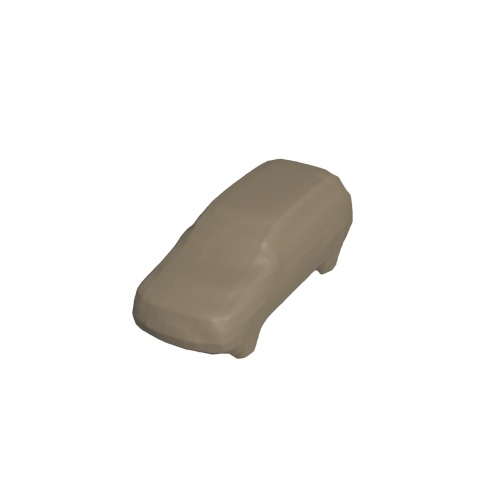}
    \includegraphics[width=0.115\linewidth, trim={\trimR cm \trimR cm \trimR cm \trimR cm}, clip]{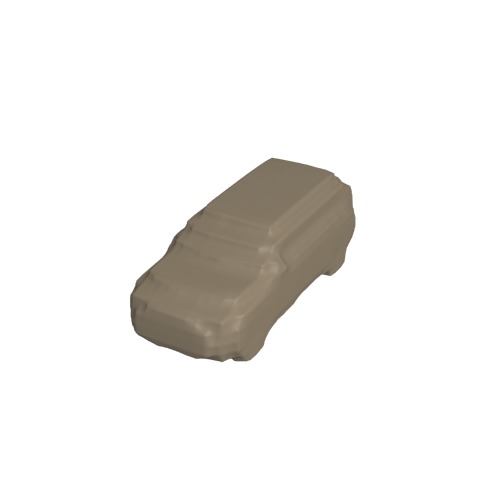}\\
    \includegraphics[width=0.115\linewidth, trim={\trimI cm \trimI cm \trimI cm \trimI cm}, clip]{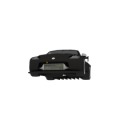}
    \includegraphics[width=0.115\linewidth, trim={\trimR cm \trimR cm \trimR cm \trimR cm}, clip]{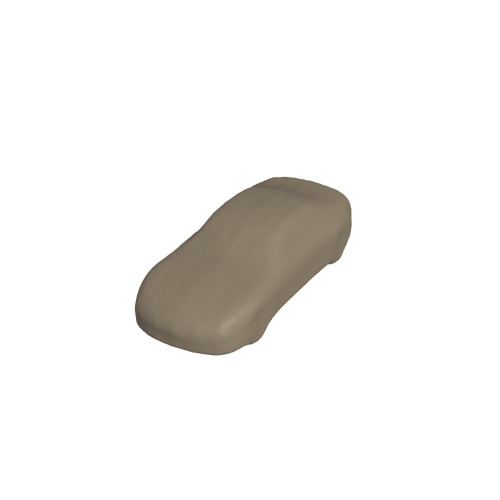}
    \includegraphics[width=0.115\linewidth, trim={\trimR cm \trimR cm \trimR cm \trimR cm}, clip]{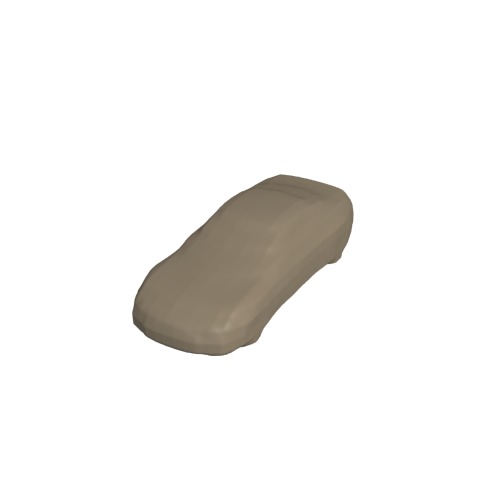}
    \includegraphics[width=0.115\linewidth, trim={\trimR cm \trimR cm \trimR cm \trimR cm}, clip]{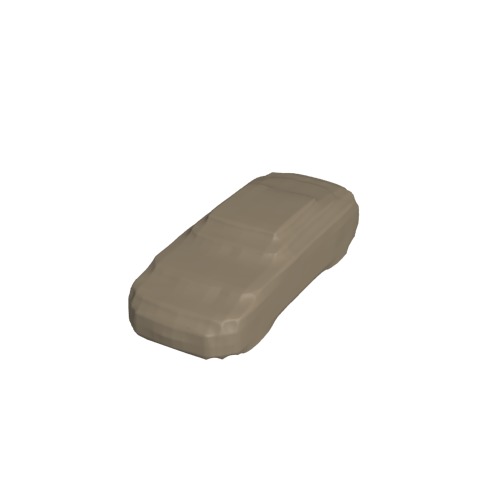}\hfill
    \includegraphics[width=0.115\linewidth, trim={\trimI cm \trimI cm \trimI cm \trimI cm}, clip]{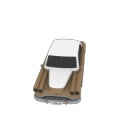}
    \includegraphics[width=0.115\linewidth, trim={\trimR cm \trimR cm \trimR cm \trimR cm}, clip]{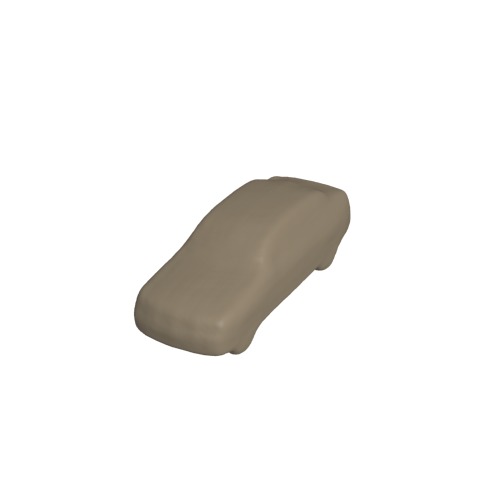}
    \includegraphics[width=0.115\linewidth, trim={\trimR cm \trimR cm \trimR cm \trimR cm}, clip]{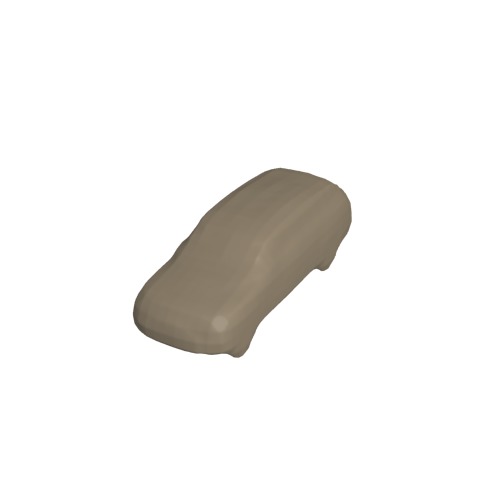}
    \includegraphics[width=0.115\linewidth, trim={\trimR cm \trimR cm \trimR cm \trimR cm}, clip]{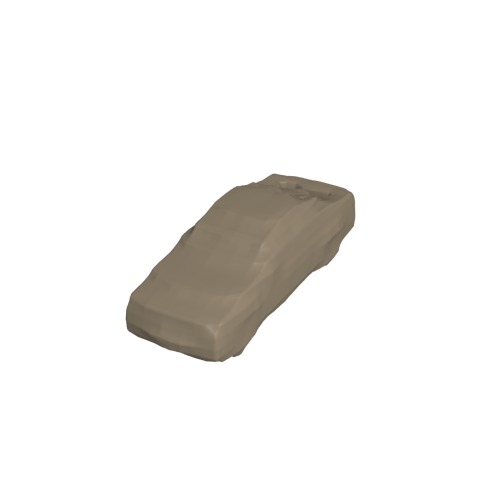}\\
    \includegraphics[width=0.115\linewidth, trim={\trimI cm \trimI cm \trimI cm \trimI cm}, clip]{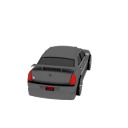}
    \includegraphics[width=0.115\linewidth, trim={\trimR cm \trimR cm \trimR cm \trimR cm}, clip]{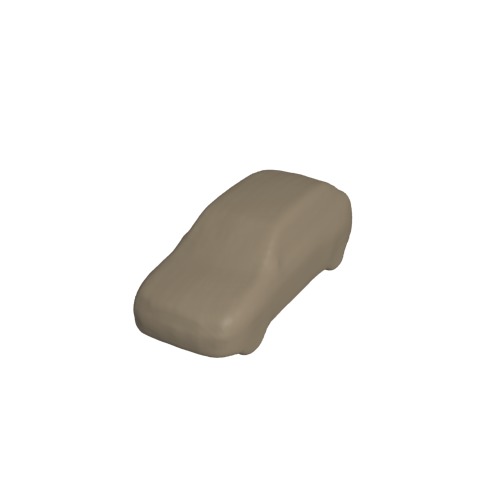}
    \includegraphics[width=0.115\linewidth, trim={\trimR cm \trimR cm \trimR cm \trimR cm}, clip]{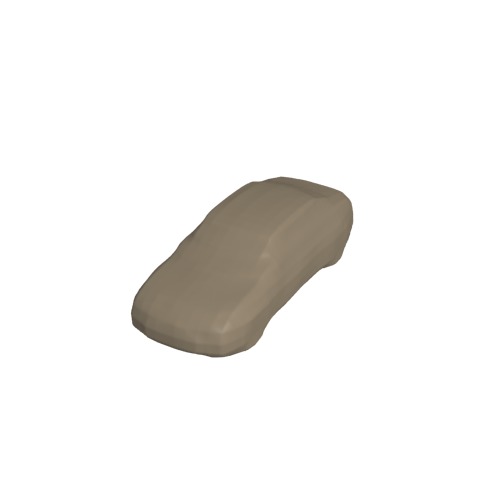}
    \includegraphics[width=0.115\linewidth, trim={\trimR cm \trimR cm \trimR cm \trimR cm}, clip]{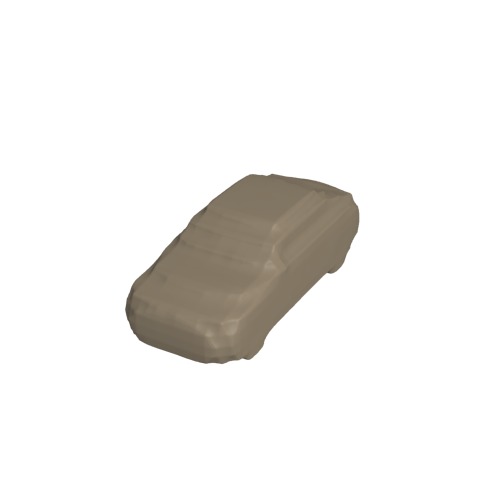}\hfill
    \includegraphics[width=0.115\linewidth, trim={\trimI cm \trimI cm \trimI cm \trimI cm}, clip]{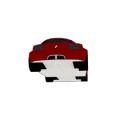}
    \includegraphics[width=0.115\linewidth, trim={\trimR cm \trimR cm \trimR cm \trimR cm}, clip]{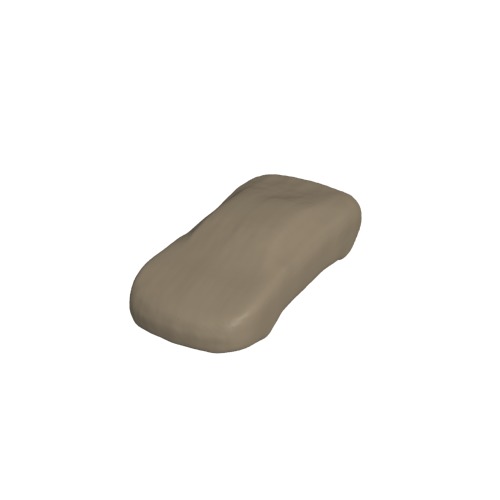}
    \includegraphics[width=0.115\linewidth, trim={\trimR cm \trimR cm \trimR cm \trimR cm}, clip]{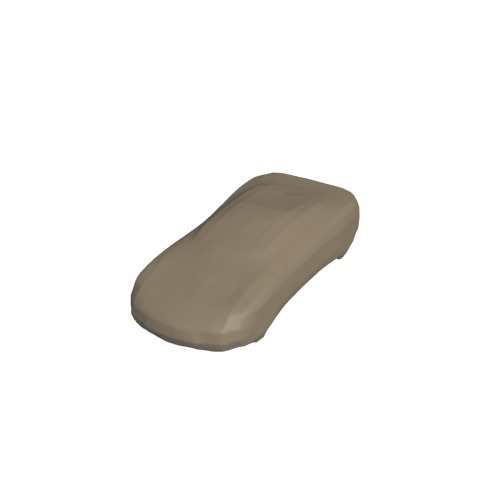}
    \includegraphics[width=0.115\linewidth, trim={\trimR cm \trimR cm \trimR cm \trimR cm}, clip]{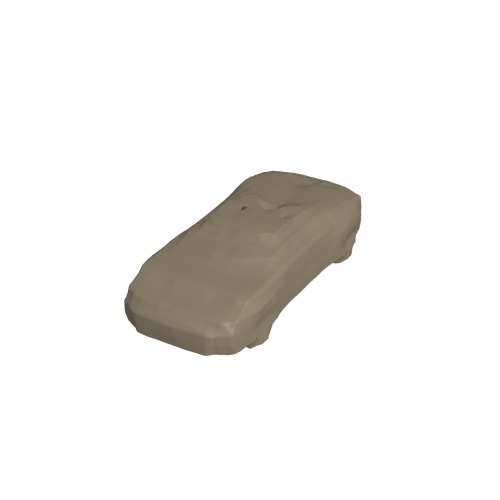}\\
    \includegraphics[width=0.115\linewidth, trim={\trimI cm \trimI cm \trimI cm \trimI cm}, clip]{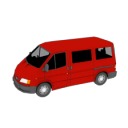}
    \includegraphics[width=0.115\linewidth, trim={\trimR cm \trimR cm \trimR cm \trimR cm}, clip]{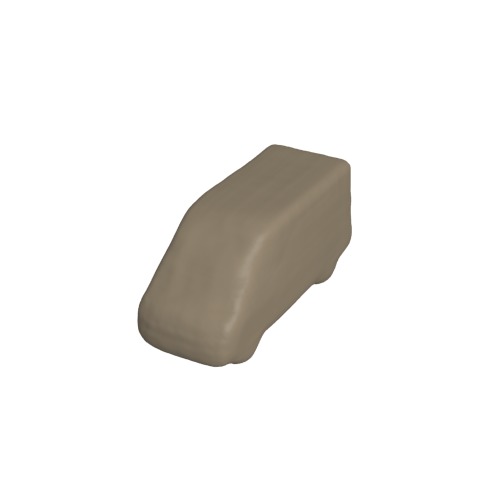}
    \includegraphics[width=0.115\linewidth, trim={\trimR cm \trimR cm \trimR cm \trimR cm}, clip]{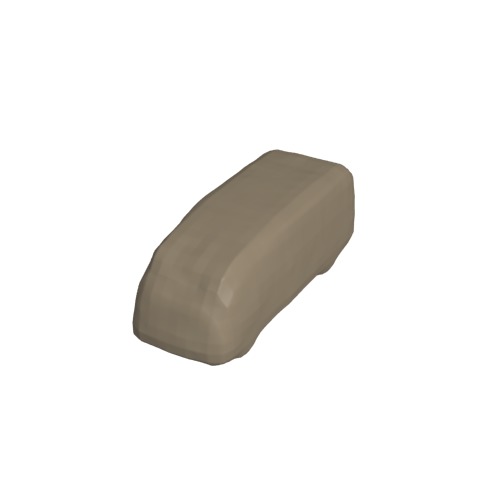}
    \includegraphics[width=0.115\linewidth, trim={\trimR cm \trimR cm \trimR cm \trimR cm}, clip]{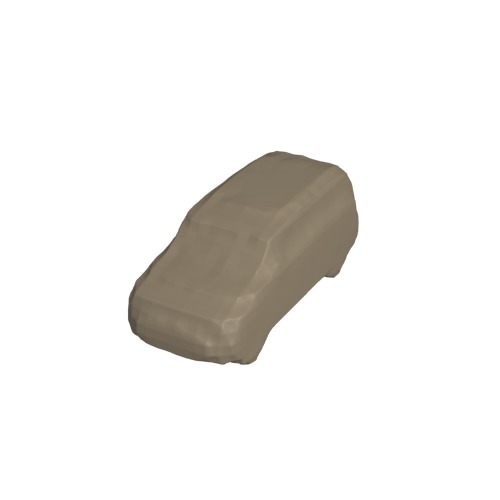}\hfill
    \includegraphics[width=0.115\linewidth, trim={\trimI cm \trimI cm \trimI cm \trimI cm}, clip]{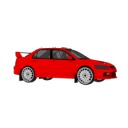}
    \includegraphics[width=0.115\linewidth, trim={\trimR cm \trimR cm \trimR cm \trimR cm}, clip]{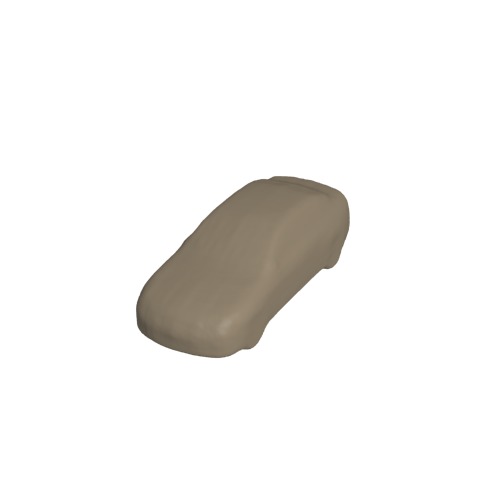}
    \includegraphics[width=0.115\linewidth, trim={\trimR cm \trimR cm \trimR cm \trimR cm}, clip]{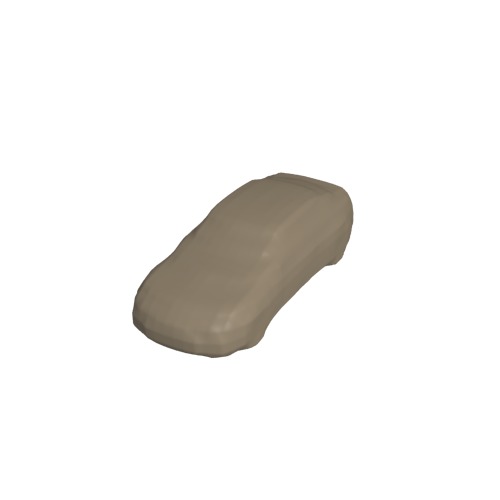}
    \includegraphics[width=0.115\linewidth, trim={\trimR cm \trimR cm \trimR cm \trimR cm}, clip]{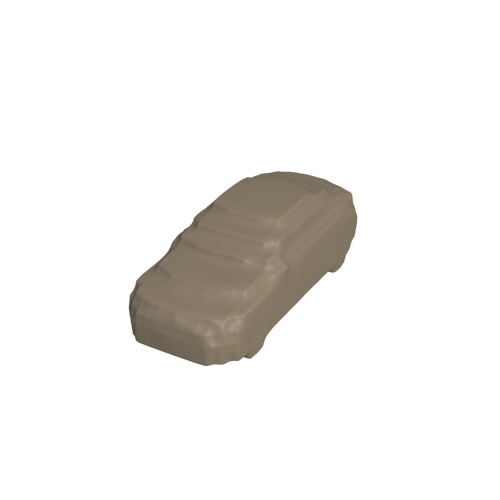}\\
    \includegraphics[width=0.115\linewidth, trim={\trimI cm \trimI cm \trimI cm \trimI cm}, clip]{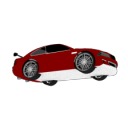}
    \includegraphics[width=0.115\linewidth, trim={\trimR cm \trimR cm \trimR cm \trimR cm}, clip]{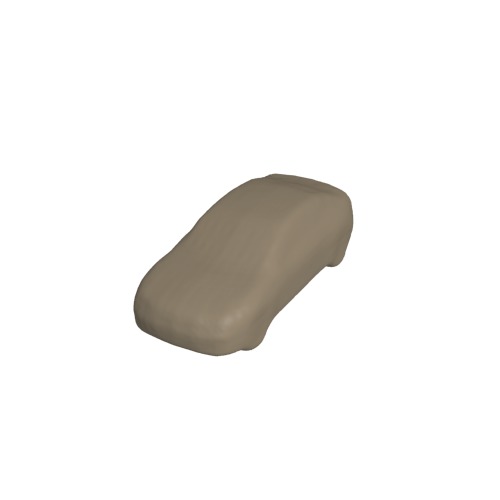}
    \includegraphics[width=0.115\linewidth, trim={\trimR cm \trimR cm \trimR cm \trimR cm}, clip]{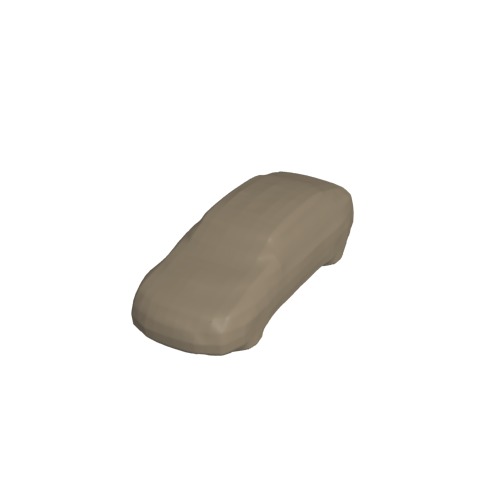}
    \includegraphics[width=0.115\linewidth, trim={\trimR cm \trimR cm \trimR cm \trimR cm}, clip]{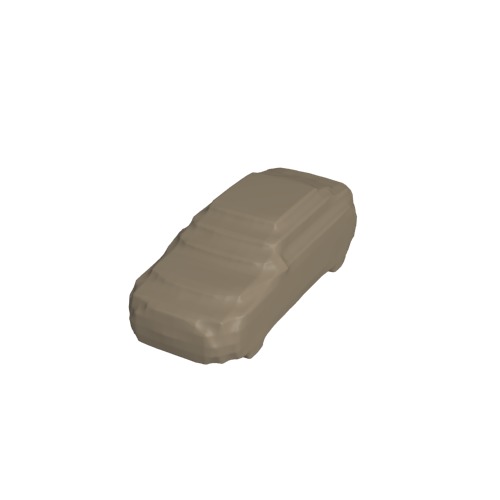}\hfill
    \includegraphics[width=0.115\linewidth, trim={\trimI cm \trimI cm \trimI cm \trimI cm}, clip]{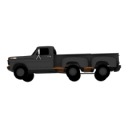}
    \includegraphics[width=0.115\linewidth, trim={\trimR cm \trimR cm \trimR cm \trimR cm}, clip]{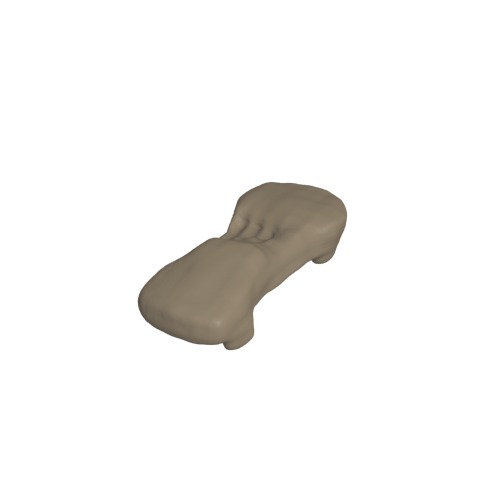}
    \includegraphics[width=0.115\linewidth, trim={\trimR cm \trimR cm \trimR cm \trimR cm}, clip]{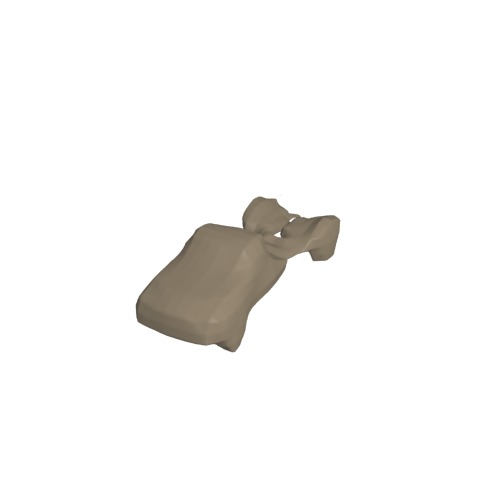}
    \includegraphics[width=0.115\linewidth, trim={\trimR cm \trimR cm \trimR cm \trimR cm}, clip]{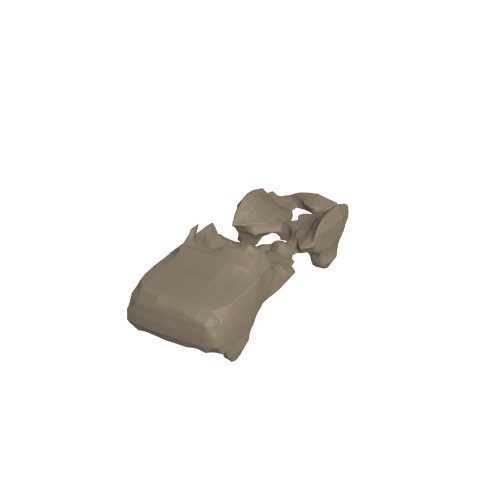}\\
    \includegraphics[width=0.115\linewidth, trim={\trimI cm \trimI cm \trimI cm \trimI cm}, clip]{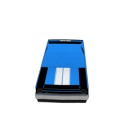}
    \includegraphics[width=0.115\linewidth, trim={\trimR cm \trimR cm \trimR cm \trimR cm}, clip]{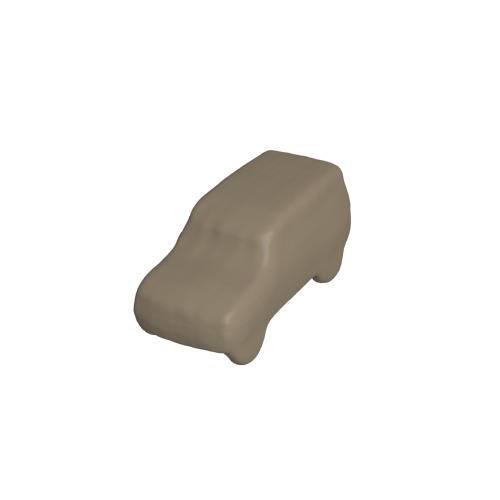}
    \includegraphics[width=0.115\linewidth, trim={\trimR cm \trimR cm \trimR cm \trimR cm}, clip]{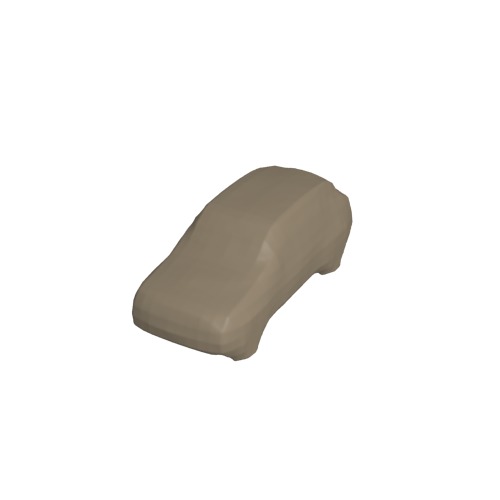}
    \includegraphics[width=0.115\linewidth, trim={\trimR cm \trimR cm \trimR cm \trimR cm}, clip]{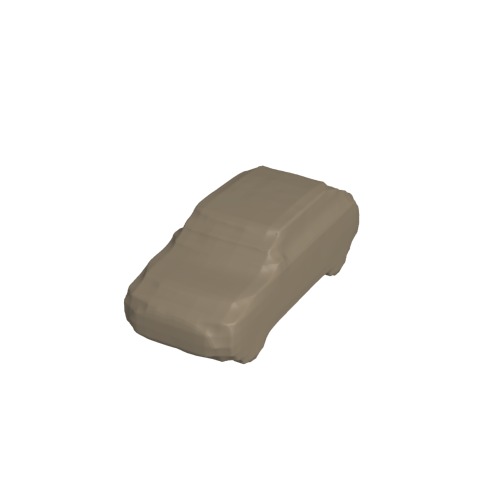}\hfill
    \includegraphics[width=0.115\linewidth, trim={\trimI cm \trimI cm \trimI cm \trimI cm}, clip]{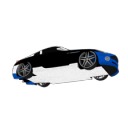}
    \includegraphics[width=0.115\linewidth, trim={\trimR cm \trimR cm \trimR cm \trimR cm}, clip]{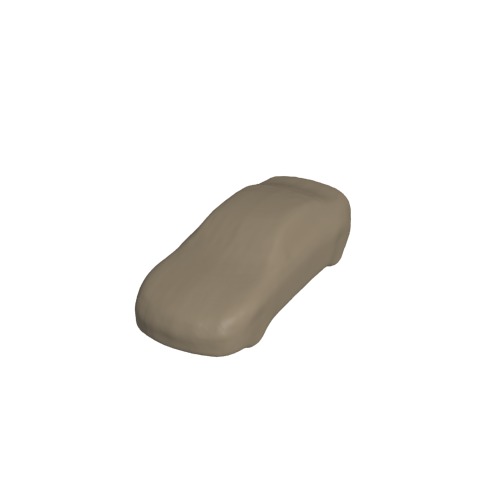}
    \includegraphics[width=0.115\linewidth, trim={\trimR cm \trimR cm \trimR cm \trimR cm}, clip]{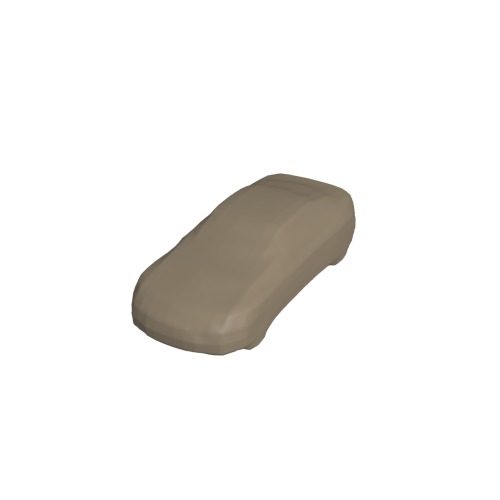}
    \includegraphics[width=0.115\linewidth, trim={\trimR cm \trimR cm \trimR cm \trimR cm}, clip]{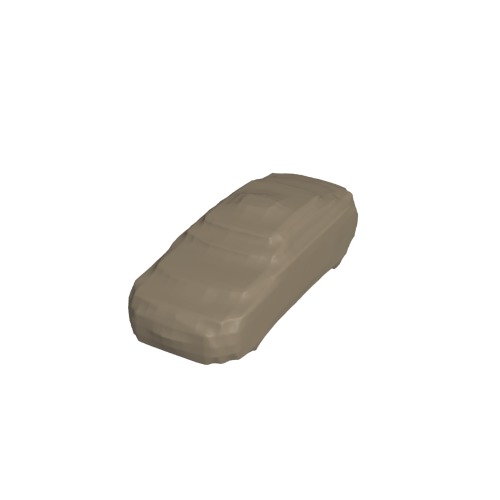}
    \caption{Results for the category car continued. Each block from left to right, input image, our proposed HSP, LR Soft, LR Hard.}
    \label{fig:car2}
  \end{figure*}
  \begin{figure*}
    \includegraphics[width=0.115\linewidth, trim={\trimI cm \trimI cm \trimI cm \trimI cm}, clip]{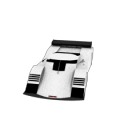}
    \includegraphics[width=0.115\linewidth, trim={\trimR cm \trimR cm \trimR cm \trimR cm}, clip]{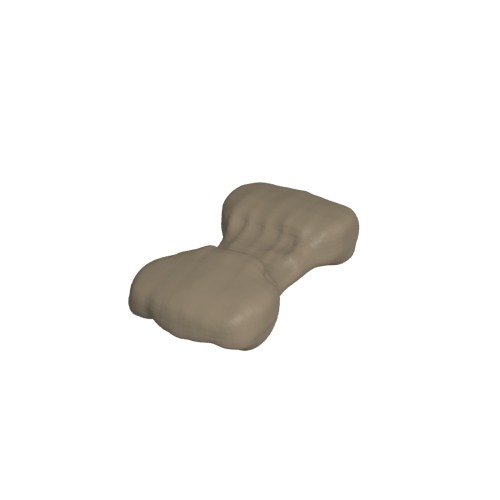}
    \includegraphics[width=0.115\linewidth, trim={\trimR cm \trimR cm \trimR cm \trimR cm}, clip]{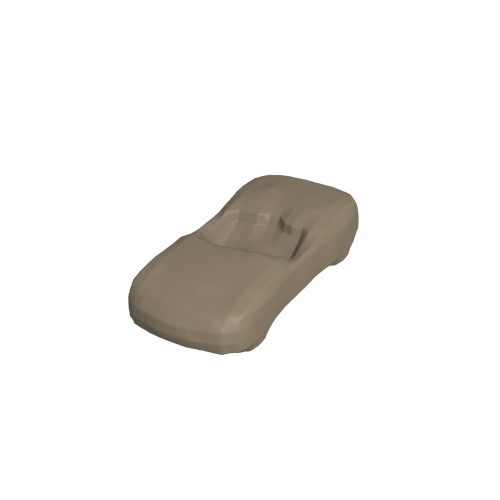}
    \includegraphics[width=0.115\linewidth, trim={\trimR cm \trimR cm \trimR cm \trimR cm}, clip]{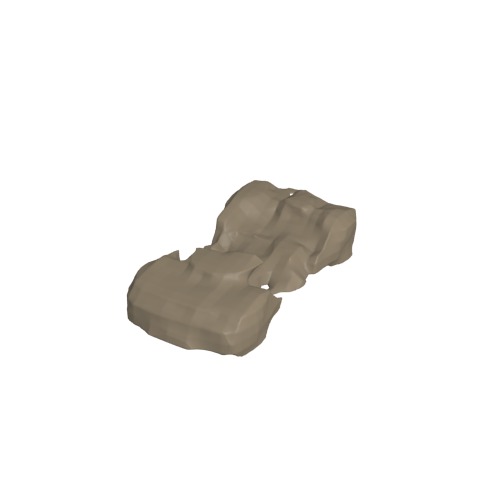}\hfill
    \includegraphics[width=0.115\linewidth, trim={\trimI cm \trimI cm \trimI cm \trimI cm}, clip]{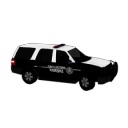}
    \includegraphics[width=0.115\linewidth, trim={\trimR cm \trimR cm \trimR cm \trimR cm}, clip]{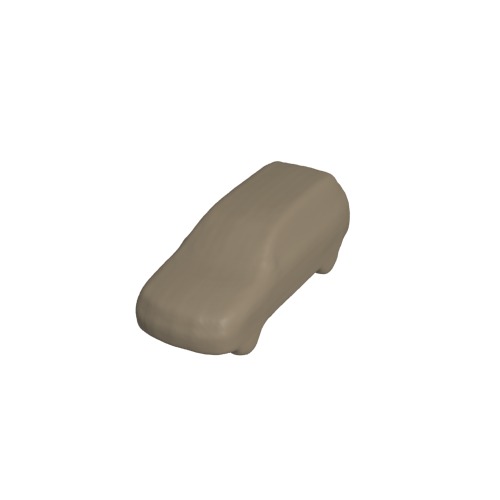}
    \includegraphics[width=0.115\linewidth, trim={\trimR cm \trimR cm \trimR cm \trimR cm}, clip]{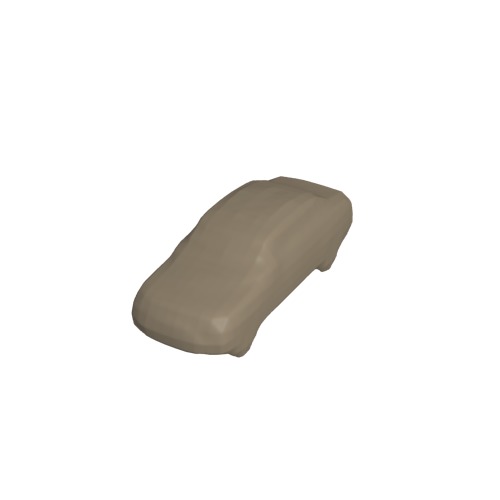}
    \includegraphics[width=0.115\linewidth, trim={\trimR cm \trimR cm \trimR cm \trimR cm}, clip]{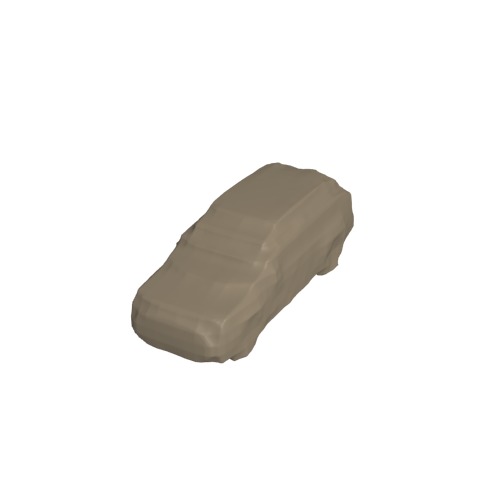}\\
    \includegraphics[width=0.115\linewidth, trim={\trimI cm \trimI cm \trimI cm \trimI cm}, clip]{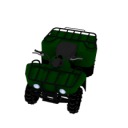}
    \includegraphics[width=0.115\linewidth, trim={\trimR cm \trimR cm \trimR cm \trimR cm}, clip]{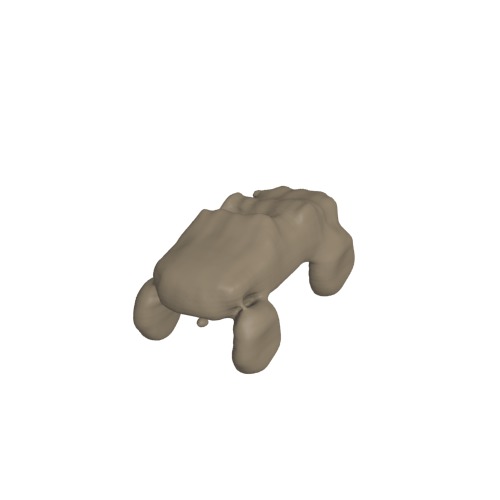}
    \includegraphics[width=0.115\linewidth, trim={\trimR cm \trimR cm \trimR cm \trimR cm}, clip]{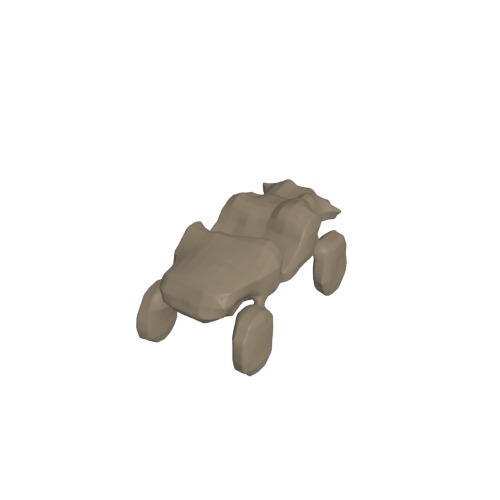}
    \includegraphics[width=0.115\linewidth, trim={\trimR cm \trimR cm \trimR cm \trimR cm}, clip]{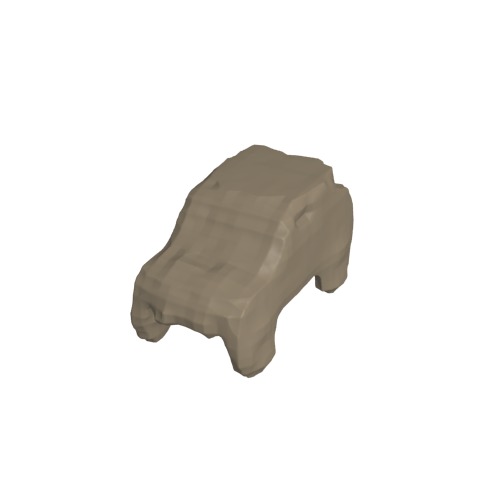}\hfill
    \includegraphics[width=0.115\linewidth, trim={\trimI cm \trimI cm \trimI cm \trimI cm}, clip]{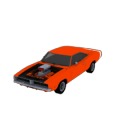}
    \includegraphics[width=0.115\linewidth, trim={\trimR cm \trimR cm \trimR cm \trimR cm}, clip]{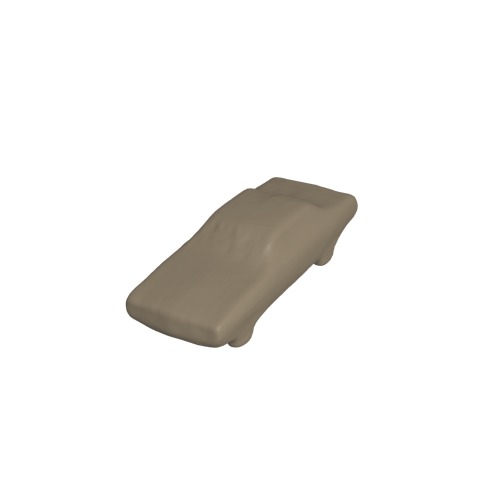}
    \includegraphics[width=0.115\linewidth, trim={\trimR cm \trimR cm \trimR cm \trimR cm}, clip]{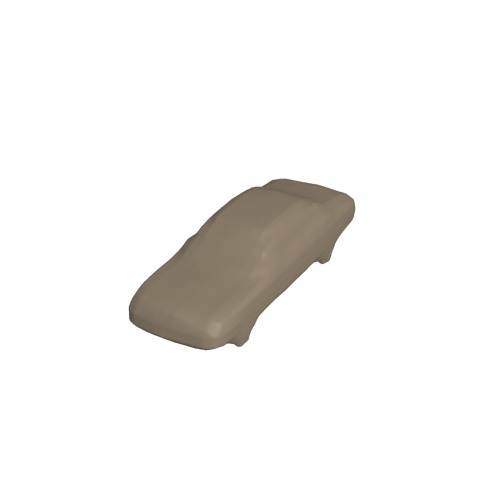}
    \includegraphics[width=0.115\linewidth, trim={\trimR cm \trimR cm \trimR cm \trimR cm}, clip]{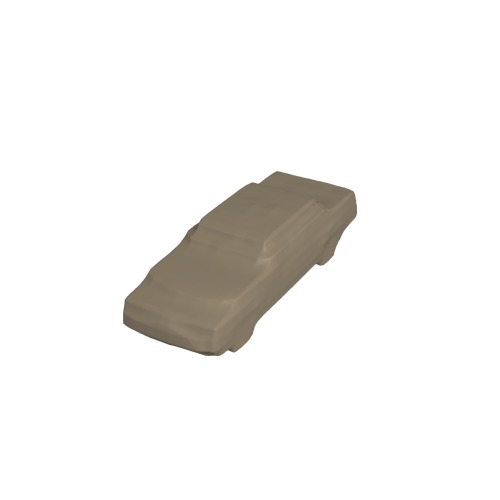}\\
    \includegraphics[width=0.115\linewidth, trim={\trimI cm \trimI cm \trimI cm \trimI cm}, clip]{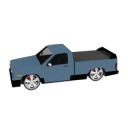}
    \includegraphics[width=0.115\linewidth, trim={\trimR cm \trimR cm \trimR cm \trimR cm}, clip]{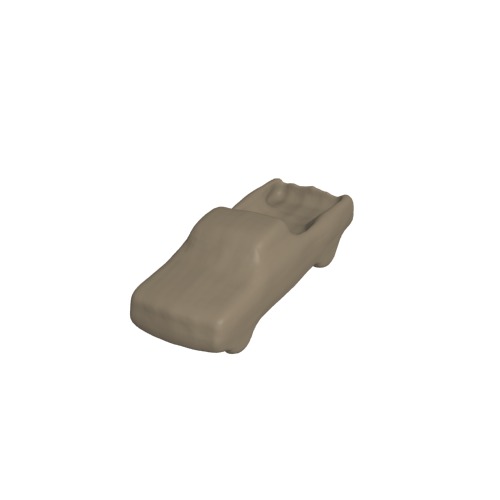}
    \includegraphics[width=0.115\linewidth, trim={\trimR cm \trimR cm \trimR cm \trimR cm}, clip]{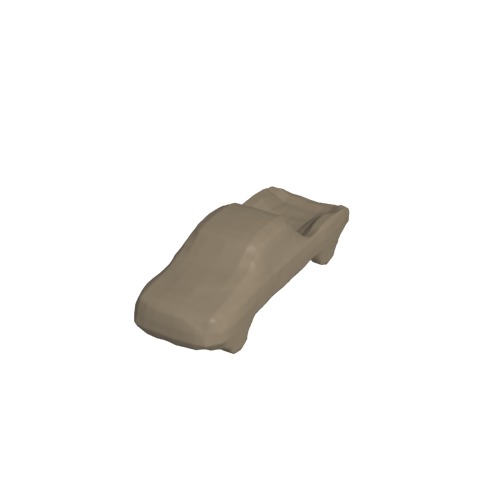}
    \includegraphics[width=0.115\linewidth, trim={\trimR cm \trimR cm \trimR cm \trimR cm}, clip]{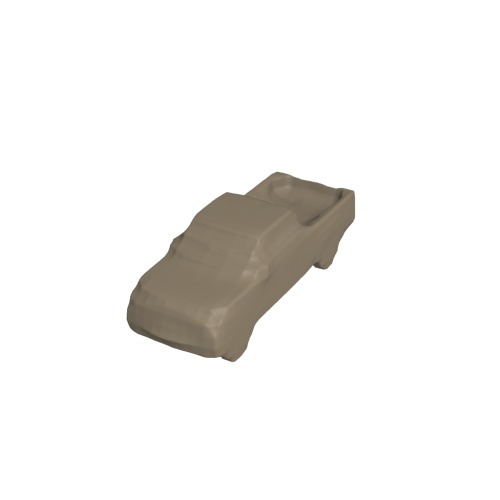}\hfill
    \includegraphics[width=0.115\linewidth, trim={\trimI cm \trimI cm \trimI cm \trimI cm}, clip]{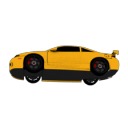}
    \includegraphics[width=0.115\linewidth, trim={\trimR cm \trimR cm \trimR cm \trimR cm}, clip]{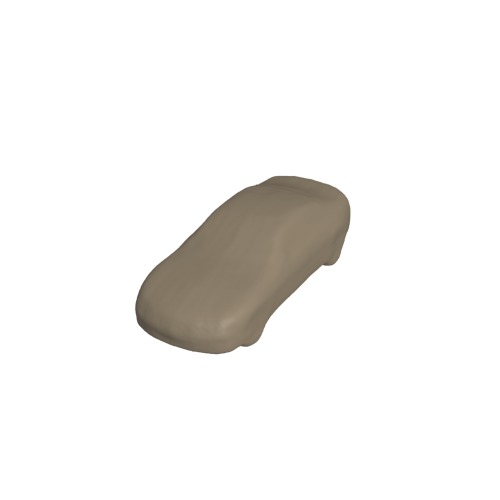}
    \includegraphics[width=0.115\linewidth, trim={\trimR cm \trimR cm \trimR cm \trimR cm}, clip]{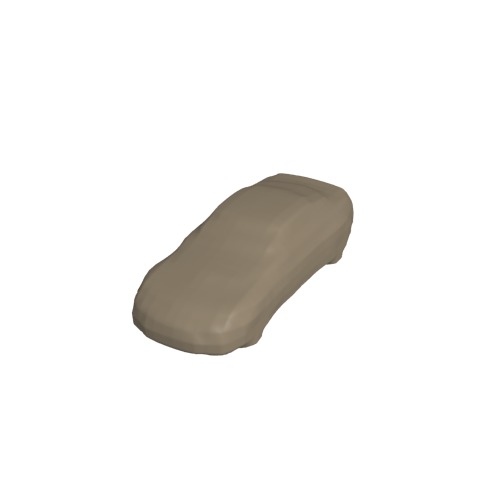}
    \includegraphics[width=0.115\linewidth, trim={\trimR cm \trimR cm \trimR cm \trimR cm}, clip]{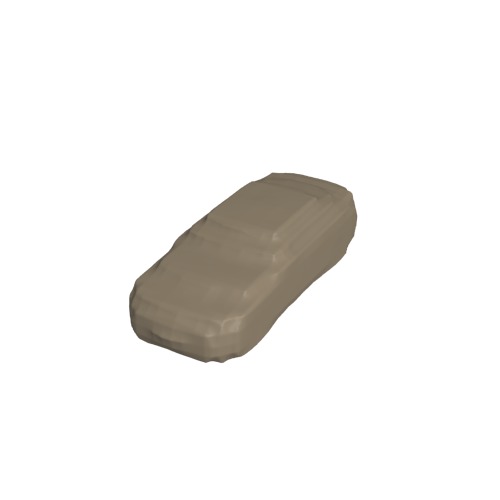}
    \caption{Results for the category car continued. Each block from left to right, input image, our proposed HSP, LR Soft, LR Hard.}
    \label{fig:car3}
  \end{figure*}

    \begin{figure*}
    \includegraphics[width=0.115\linewidth, trim={\trimI cm \trimI cm \trimI cm \trimI cm}, clip]{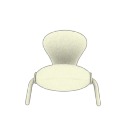}
    \includegraphics[width=0.115\linewidth, trim={\trimR cm \trimR cm \trimR cm \trimR cm}, clip]{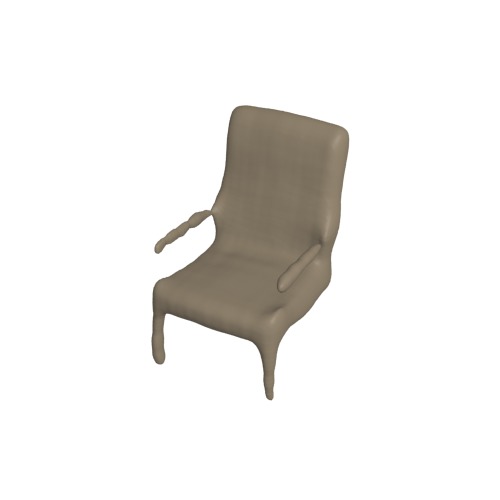}
    \includegraphics[width=0.115\linewidth, trim={\trimR cm \trimR cm \trimR cm \trimR cm}, clip]{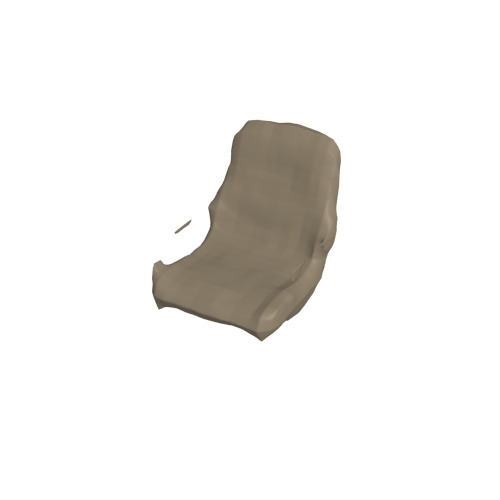}
    \includegraphics[width=0.115\linewidth, trim={\trimR cm \trimR cm \trimR cm \trimR cm}, clip]{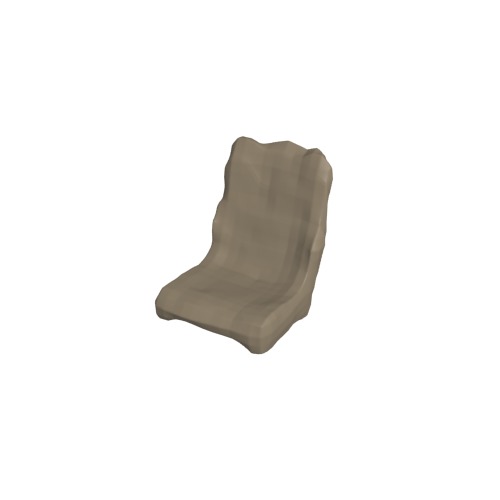}\hfill
    \includegraphics[width=0.115\linewidth, trim={\trimI cm \trimI cm \trimI cm \trimI cm}, clip]{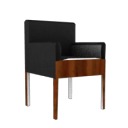}
    \includegraphics[width=0.115\linewidth, trim={\trimR cm \trimR cm \trimR cm \trimR cm}, clip]{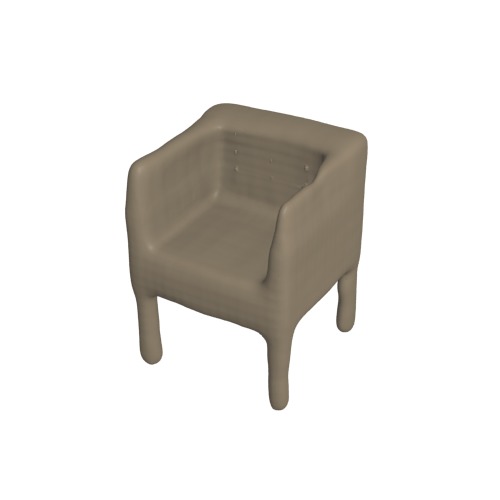}
    \includegraphics[width=0.115\linewidth, trim={\trimR cm \trimR cm \trimR cm \trimR cm}, clip]{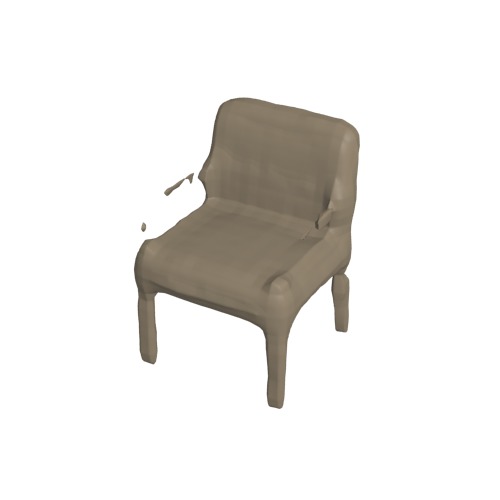}
    \includegraphics[width=0.115\linewidth, trim={\trimR cm \trimR cm \trimR cm \trimR cm}, clip]{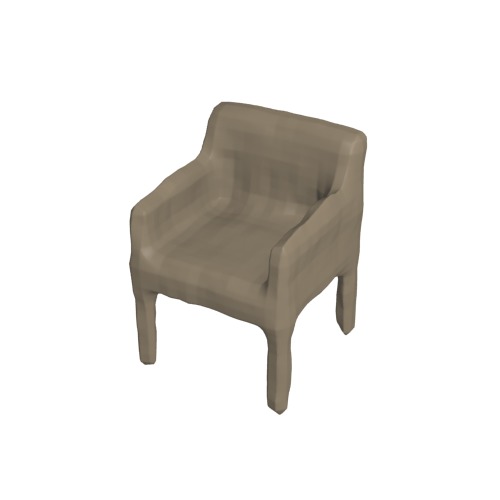}\\
    \includegraphics[width=0.115\linewidth, trim={\trimI cm \trimI cm \trimI cm \trimI cm}, clip]{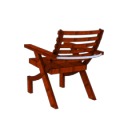}
    \includegraphics[width=0.115\linewidth, trim={\trimR cm \trimR cm \trimR cm \trimR cm}, clip]{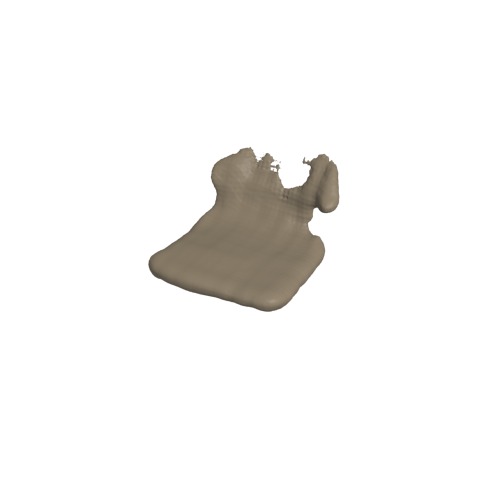}
    \includegraphics[width=0.115\linewidth, trim={\trimR cm \trimR cm \trimR cm \trimR cm}, clip]{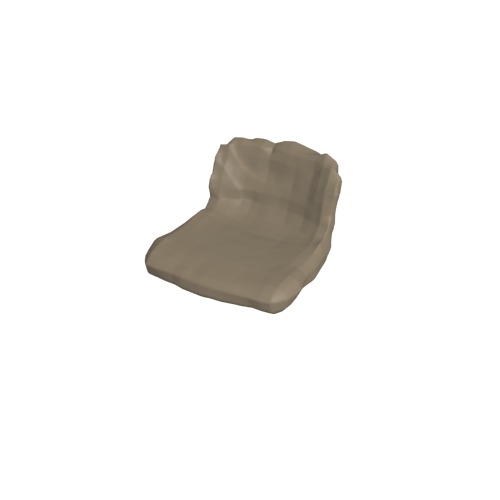}
    \includegraphics[width=0.115\linewidth, trim={\trimR cm \trimR cm \trimR cm \trimR cm}, clip]{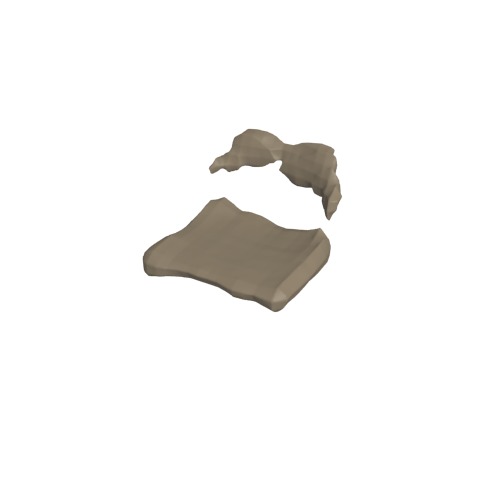}\hfill
    \includegraphics[width=0.115\linewidth, trim={\trimI cm \trimI cm \trimI cm \trimI cm}, clip]{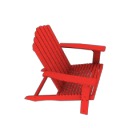}
    \includegraphics[width=0.115\linewidth, trim={\trimR cm \trimR cm \trimR cm \trimR cm}, clip]{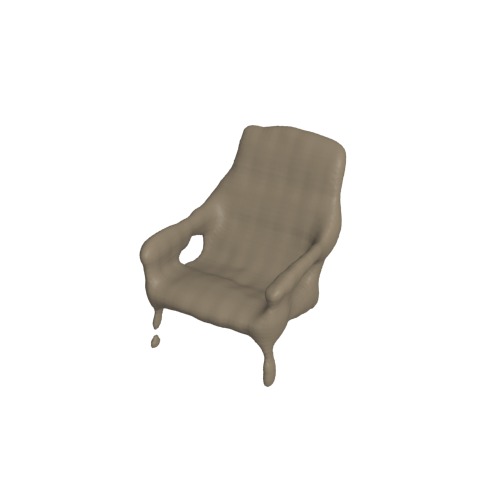}
    \includegraphics[width=0.115\linewidth, trim={\trimR cm \trimR cm \trimR cm \trimR cm}, clip]{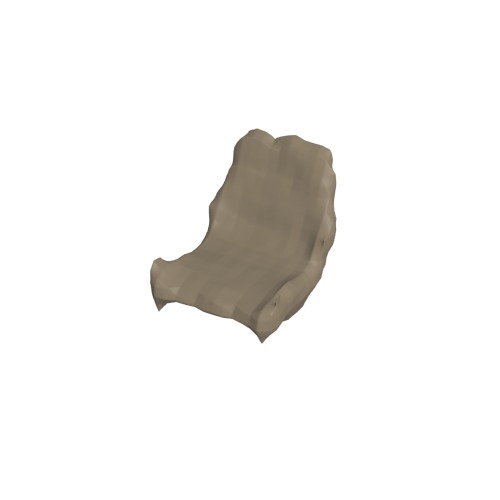}
    \includegraphics[width=0.115\linewidth, trim={\trimR cm \trimR cm \trimR cm \trimR cm}, clip]{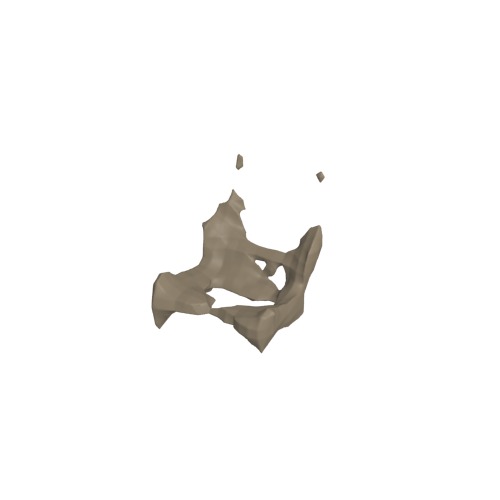}\\
    \includegraphics[width=0.115\linewidth, trim={\trimI cm \trimI cm \trimI cm \trimI cm}, clip]{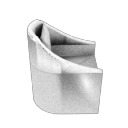}
    \includegraphics[width=0.115\linewidth, trim={\trimR cm \trimR cm \trimR cm \trimR cm}, clip]{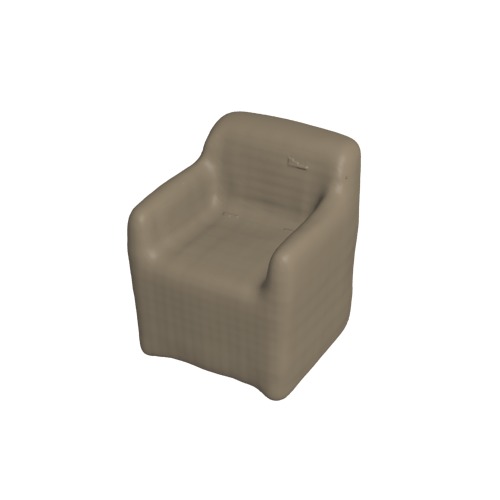}
    \includegraphics[width=0.115\linewidth, trim={\trimR cm \trimR cm \trimR cm \trimR cm}, clip]{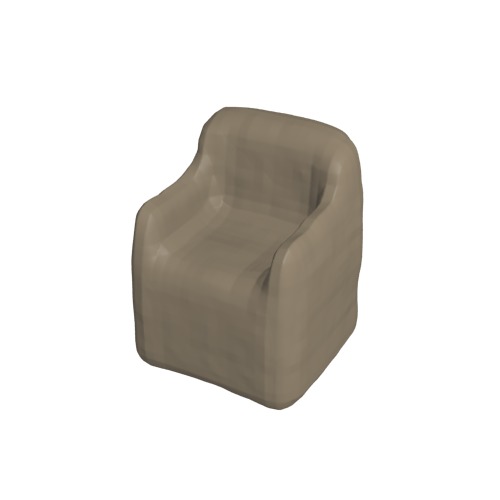}
    \includegraphics[width=0.115\linewidth, trim={\trimR cm \trimR cm \trimR cm \trimR cm}, clip]{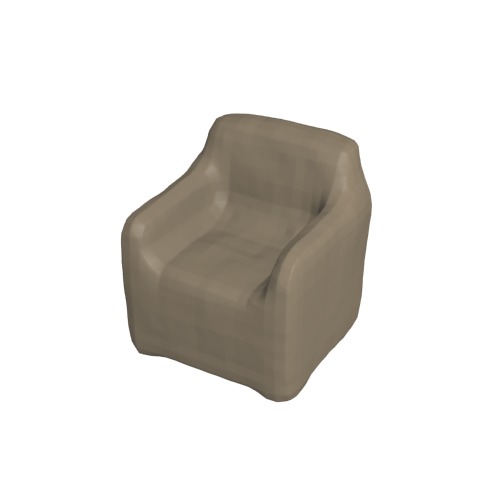}\hfill
    \includegraphics[width=0.115\linewidth, trim={\trimI cm \trimI cm \trimI cm \trimI cm}, clip]{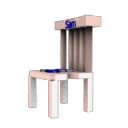}
    \includegraphics[width=0.115\linewidth, trim={\trimR cm \trimR cm \trimR cm \trimR cm}, clip]{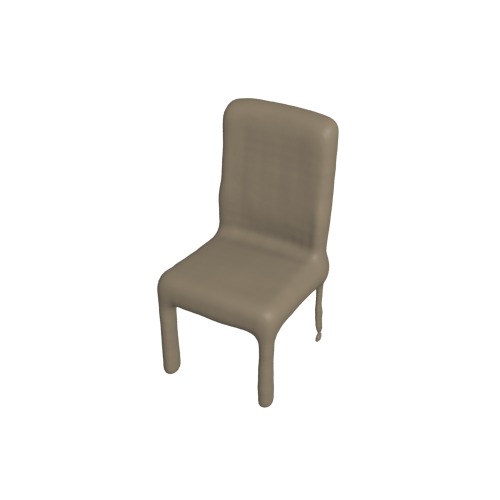}
    \includegraphics[width=0.115\linewidth, trim={\trimR cm \trimR cm \trimR cm \trimR cm}, clip]{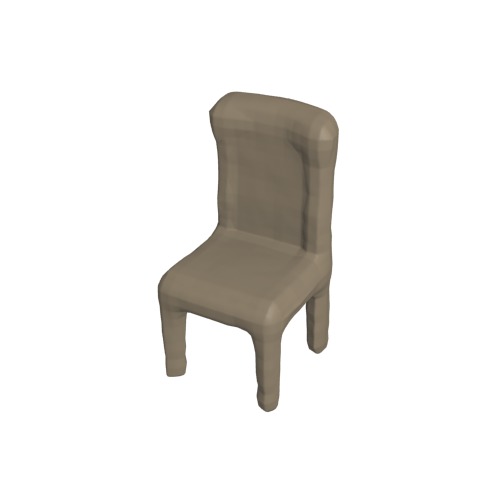}
    \includegraphics[width=0.115\linewidth, trim={\trimR cm \trimR cm \trimR cm \trimR cm}, clip]{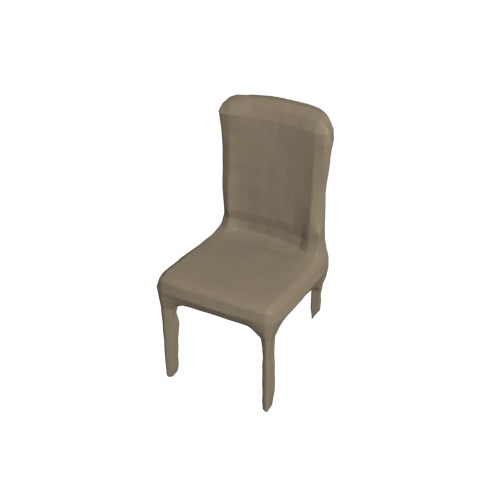}\\
    \includegraphics[width=0.115\linewidth, trim={\trimI cm \trimI cm \trimI cm \trimI cm}, clip]{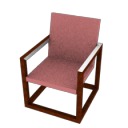}
    \includegraphics[width=0.115\linewidth, trim={\trimR cm \trimR cm \trimR cm \trimR cm}, clip]{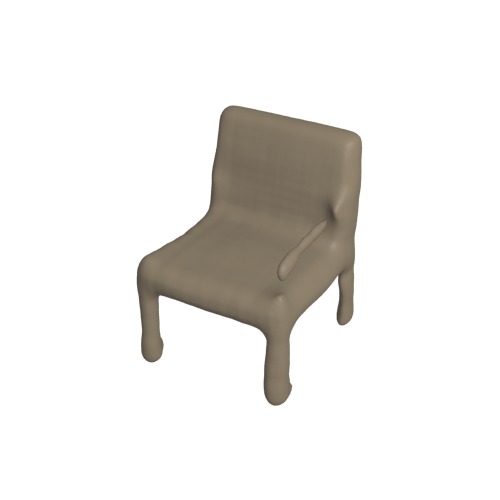}
    \includegraphics[width=0.115\linewidth, trim={\trimR cm \trimR cm \trimR cm \trimR cm}, clip]{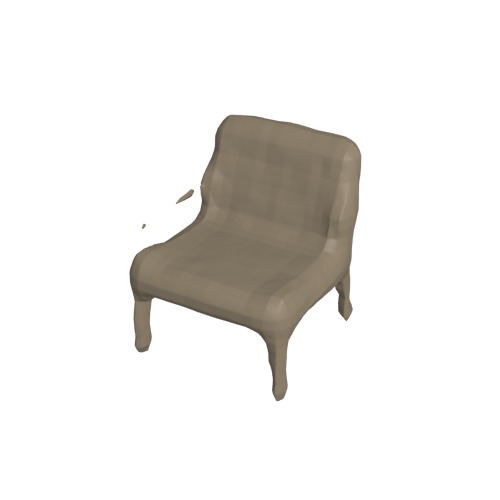}
    \includegraphics[width=0.115\linewidth, trim={\trimR cm \trimR cm \trimR cm \trimR cm}, clip]{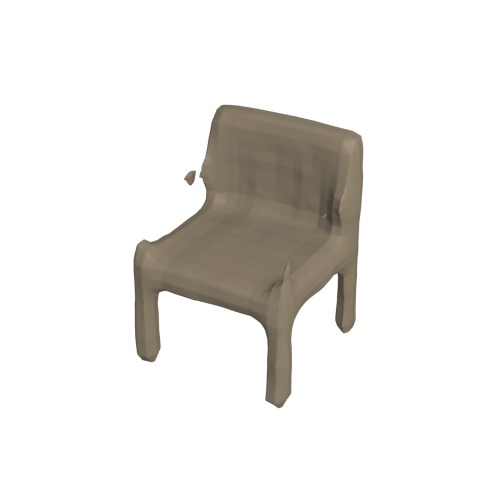}\hfill
    \includegraphics[width=0.115\linewidth, trim={\trimI cm \trimI cm \trimI cm \trimI cm}, clip]{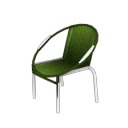}
    \includegraphics[width=0.115\linewidth, trim={\trimR cm \trimR cm \trimR cm \trimR cm}, clip]{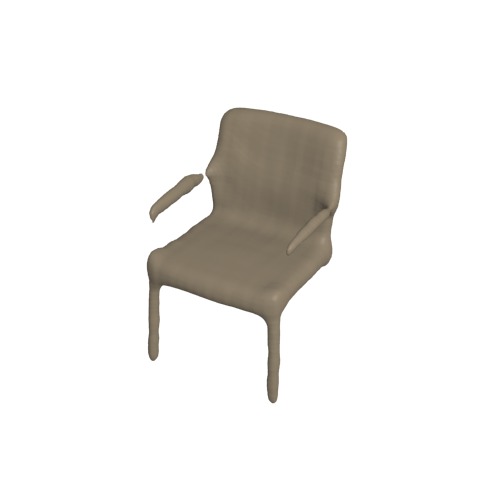}
    \includegraphics[width=0.115\linewidth, trim={\trimR cm \trimR cm \trimR cm \trimR cm}, clip]{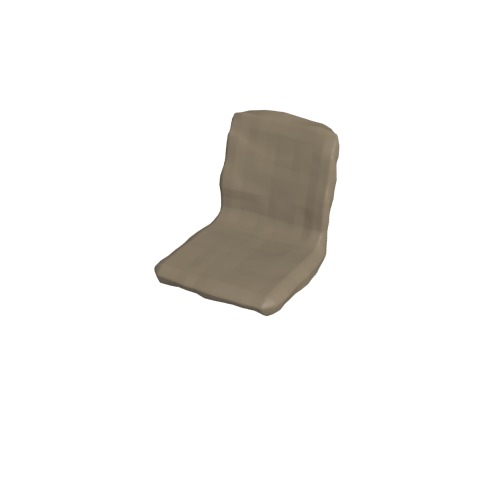}
    \includegraphics[width=0.115\linewidth, trim={\trimR cm \trimR cm \trimR cm \trimR cm}, clip]{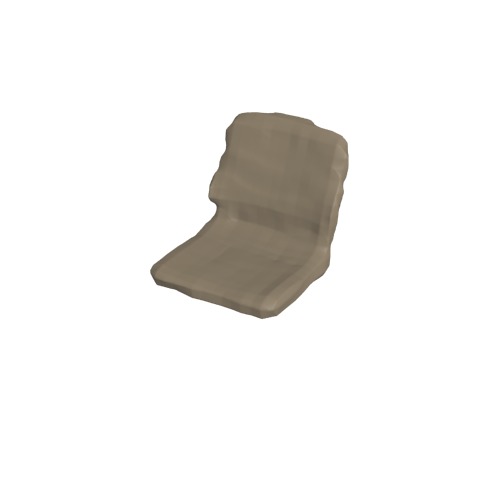}\\
    \includegraphics[width=0.115\linewidth, trim={\trimI cm \trimI cm \trimI cm \trimI cm}, clip]{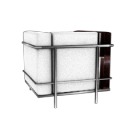}
    \includegraphics[width=0.115\linewidth, trim={\trimR cm \trimR cm \trimR cm \trimR cm}, clip]{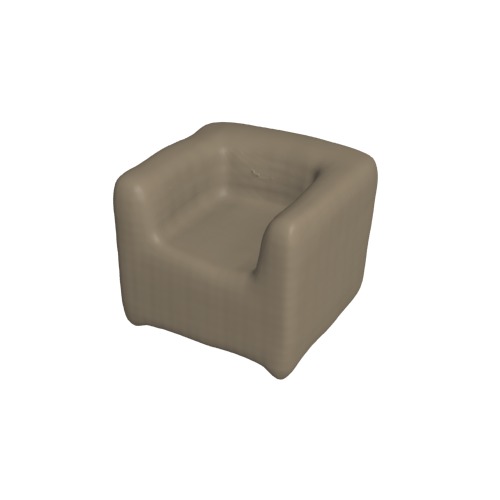}
    \includegraphics[width=0.115\linewidth, trim={\trimR cm \trimR cm \trimR cm \trimR cm}, clip]{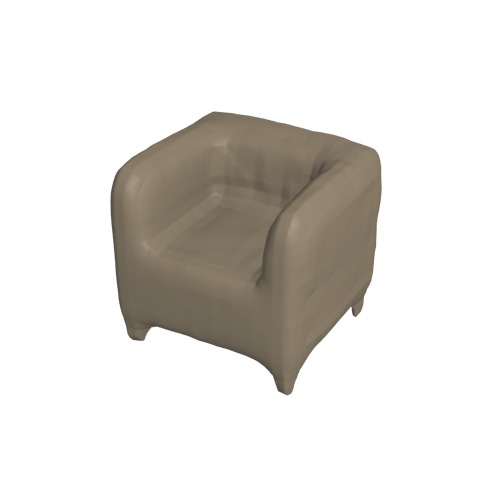}
    \includegraphics[width=0.115\linewidth, trim={\trimR cm \trimR cm \trimR cm \trimR cm}, clip]{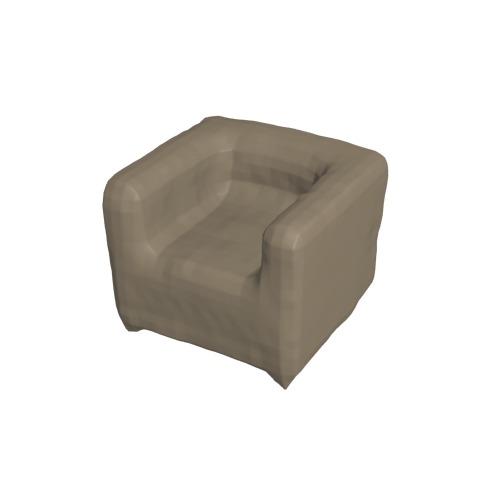}\hfill
    \includegraphics[width=0.115\linewidth, trim={\trimI cm \trimI cm \trimI cm \trimI cm}, clip]{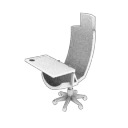}
    \includegraphics[width=0.115\linewidth, trim={\trimR cm \trimR cm \trimR cm \trimR cm}, clip]{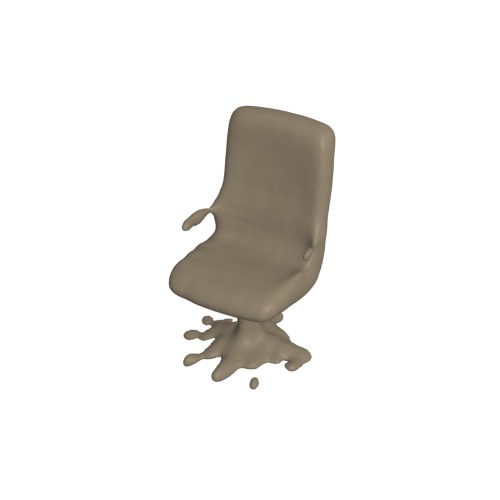}
    \includegraphics[width=0.115\linewidth, trim={\trimR cm \trimR cm \trimR cm \trimR cm}, clip]{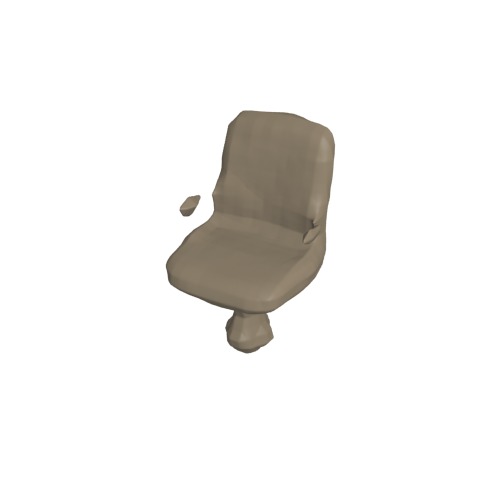}
    \includegraphics[width=0.115\linewidth, trim={\trimR cm \trimR cm \trimR cm \trimR cm}, clip]{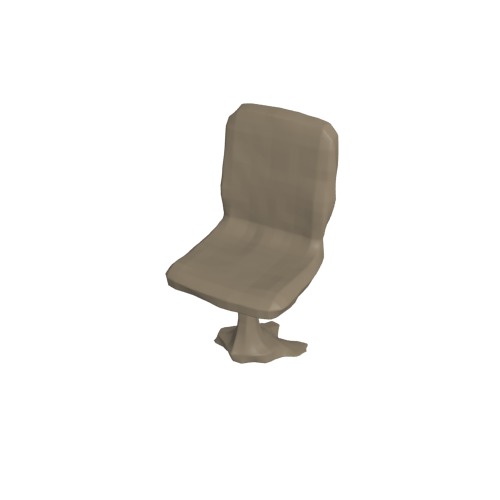}\\
    \includegraphics[width=0.115\linewidth, trim={\trimI cm \trimI cm \trimI cm \trimI cm}, clip]{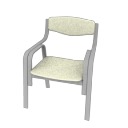}
    \includegraphics[width=0.115\linewidth, trim={\trimR cm \trimR cm \trimR cm \trimR cm}, clip]{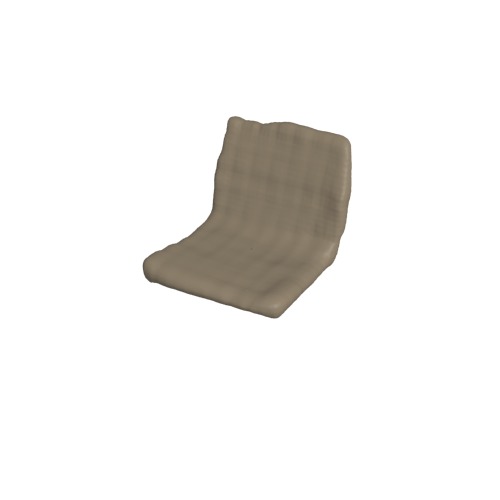}
    \includegraphics[width=0.115\linewidth, trim={\trimR cm \trimR cm \trimR cm \trimR cm}, clip]{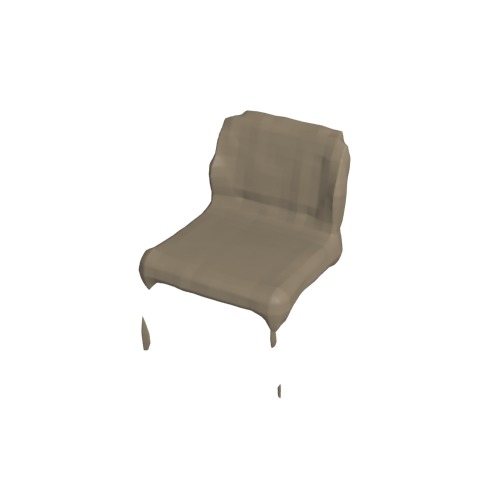}
    \includegraphics[width=0.115\linewidth, trim={\trimR cm \trimR cm \trimR cm \trimR cm}, clip]{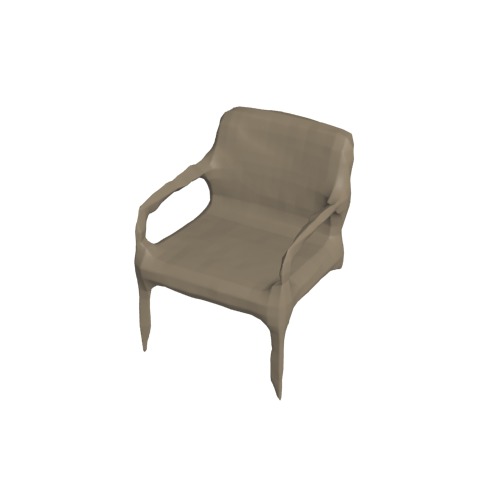}\hfill
    \includegraphics[width=0.115\linewidth, trim={\trimI cm \trimI cm \trimI cm \trimI cm}, clip]{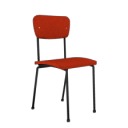}
    \includegraphics[width=0.115\linewidth, trim={\trimR cm \trimR cm \trimR cm \trimR cm}, clip]{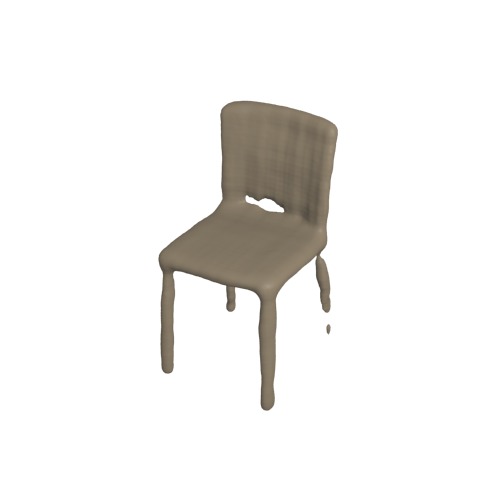}
    \includegraphics[width=0.115\linewidth, trim={\trimR cm \trimR cm \trimR cm \trimR cm}, clip]{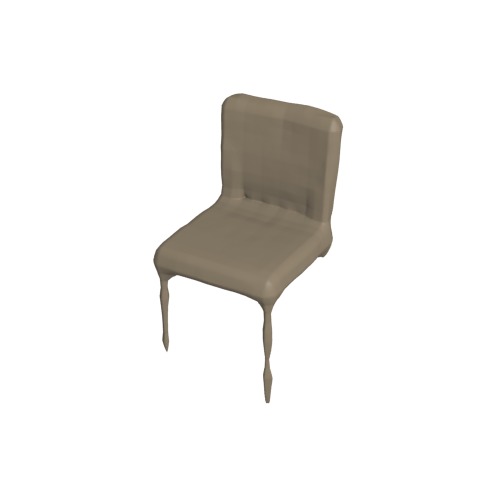}
    \includegraphics[width=0.115\linewidth, trim={\trimR cm \trimR cm \trimR cm \trimR cm}, clip]{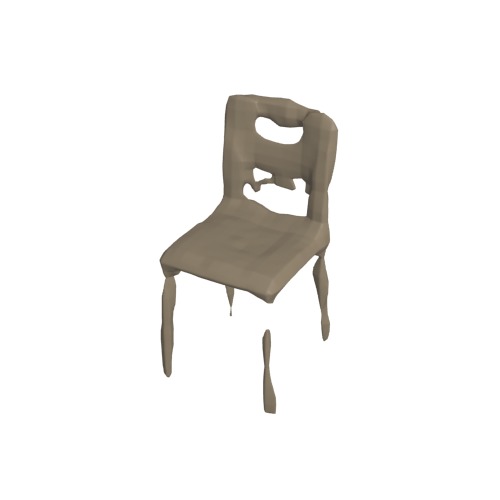} \\
    \includegraphics[width=0.115\linewidth, trim={\trimI cm \trimI cm \trimI cm \trimI cm}, clip]{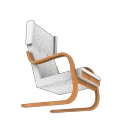}
    \includegraphics[width=0.115\linewidth, trim={\trimR cm \trimR cm \trimR cm \trimR cm}, clip]{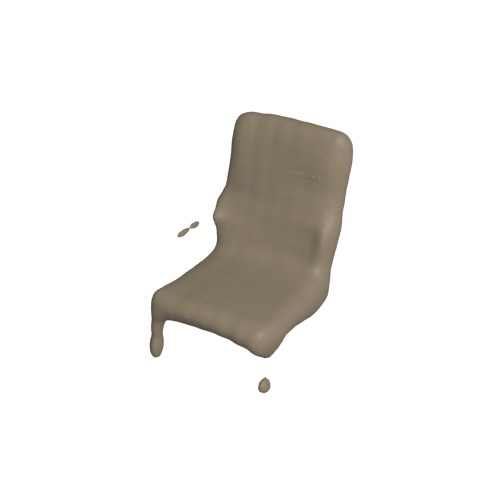}
    \includegraphics[width=0.115\linewidth, trim={\trimR cm \trimR cm \trimR cm \trimR cm}, clip]{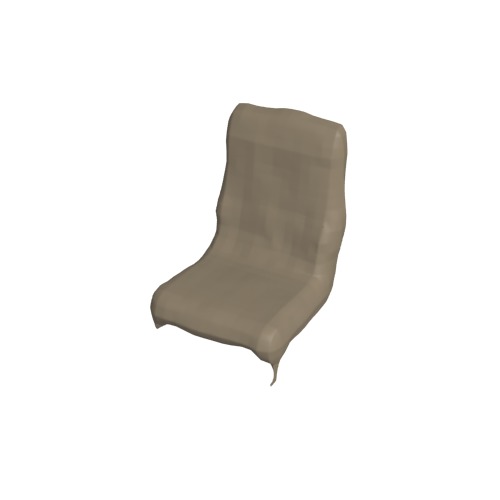}
    \includegraphics[width=0.115\linewidth, trim={\trimR cm \trimR cm \trimR cm \trimR cm}, clip]{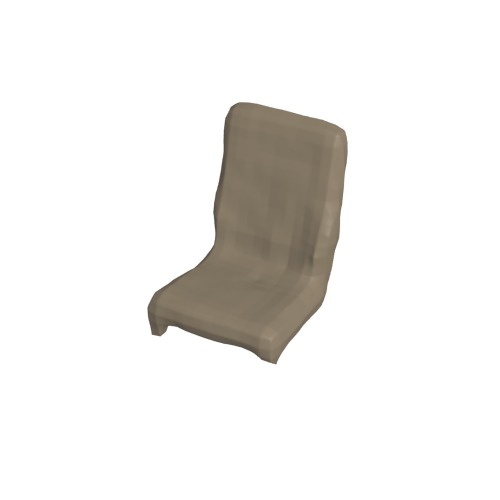}\hfill
    \includegraphics[width=0.115\linewidth, trim={\trimI cm \trimI cm \trimI cm \trimI cm}, clip]{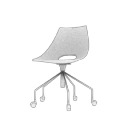}
    \includegraphics[width=0.115\linewidth, trim={\trimR cm \trimR cm \trimR cm \trimR cm}, clip]{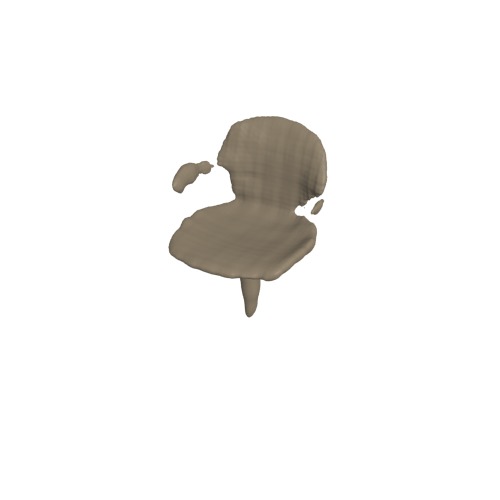}
    \includegraphics[width=0.115\linewidth, trim={\trimR cm \trimR cm \trimR cm \trimR cm}, clip]{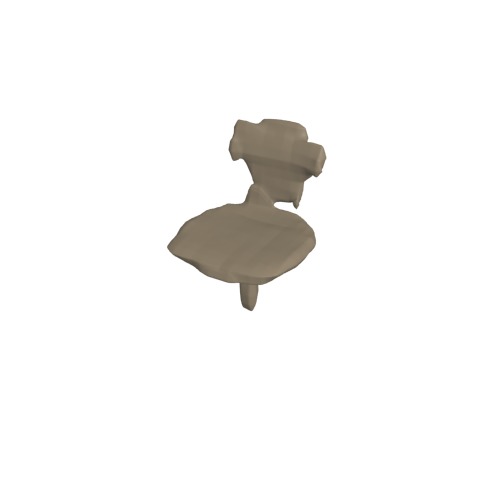}
    \includegraphics[width=0.115\linewidth, trim={\trimR cm \trimR cm \trimR cm \trimR cm}, clip]{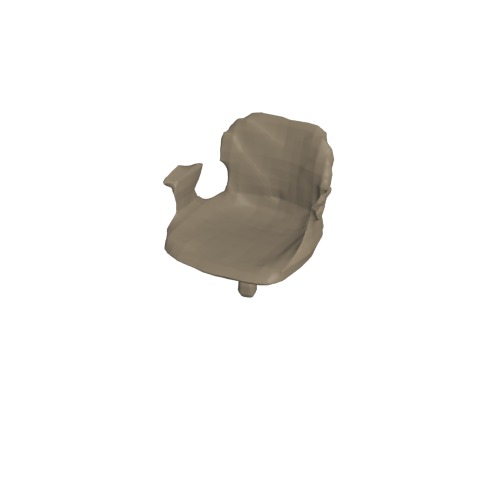}
    \caption{Results for the category chair. Each block from left to right, input image, our proposed HSP, LR Soft, LR Hard.}
    \label{fig:chair1}
  \end{figure*}
  \begin{figure*}
    \includegraphics[width=0.115\linewidth, trim={\trimI cm \trimI cm \trimI cm \trimI cm}, clip]{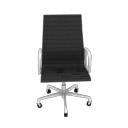}
    \includegraphics[width=0.115\linewidth, trim={\trimR cm \trimR cm \trimR cm \trimR cm}, clip]{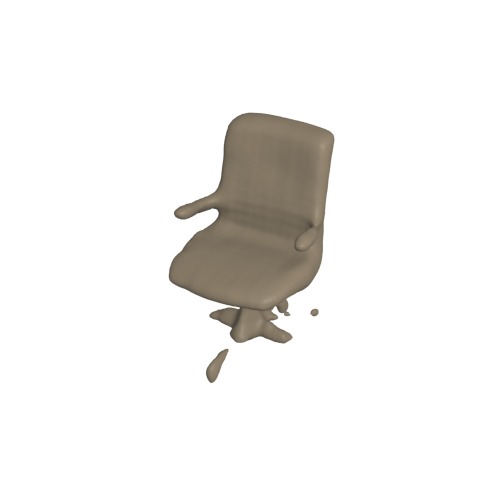}
    \includegraphics[width=0.115\linewidth, trim={\trimR cm \trimR cm \trimR cm \trimR cm}, clip]{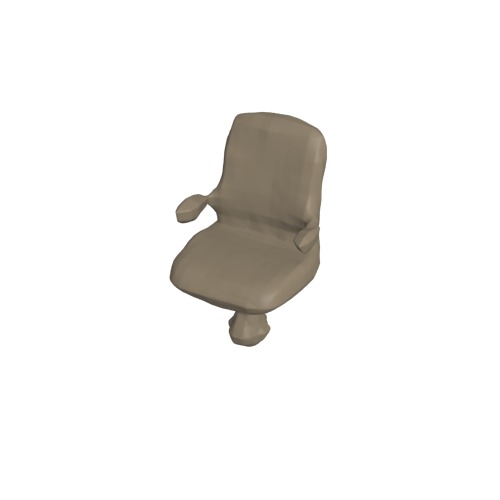}
    \includegraphics[width=0.115\linewidth, trim={\trimR cm \trimR cm \trimR cm \trimR cm}, clip]{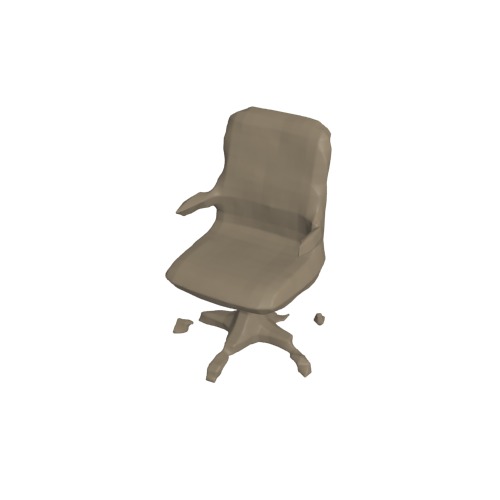}\hfill
    \includegraphics[width=0.115\linewidth, trim={\trimI cm \trimI cm \trimI cm \trimI cm}, clip]{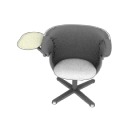}
    \includegraphics[width=0.115\linewidth, trim={\trimR cm \trimR cm \trimR cm \trimR cm}, clip]{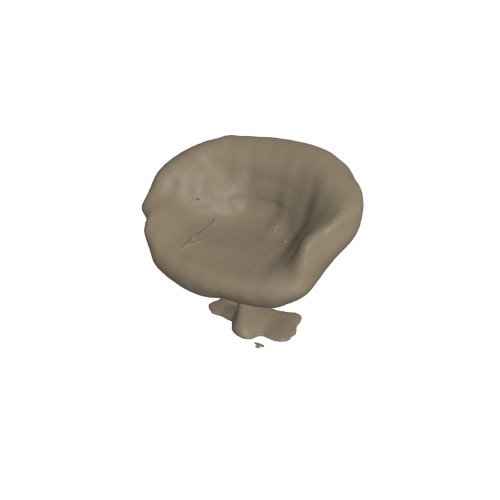}
    \includegraphics[width=0.115\linewidth, trim={\trimR cm \trimR cm \trimR cm \trimR cm}, clip]{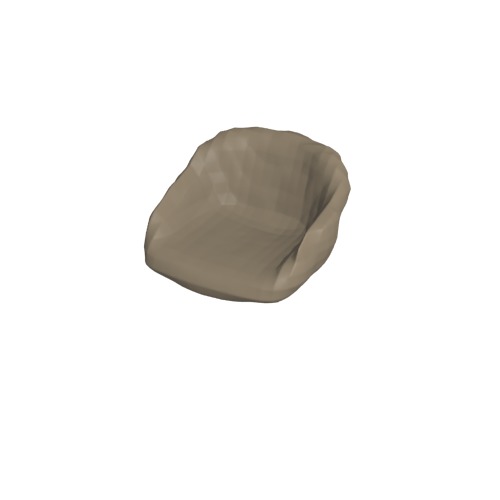}
    \includegraphics[width=0.115\linewidth, trim={\trimR cm \trimR cm \trimR cm \trimR cm}, clip]{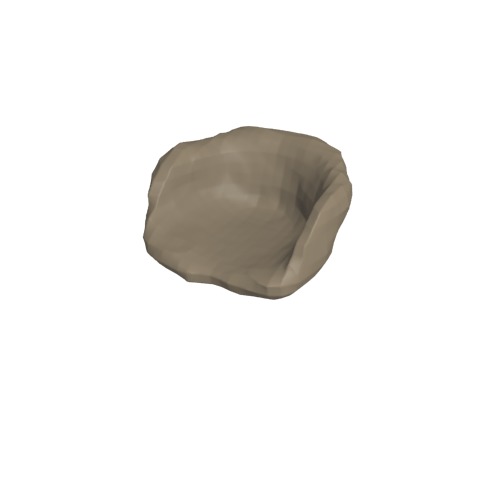} \\
    \includegraphics[width=0.115\linewidth, trim={\trimI cm \trimI cm \trimI cm \trimI cm}, clip]{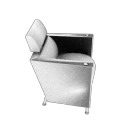}
    \includegraphics[width=0.115\linewidth, trim={\trimR cm \trimR cm \trimR cm \trimR cm}, clip]{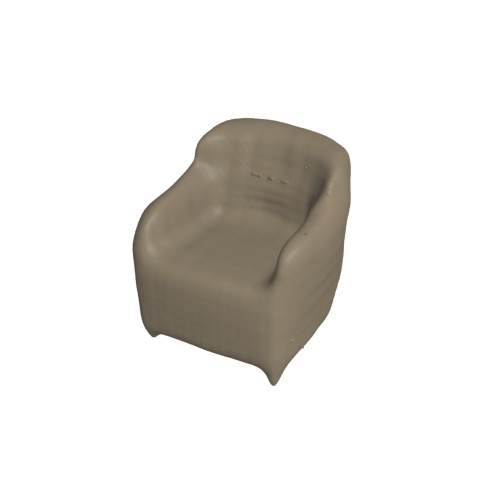}
    \includegraphics[width=0.115\linewidth, trim={\trimR cm \trimR cm \trimR cm \trimR cm}, clip]{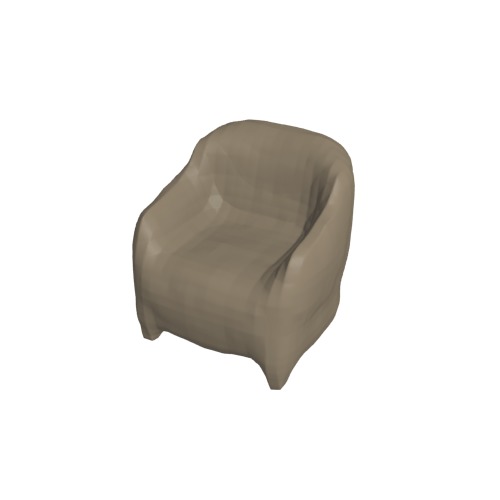}
    \includegraphics[width=0.115\linewidth, trim={\trimR cm \trimR cm \trimR cm \trimR cm}, clip]{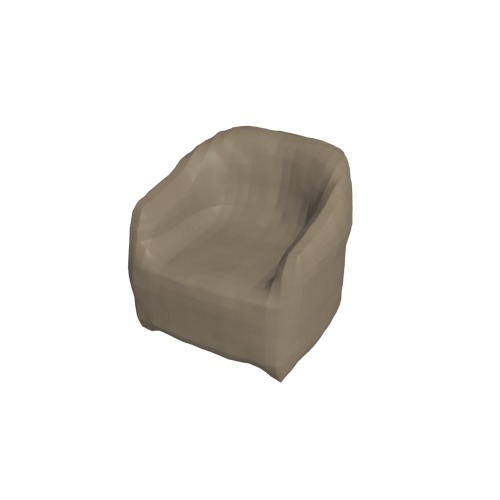}\hfill
    \includegraphics[width=0.115\linewidth, trim={\trimI cm \trimI cm \trimI cm \trimI cm}, clip]{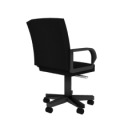}
    \includegraphics[width=0.115\linewidth, trim={\trimR cm \trimR cm \trimR cm \trimR cm}, clip]{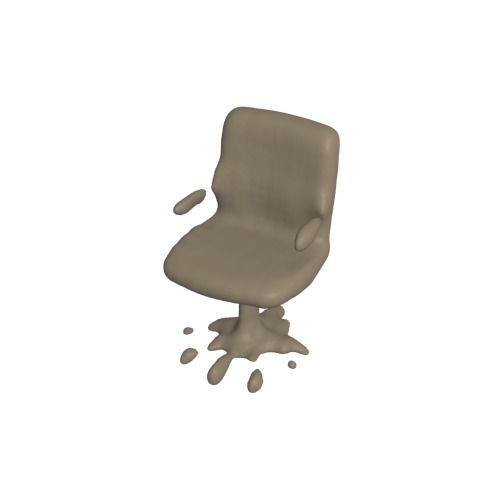}
    \includegraphics[width=0.115\linewidth, trim={\trimR cm \trimR cm \trimR cm \trimR cm}, clip]{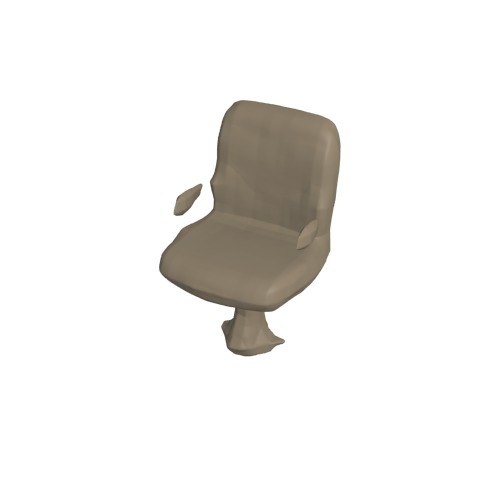}
    \includegraphics[width=0.115\linewidth, trim={\trimR cm \trimR cm \trimR cm \trimR cm}, clip]{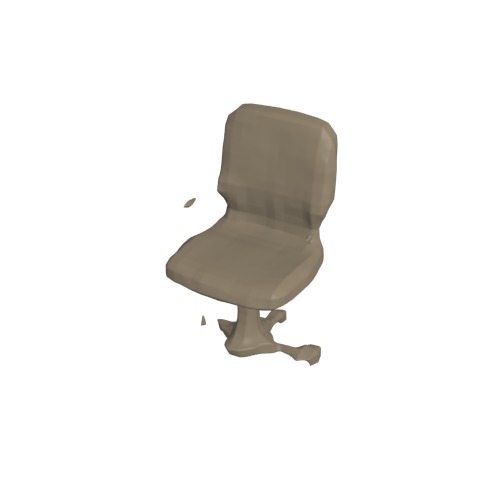}\\
    \includegraphics[width=0.115\linewidth, trim={\trimI cm \trimI cm \trimI cm \trimI cm}, clip]{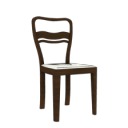}
    \includegraphics[width=0.115\linewidth, trim={\trimR cm \trimR cm \trimR cm \trimR cm}, clip]{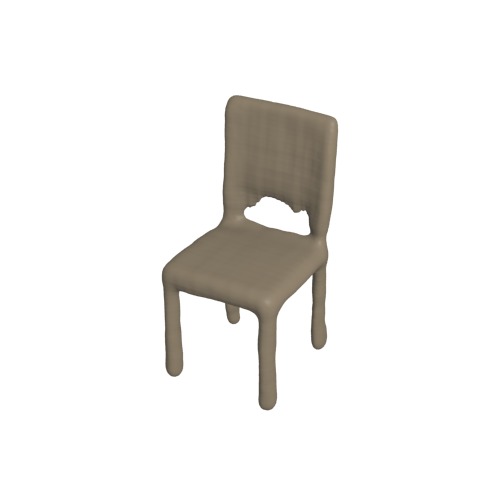}
    \includegraphics[width=0.115\linewidth, trim={\trimR cm \trimR cm \trimR cm \trimR cm}, clip]{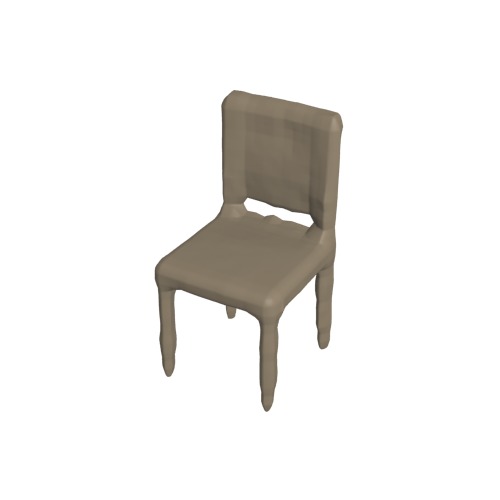}
    \includegraphics[width=0.115\linewidth, trim={\trimR cm \trimR cm \trimR cm \trimR cm}, clip]{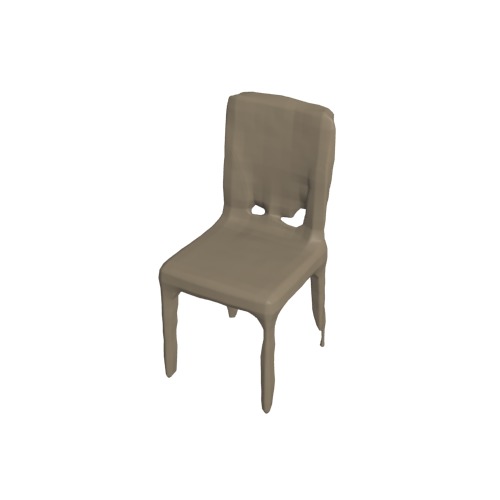}\hfill
    \includegraphics[width=0.115\linewidth, trim={\trimI cm \trimI cm \trimI cm \trimI cm}, clip]{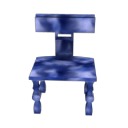}
    \includegraphics[width=0.115\linewidth, trim={\trimR cm \trimR cm \trimR cm \trimR cm}, clip]{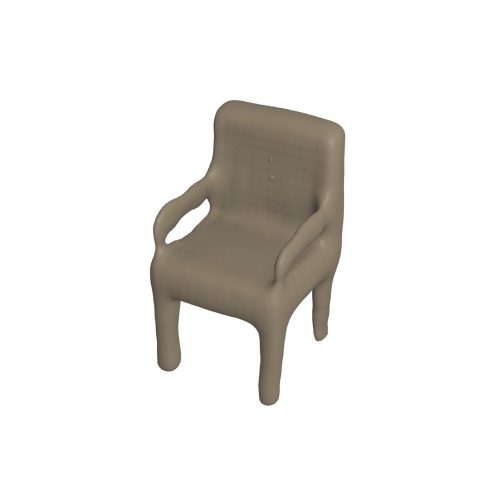}
    \includegraphics[width=0.115\linewidth, trim={\trimR cm \trimR cm \trimR cm \trimR cm}, clip]{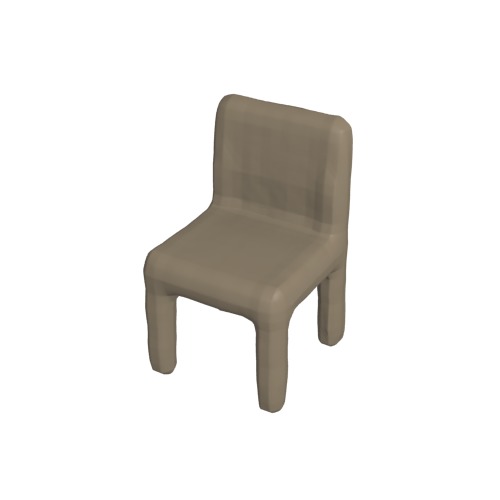}
    \includegraphics[width=0.115\linewidth, trim={\trimR cm \trimR cm \trimR cm \trimR cm}, clip]{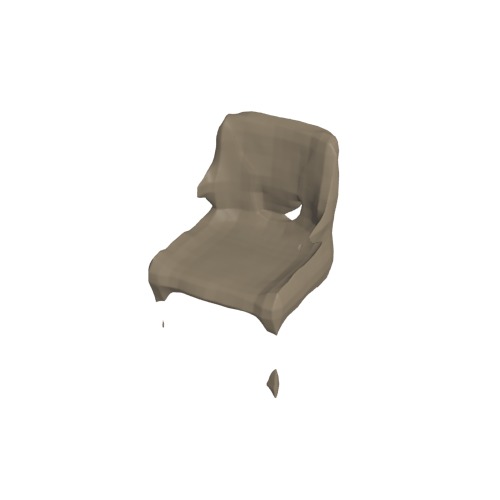}\\
    \includegraphics[width=0.115\linewidth, trim={\trimI cm \trimI cm \trimI cm \trimI cm}, clip]{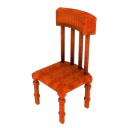}
    \includegraphics[width=0.115\linewidth, trim={\trimR cm \trimR cm \trimR cm \trimR cm}, clip]{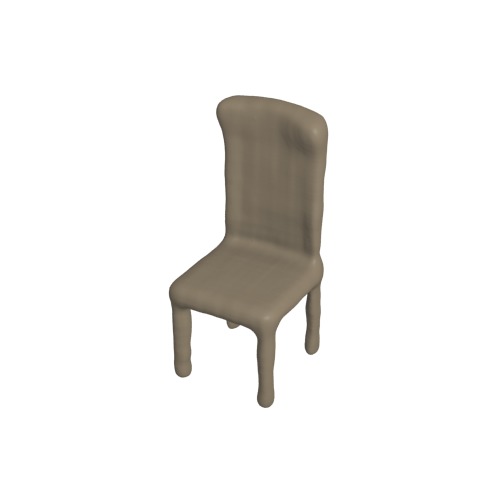}
    \includegraphics[width=0.115\linewidth, trim={\trimR cm \trimR cm \trimR cm \trimR cm}, clip]{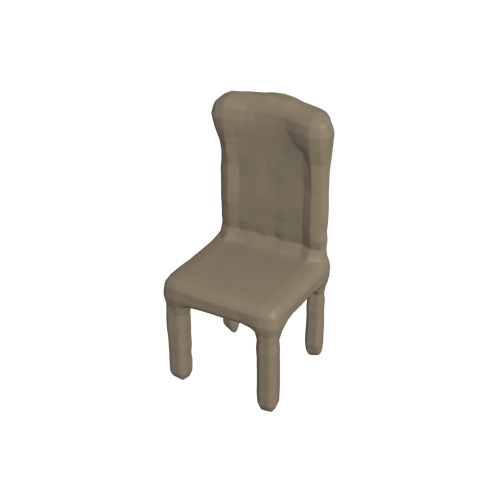}
    \includegraphics[width=0.115\linewidth, trim={\trimR cm \trimR cm \trimR cm \trimR cm}, clip]{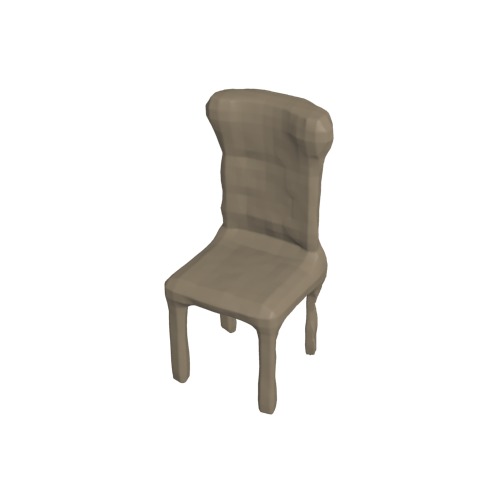}\hfill
    \includegraphics[width=0.115\linewidth, trim={\trimI cm \trimI cm \trimI cm \trimI cm}, clip]{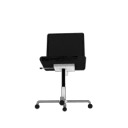}
    \includegraphics[width=0.115\linewidth, trim={\trimR cm \trimR cm \trimR cm \trimR cm}, clip]{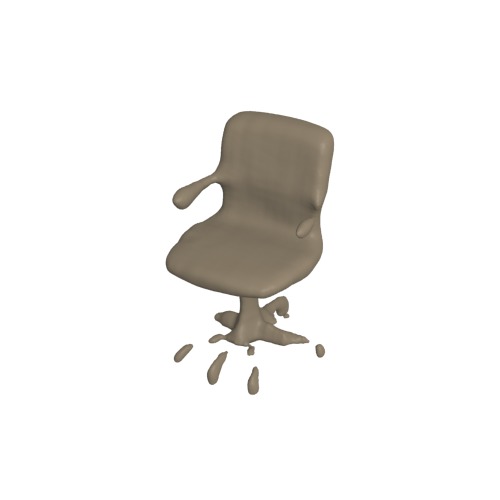}
    \includegraphics[width=0.115\linewidth, trim={\trimR cm \trimR cm \trimR cm \trimR cm}, clip]{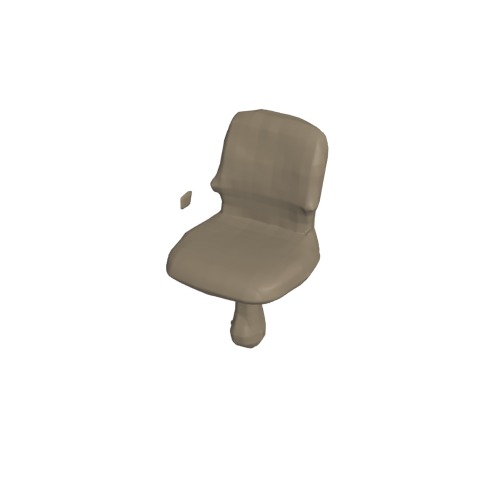}
    \includegraphics[width=0.115\linewidth, trim={\trimR cm \trimR cm \trimR cm \trimR cm}, clip]{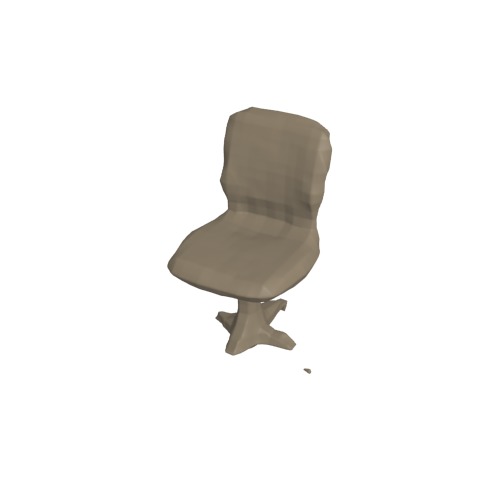}\\
    \includegraphics[width=0.115\linewidth, trim={\trimI cm \trimI cm \trimI cm \trimI cm}, clip]{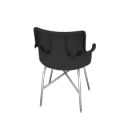}
    \includegraphics[width=0.115\linewidth, trim={\trimR cm \trimR cm \trimR cm \trimR cm}, clip]{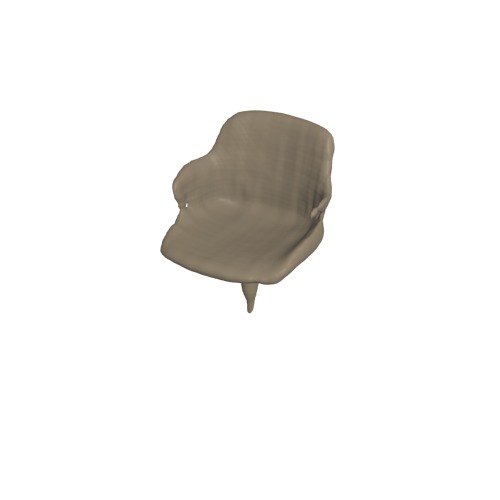}
    \includegraphics[width=0.115\linewidth, trim={\trimR cm \trimR cm \trimR cm \trimR cm}, clip]{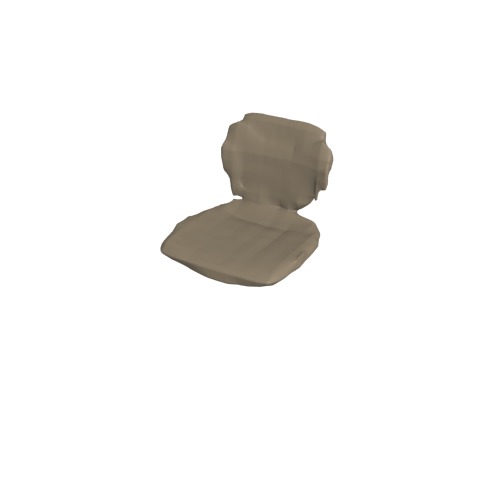}
    \includegraphics[width=0.115\linewidth, trim={\trimR cm \trimR cm \trimR cm \trimR cm}, clip]{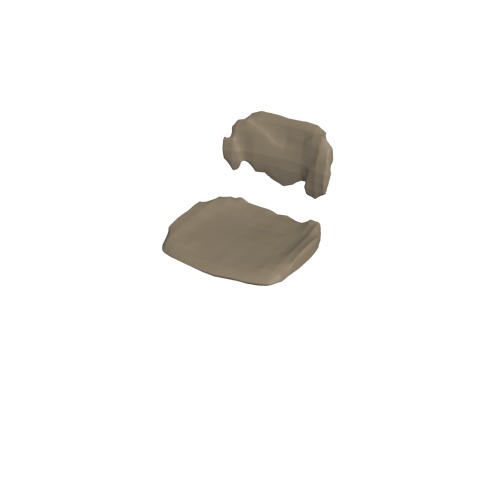}\hfill
    \includegraphics[width=0.115\linewidth, trim={\trimI cm \trimI cm \trimI cm \trimI cm}, clip]{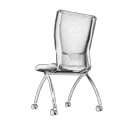}
    \includegraphics[width=0.115\linewidth, trim={\trimR cm \trimR cm \trimR cm \trimR cm}, clip]{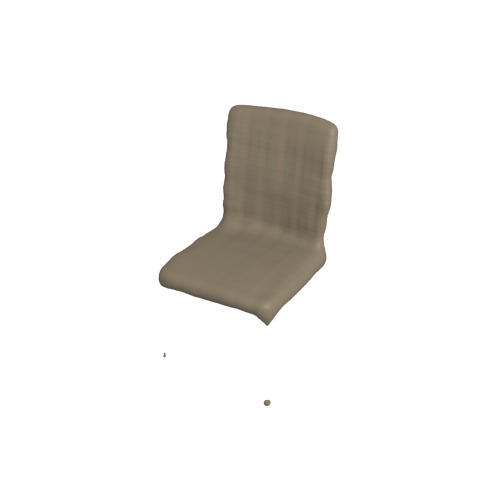}
    \includegraphics[width=0.115\linewidth, trim={\trimR cm \trimR cm \trimR cm \trimR cm}, clip]{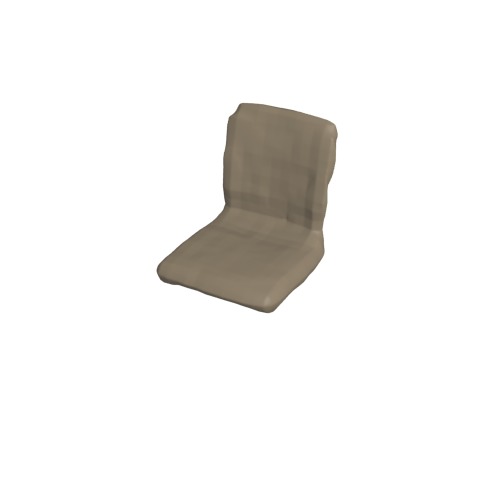}
    \includegraphics[width=0.115\linewidth, trim={\trimR cm \trimR cm \trimR cm \trimR cm}, clip]{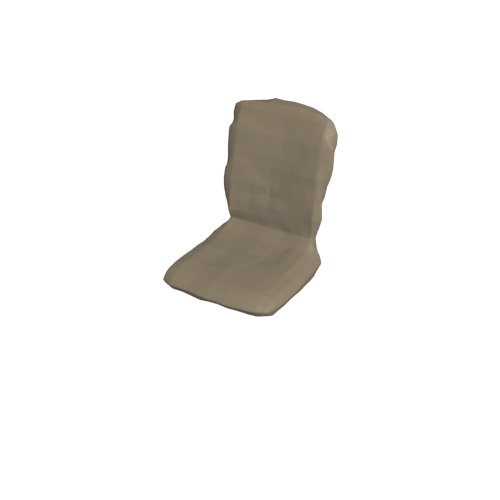}\\
    \includegraphics[width=0.115\linewidth, trim={\trimI cm \trimI cm \trimI cm \trimI cm}, clip]{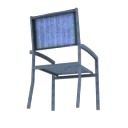}
    \includegraphics[width=0.115\linewidth, trim={\trimR cm \trimR cm \trimR cm \trimR cm}, clip]{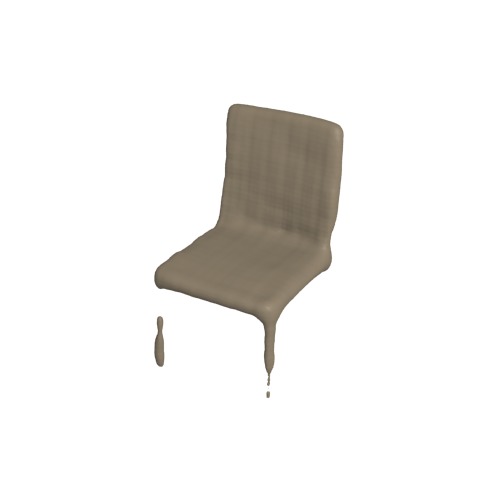}
    \includegraphics[width=0.115\linewidth, trim={\trimR cm \trimR cm \trimR cm \trimR cm}, clip]{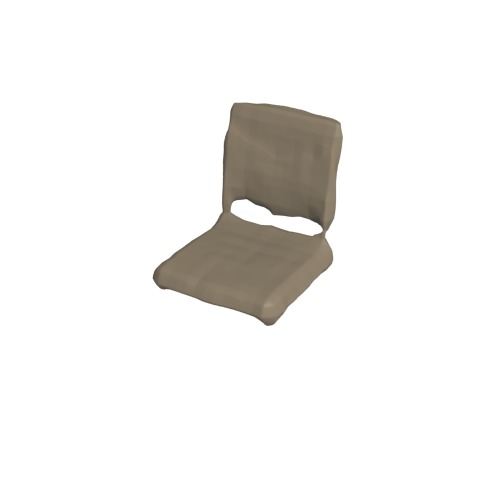}
    \includegraphics[width=0.115\linewidth, trim={\trimR cm \trimR cm \trimR cm \trimR cm}, clip]{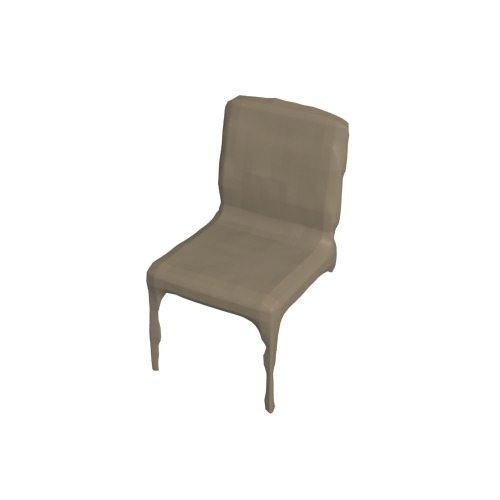}\hfill
    \includegraphics[width=0.115\linewidth, trim={\trimI cm \trimI cm \trimI cm \trimI cm}, clip]{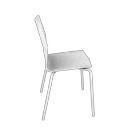}
    \includegraphics[width=0.115\linewidth, trim={\trimR cm \trimR cm \trimR cm \trimR cm}, clip]{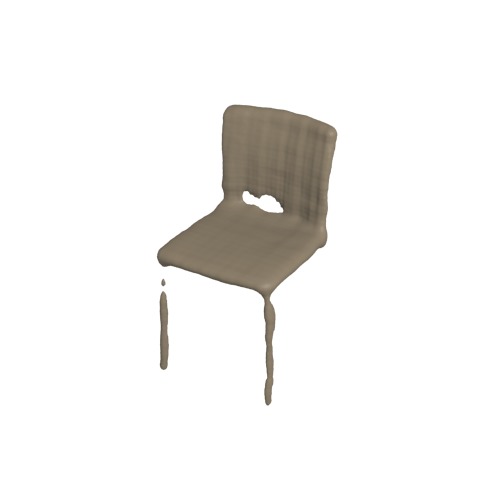}
    \includegraphics[width=0.115\linewidth, trim={\trimR cm \trimR cm \trimR cm \trimR cm}, clip]{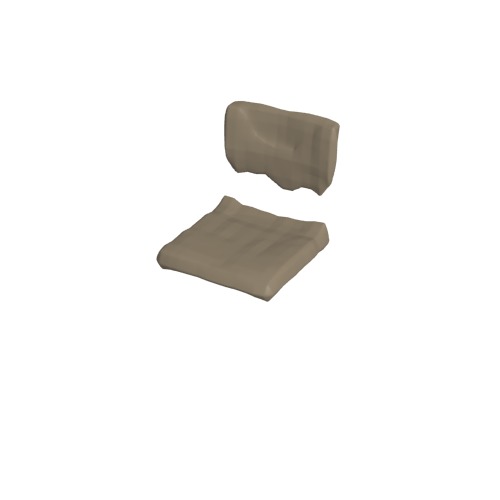}
    \includegraphics[width=0.115\linewidth, trim={\trimR cm \trimR cm \trimR cm \trimR cm}, clip]{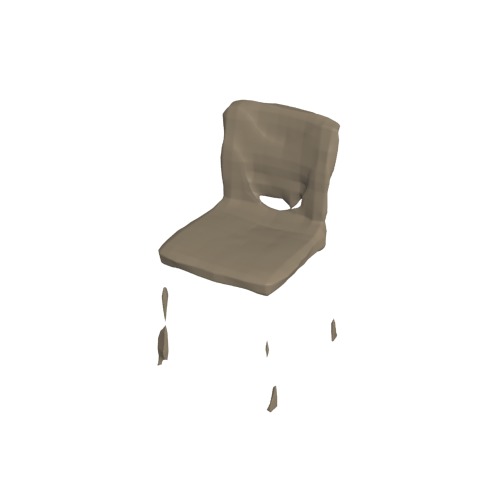}\\
    \includegraphics[width=0.115\linewidth, trim={\trimI cm \trimI cm \trimI cm \trimI cm}, clip]{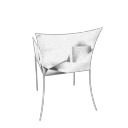}
    \includegraphics[width=0.115\linewidth, trim={\trimR cm \trimR cm \trimR cm \trimR cm}, clip]{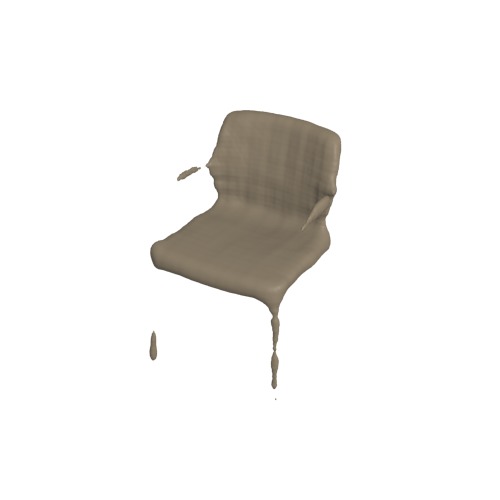}
    \includegraphics[width=0.115\linewidth, trim={\trimR cm \trimR cm \trimR cm \trimR cm}, clip]{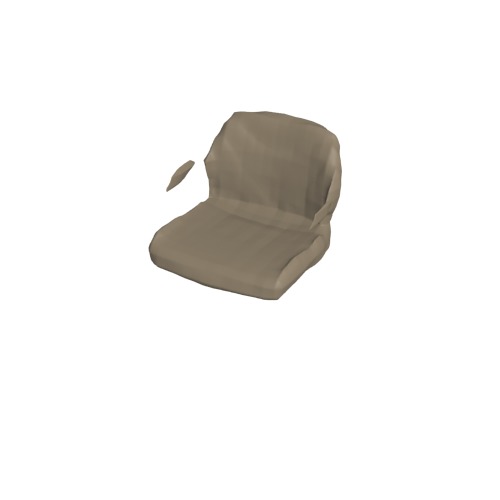}
    \includegraphics[width=0.115\linewidth, trim={\trimR cm \trimR cm \trimR cm \trimR cm}, clip]{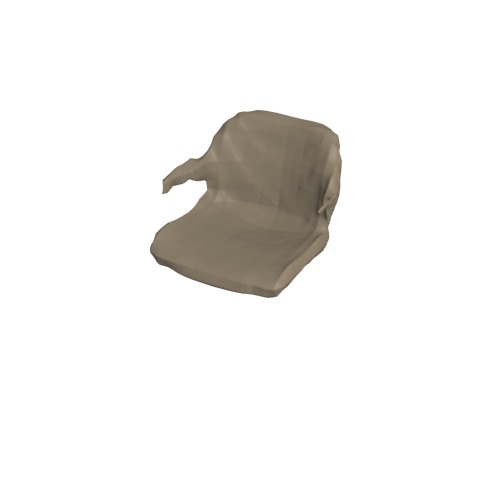}\hfill
    \includegraphics[width=0.115\linewidth, trim={\trimI cm \trimI cm \trimI cm \trimI cm}, clip]{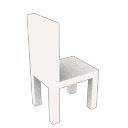}
    \includegraphics[width=0.115\linewidth, trim={\trimR cm \trimR cm \trimR cm \trimR cm}, clip]{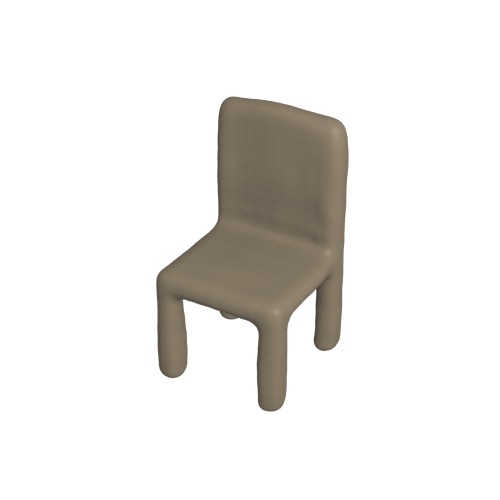}
    \includegraphics[width=0.115\linewidth, trim={\trimR cm \trimR cm \trimR cm \trimR cm}, clip]{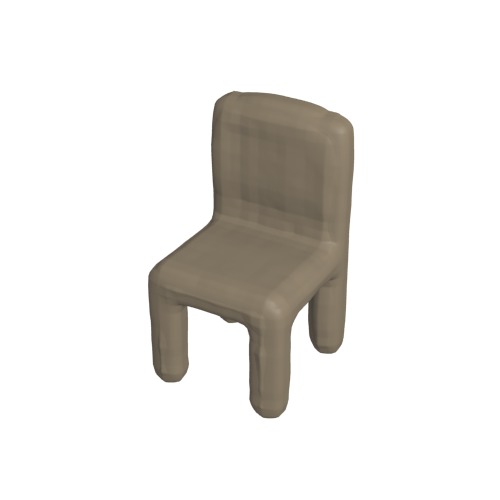}
    \includegraphics[width=0.115\linewidth, trim={\trimR cm \trimR cm \trimR cm \trimR cm}, clip]{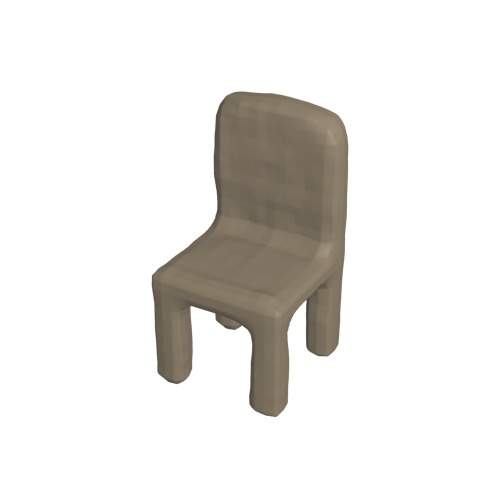}\\
    \includegraphics[width=0.115\linewidth, trim={\trimI cm \trimI cm \trimI cm \trimI cm}, clip]{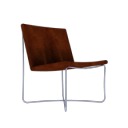}
    \includegraphics[width=0.115\linewidth, trim={\trimR cm \trimR cm \trimR cm \trimR cm}, clip]{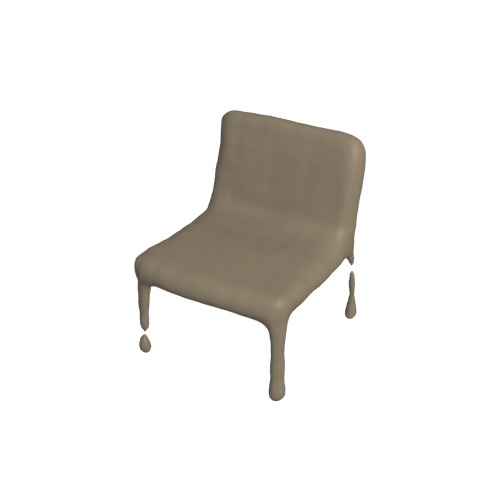}
    \includegraphics[width=0.115\linewidth, trim={\trimR cm \trimR cm \trimR cm \trimR cm}, clip]{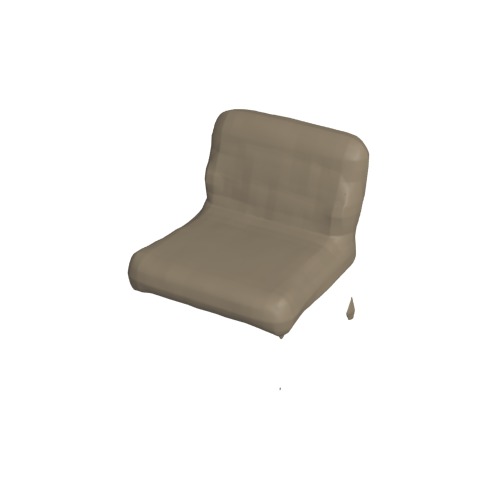}
    \includegraphics[width=0.115\linewidth, trim={\trimR cm \trimR cm \trimR cm \trimR cm}, clip]{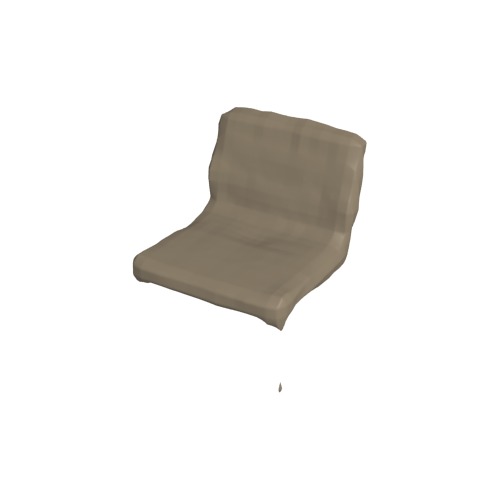}\hfill
    \includegraphics[width=0.115\linewidth, trim={\trimI cm \trimI cm \trimI cm \trimI cm}, clip]{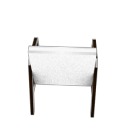}
    \includegraphics[width=0.115\linewidth, trim={\trimR cm \trimR cm \trimR cm \trimR cm}, clip]{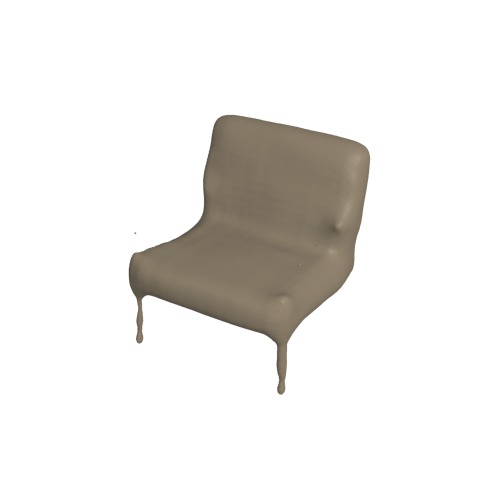}
    \includegraphics[width=0.115\linewidth, trim={\trimR cm \trimR cm \trimR cm \trimR cm}, clip]{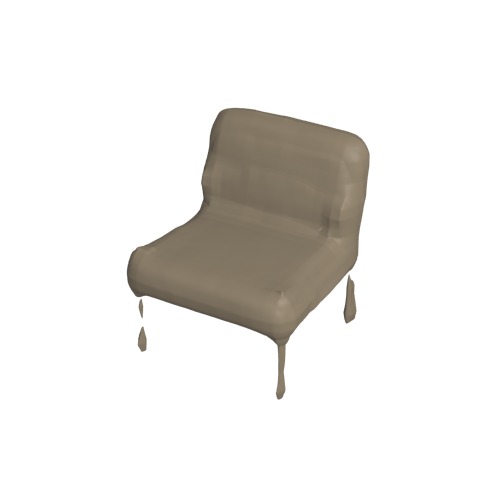}
    \includegraphics[width=0.115\linewidth, trim={\trimR cm \trimR cm \trimR cm \trimR cm}, clip]{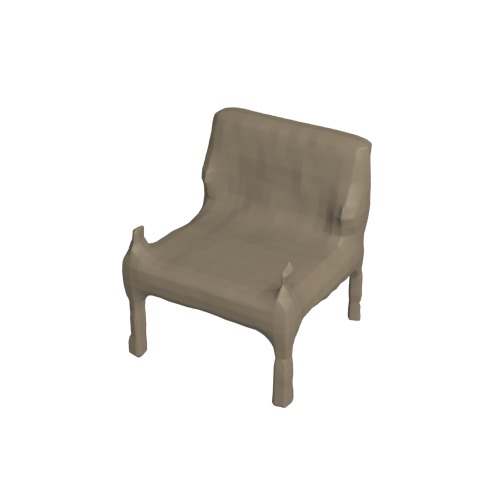} \\
    \includegraphics[width=0.115\linewidth, trim={\trimI cm \trimI cm \trimI cm \trimI cm}, clip]{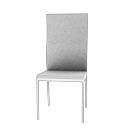}
    \includegraphics[width=0.115\linewidth, trim={\trimR cm \trimR cm \trimR cm \trimR cm}, clip]{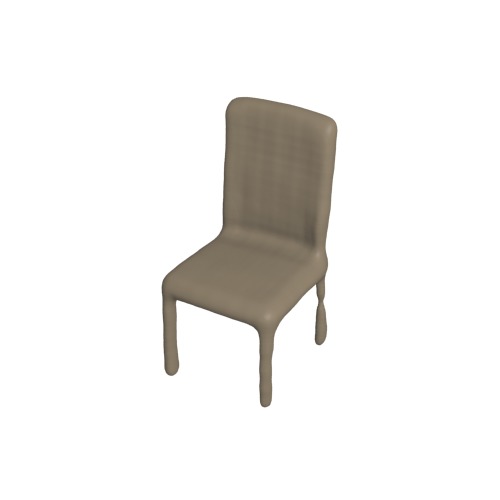}
    \includegraphics[width=0.115\linewidth, trim={\trimR cm \trimR cm \trimR cm \trimR cm}, clip]{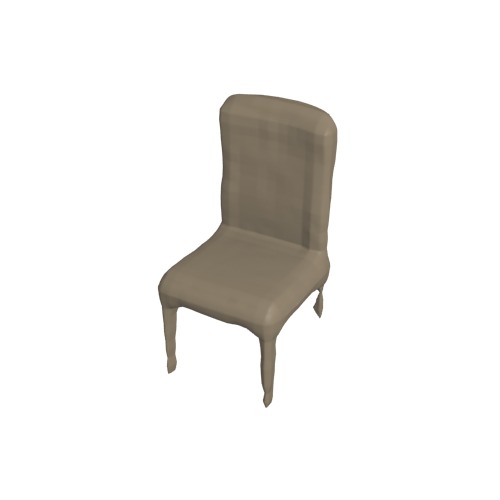}
    \includegraphics[width=0.115\linewidth, trim={\trimR cm \trimR cm \trimR cm \trimR cm}, clip]{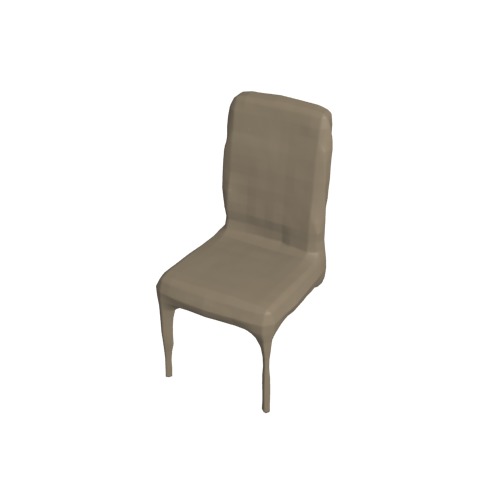}\hfill
    \includegraphics[width=0.115\linewidth, trim={\trimI cm \trimI cm \trimI cm \trimI cm}, clip]{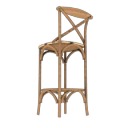}
    \includegraphics[width=0.115\linewidth, trim={\trimR cm \trimR cm \trimR cm \trimR cm}, clip]{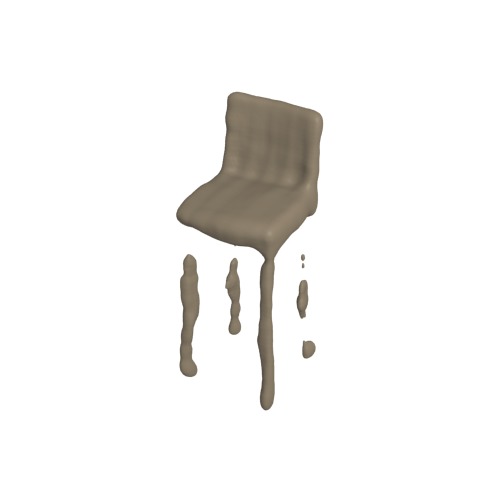}
    \includegraphics[width=0.115\linewidth, trim={\trimR cm \trimR cm \trimR cm \trimR cm}, clip]{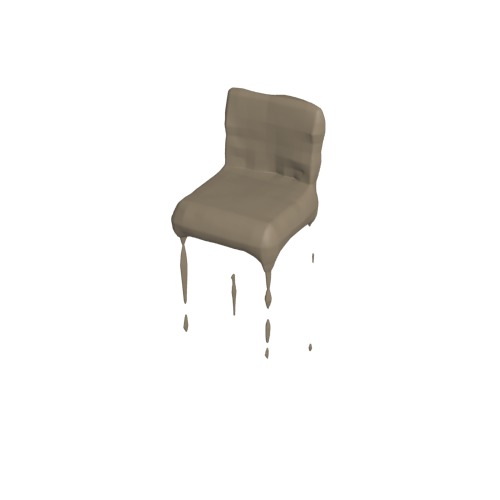}
    \includegraphics[width=0.115\linewidth, trim={\trimR cm \trimR cm \trimR cm \trimR cm}, clip]{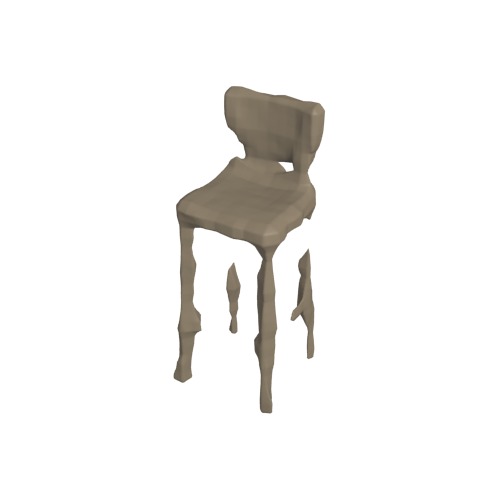} \\
    \includegraphics[width=0.115\linewidth, trim={\trimI cm \trimI cm \trimI cm \trimI cm}, clip]{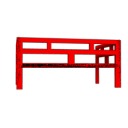}
    \includegraphics[width=0.115\linewidth, trim={\trimR cm \trimR cm \trimR cm \trimR cm}, clip]{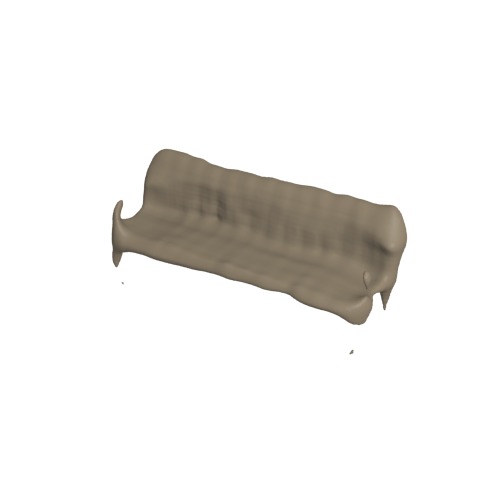}
    \includegraphics[width=0.115\linewidth, trim={\trimR cm \trimR cm \trimR cm \trimR cm}, clip]{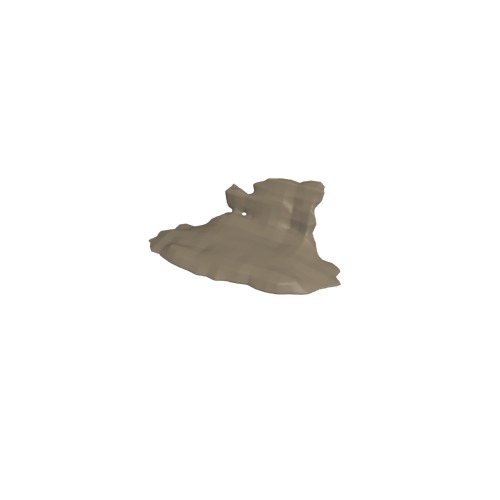}
    \includegraphics[width=0.115\linewidth, trim={\trimR cm \trimR cm \trimR cm \trimR cm}, clip]{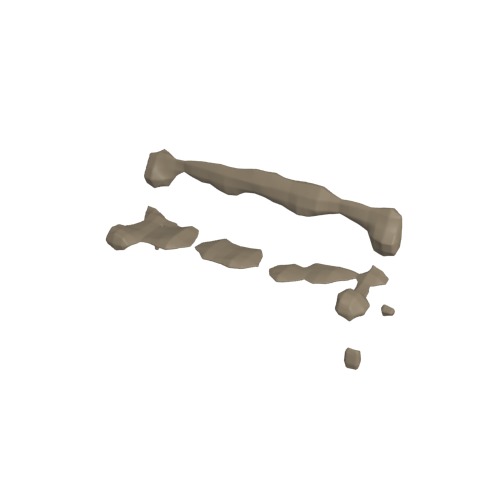}\hfill
    \includegraphics[width=0.115\linewidth, trim={\trimI cm \trimI cm \trimI cm \trimI cm}, clip]{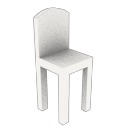}
    \includegraphics[width=0.115\linewidth, trim={\trimR cm \trimR cm \trimR cm \trimR cm}, clip]{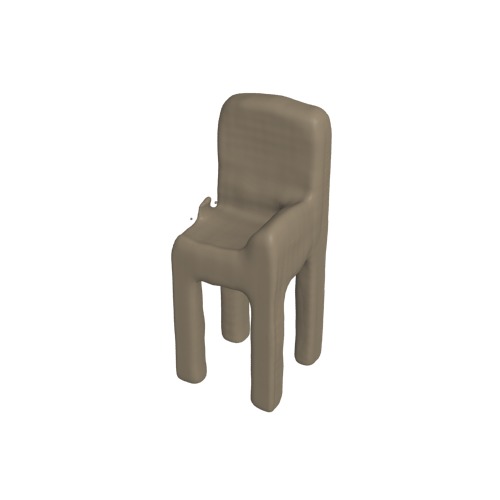}
    \includegraphics[width=0.115\linewidth, trim={\trimR cm \trimR cm \trimR cm \trimR cm}, clip]{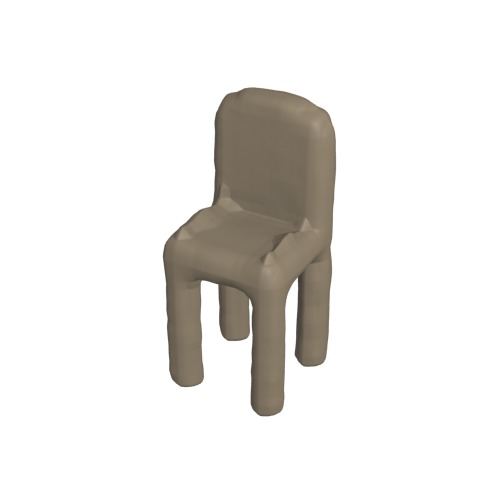}
    \includegraphics[width=0.115\linewidth, trim={\trimR cm \trimR cm \trimR cm \trimR cm}, clip]{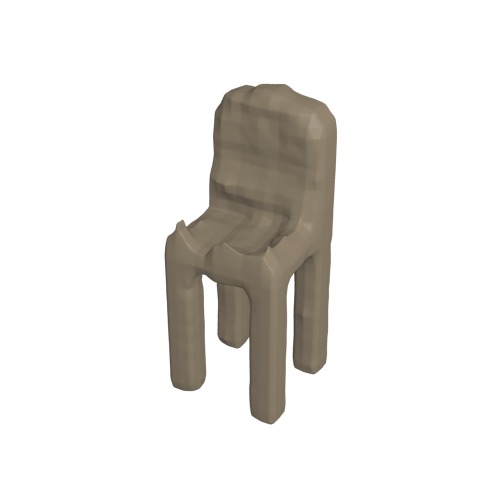}
    \caption{Results for the category chair continued. Each block from left to right, input image, our proposed HSP, LR Soft, LR Hard.} 
    \label{fig:chair2}
  \end{figure*}

\end{appendices}

\end{document}